\documentclass[journal]{IEEEtran}

\usepackage{cite}
\usepackage[pdftex]{graphicx}
\DeclareGraphicsExtensions{.pdf,.jpeg,.png}
\usepackage[caption=false,font=normalsize,labelfont=sf,textfont=sf]{subfig}
\usepackage{algorithmic}
\usepackage{url}

\usepackage{mathtools,amssymb}
\usepackage{multirow,booktabs,longtable}
\newcommand{\spheading}[2][8em]{\rotatebox{90}{\parbox{#1}{\raggedright #2}}}
\usepackage{pbox}

\begin{document}

\title{Fuzzy Clustering to Identify Clusters at Different Levels of Fuzziness: An Evolutionary Multi-Objective Optimization Approach}

\author{Avisek~Gupta,
        Shounak~Datta,
        and~Swagatam~Das,~\IEEEmembership{Senior~Member,~IEEE}% <-this % stops a space
\thanks{A. Gupta, S. Datta, and S.Das are with the Electronics and Communication Sciences Unit, Indian Statistical Institute, 203 B. T. Road, Kolkata 700108, India.}% <-this % stops a space
\thanks{(Corresponding author: Swagatam Das)}%
\thanks{e-mail: avisek003@gmail.com, swagatam.das@isical.ac.in}%
}

%\markboth{IEEE TRANSACTIONS ON SYSTEMS, MAN, AND CYBERNETICS---PART B: CYBERNETICS}%
%{IEEE TRANSACTIONS ON SYSTEMS, MAN, AND CYBERNETICS---PART B: CYBERNETICS}
%{Gupta \MakeLowercase{\textit{et al.}}: Fuzzy Clustering to Identify Clusters at Different Levels of Fuzziness: An Evolutionary Multi-Objective Optimization Approach}

% Remember, if you use this you must call \IEEEpubidadjcol in the second column for its text to clear the IEEEpubid mark.

\maketitle

% ~~~~~~~~~~~~~~~~~~~~~~~~~~~~~ %

\begin{abstract}
Fuzzy clustering methods identify naturally occurring clusters in a dataset, where the extent to which different clusters are overlapped can differ. Most methods have a parameter to fix the level of fuzziness. However, the appropriate level of fuzziness depends on the application at hand. This paper presents Entropy $c$-Means (ECM), a method of fuzzy clustering that simultaneously optimizes two contradictory objective functions, resulting in the creation of fuzzy clusters with different levels of fuzziness. This allows ECM to identify clusters with different degrees of overlap. ECM optimizes the two objective functions using two multi-objective optimization methods, Non-dominated Sorting Genetic Algorithm II (NSGA-II), and Multiobjective Evolutionary Algorithm based on Decomposition (MOEA/D). We also propose a method to select a suitable trade-off clustering from the Pareto front. Experiments on challenging synthetic datasets as well as real-world datasets show that ECM leads to better cluster detection compared to the conventional fuzzy clustering methods as well as previously used multi-objective methods for fuzzy clustering.
\end{abstract}

% ~~~~~~~~~~~~~~~~~~~~~~~~~~~~~ %


%\IEEEpeerreviewmaketitle

% ~~~~~~~~~~~~~~~~~~~~~~~~~~~~~ %

\section{Introduction}

The objective of data clustering is to identify meaningful groups in a collection of data points, so that points within a group are similar and points from different groups are dissimilar. The area of data clustering can be categorized in several ways \cite{xu2005survey}, among which one category is Center-Based Clustering (CBC). In CBC, a cluster is represented by a single point called the center of a cluster. CBC methods aim to estimate the centers for each cluster, which are usually few in number, therefore enabling faster computation. CBC problems can further be of two types - hard or fuzzy. In hard CBC, each data point is assigned a membership in the set $\{0,1\}$ to all clusters. By contrast, in fuzzy CBC, every data point is assigned a membership in the range $[0,1]$ to all clusters. A high membership value indicates that a data point is closer to the center of the corresponding cluster. Fuzzy CBC thus generalizes the membership values from the set $\{0,1\}$ to the interval $[0,1]$. The generalization can be done in several ways, which is called \emph{fuzzification}.

The degree of fuzzification has an important effect on the identification of overlap between clusters. The true overlap between clusters is unknown, since in the problem of data clustering we have no information on the underlying cluster structure. Low degrees of fuzzification assign high memberships for a data point to its closest cluster, and assign low memberships for it to the other clusters. This leads to the formation of clusters that are less overlapped. On the other hand, increasing the degree of fuzziness decreases the memberships of data points to their closest cluster and increases their memberships to all other clusters, forming more overlapped clusters. With the appropriate degree of fuzzification, a fuzzy CBC method can form fuzzy clusters that correspond to the underlying overlapped cluster structure \cite{Klawonn2003}.

The proper extent of fuzzification can thus help identify clusters with varying degrees of overlap. Moreover, the positions of the identified cluster centers are less sensitive to the presence of noise in the dataset compared to hard CBC. With this aim, Dunn first introduced fuzzy CBC with a specific level of fuzziness of the membership values \cite{dunn1973fuzzy}. This was later generalized by Bezdek \cite{Bezdek:1981}, who introduced a parameter $m$ with which one can choose an appropriate level of fuzziness. This gave rise to the well-known Fuzzy $c$-Means (FCM) clustering problem

\begin{equation}\label{eq_fcm_cost}
\text{\emph{minimize  }} J_m = \sum\limits_{i=1}^{N} \sum\limits_{j=1}^{c} \mu_{ij}^{m} || x_i - v_j ||^2,
\end{equation}
subject to the constraints $ \sum\limits_{j=1}^{c} \mu_{ij} = 1 $, where $x_i \in \mathbb{R}^d$, $i=1,...,N$ are the data points to be grouped into $c$ clusters having centers $v_j \in \mathbb{R}^{d}$, for $j=1,...,c$. Each $x_i$ belongs to the $j$-th cluster with a membership degree $\mu_{ij} \in [0,1]$.

The popular algorithm for FCM is a two-step Alternating Optimization (AO) method that alternately updates all $\mu_{ij}$ and then all $v_j$, so as to minimize $J_m$. Bezdek \emph{et al.} \cite{bezdek1984fcm} observed that the AO method works well for values of $m$ in the interval $[1,30]$, with $1.5 \leq m \leq 3.0$ yielding good results for the datasets considered by them. Therefore, it is evident that the same level of fuzzification is not likely to be suitable for all datasets. Indeed, the required extent of fuzzification varies with change in overlap between the clusters as well as the presence of noise. Hence, the proper level of fuzziness is likely to depend on the application at hand \cite{zhu:2009}. Surprisingly, most subsequent works using FCM fix $m$ to $2$. This has resulted in a noticeable lack of investigation on the effects of different levels of fuzziness.

An interesting study on this topic is that of Li and Mukaidono \cite{dsr95} who formulated a Maximum Entropy Inference (MEI) approach to fuzzy CBC,
\begin{equation} \label{eq_cost_mei}
\text{\emph{maximize  }} E = -\sum\limits_{i=1}^{N} \sum\limits_{j=1}^{c} \mu_{ij}log(\mu_{ij})\ ,
\end{equation}
subject to the constraint $ \sum\limits_{j=1}^{c} \mu_{ij} = 1 $, and the soft constraint $\sum\limits_{i=1}^{N} \sum\limits_{j=1}^{c} \mu_{ij} ||x_i - v_j||^2 = 0$. This formulation allowed them to design an AO algorithm for MEI which uses an \emph{admissible error radius} $\sigma$ instead of $m$, which is advantageous as $\sigma$ is easier to interpret compared to $m$. Increasing $\sigma$ leads to a physical increase in the spread of the clusters, thereby increasing the fuzziness of the membership values. A recent work by Saha and Das \cite{Saha2017} proposed an axiomatic definition of a general class of weighting functions that can be used to control the level of fuzzification. However, there is still a lack of studies on the effect of different levels of fuzziness on fuzzy CBC, which we consider to be an area worth investigating.

In this paper, we show that clusterings at different levels of fuzziness can be obtained by simultaneously optimizing two contradicting objectives in a Multi-Objective Optimization (MOO) framework. In the proposed Entropy $c$-Means (ECM) method, one objective favors the formation of compact fuzzy clusters while the other objective strives for largely overlapping clusters. In an MOO framework, multiple objective functions are optimized simultaneously, leading to a set of trade-off solutions called \emph{Pareto-optimal} solutions, where no solution is better than the other when considering all objective function values at once (see Section \ref{sec_moo}). MOO methods have long been used for data clustering \cite{mukho14,xia_13,wang_13}. If the objectives are contradictory, then a wide variety of Pareto-optimal solutions can be found. We show that the two objectives of ECM are indeed in conflict and hold a strong Pareto relation. This leads to a wide variety of Pareto-optimal clusterings corresponding to different levels of fuzziness. In Section \ref{sec_exp_results}, we conduct experiments to compare the proposed ECM method against state-of-the-art fuzzy CBC methods. We also propose a method to select a trade-off clustering from the Pareto-front identified by ECM in section \ref{sec_tradeoff}.

% ~~~~~~~~~~~~~~~~~~~~~~~~~~~~~ %

\section{Fuzzy Clustering using Multi-Objective Optimization}
\label{sec_method}

In this section we first define the framework of MOO problems, after which we elucidate the two objective functions which are optimized in ECM within an MOO framework to generate clusterings with different levels of fuzziness. Finally, we describe how the ECM problem can be solved by using two popular MOO methods, namely Non-dominated Sorting Genetic Algorithm II (NSGA-II) \cite{nsga2}, and Multiobjective Evolutionary Algorithm based on Decomposition (MOEA/D) \cite{zhang07} \cite{zhang_09}.

% ~~~~~~~~~~~~~~~~~~~~~~~~~~~~~ %

\subsection{Multi-Objective Optimization}\label{sec_moo}

Let there be $n$ objective functions to be optimized, subject to $p$ constraint functions. Problems of this form can be written as,
\begin{equation}
\begin{aligned}[b]
& \text{\emph{maximize/minimize  }} & f_i(x) \ \ &,\ i = 1,2,\hdots,n \\
& \text{\emph{subject to  }} & g_j(x) \leq 0 \ &,\ j=1,2,\hdots,p,
\end{aligned}
\end{equation}
where $x$ is a feasible solution satisfying all $g_j$. $x$ is said to dominate $y$, if for all $i$, $f_i(x)$ is at least as optimal as $f_i(y)$, and there exists at least one $f_i(x)$ more optimal than the corresponding $f_i(y)$. A feasible solution $x^*$ is said to be \emph{optimal}, if for all feasible $y$, $x^*$ dominates $y$. In MOO problems, we usually cannot find single optimal solutions as the objectives are contradictory in nature. Instead we find a set of mutually non-dominating solutions called \emph{Pareto-optimal solutions}. The set of images of the Pareto-optimal solutions in the objective space is called the \emph{Pareto front}. For a survey on methods used to solve MOO problems, see Zhou \emph{et al.} \cite{evo11}.

% ~~~~~~~~~~~~~~~~~~~~~~~~~~~~~ %

\subsection{Objective Functions for Fuzzy Clusters} \label{sec_2_obj_funcs}

\begin{figure*}
	\centering
	\subfloat[Original dataset]{\includegraphics[width=0.32\textwidth]{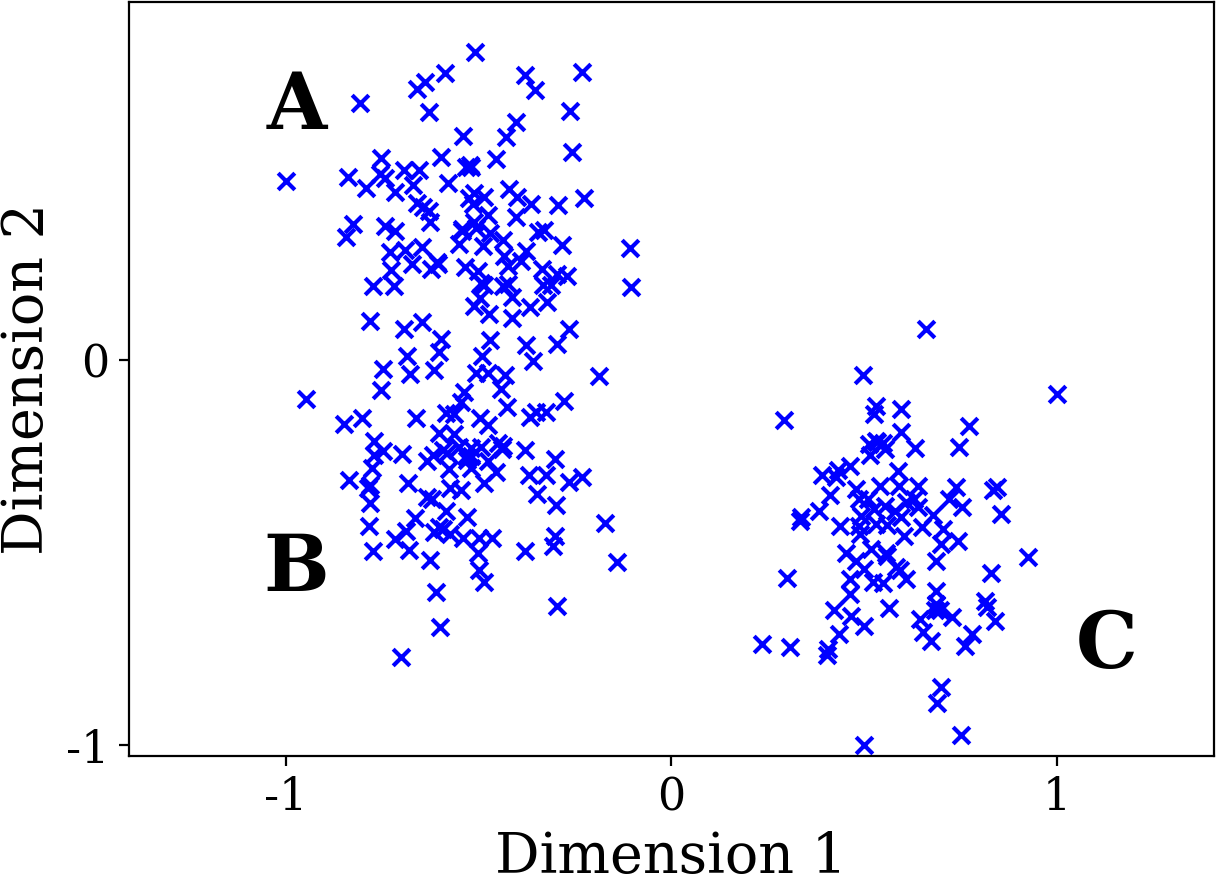}
		\label{fig_c3}}
	\hfil
	\subfloat[Pareto Front]{\includegraphics[width=0.32\textwidth]{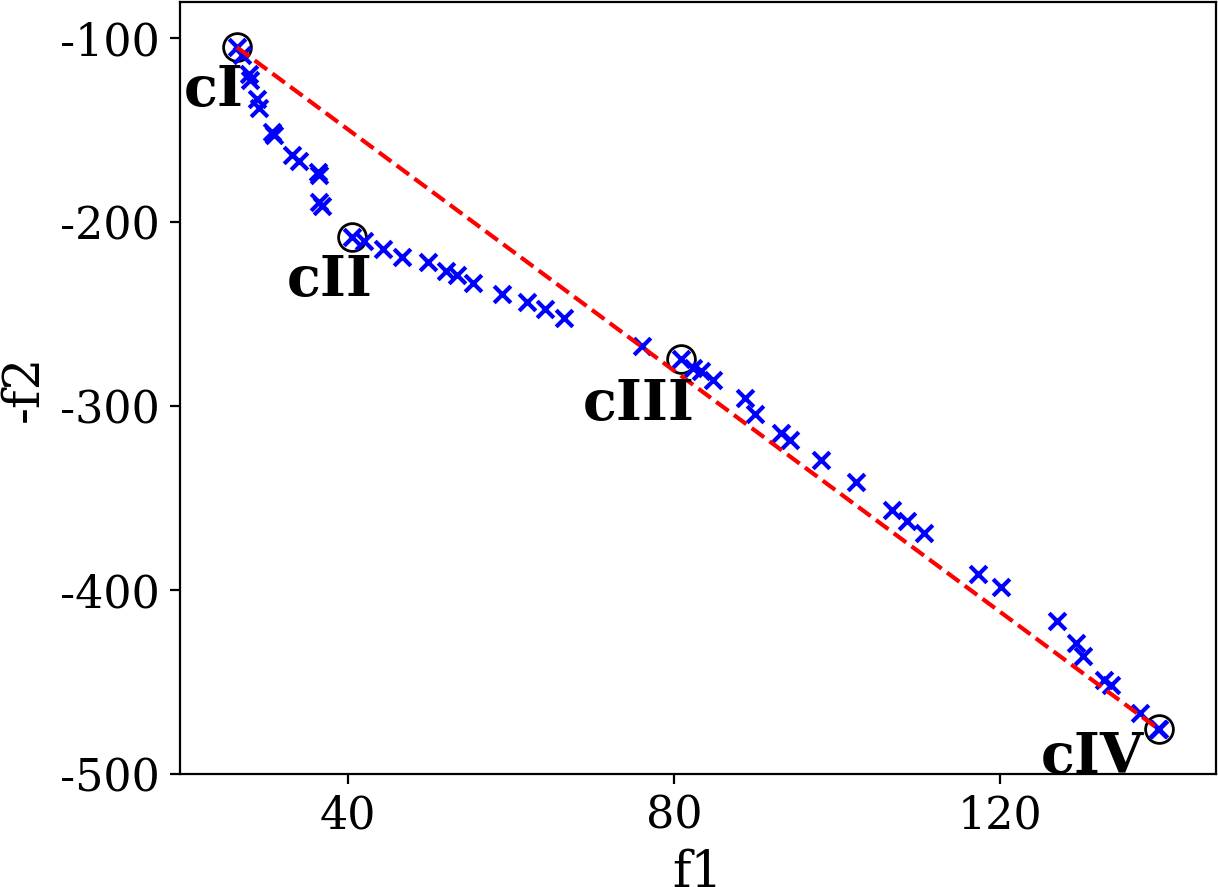}
		\label{fig_c3_pf}}
	\hfil
	\subfloat[Clusters formed in cI]{\includegraphics[width=0.32\textwidth]{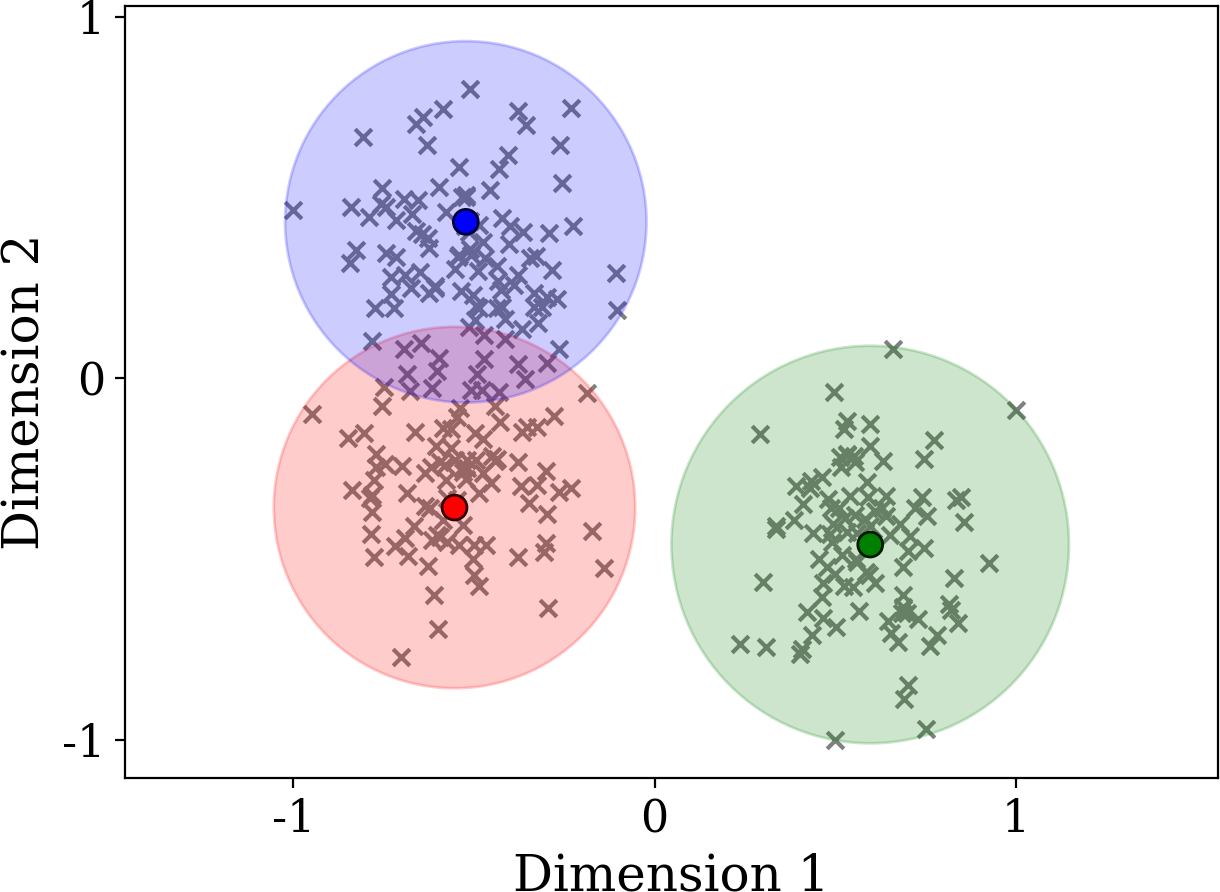}
		\label{fig_c3_cI}}

	\subfloat[Clusters formed in cII]{\includegraphics[width=0.32\textwidth]{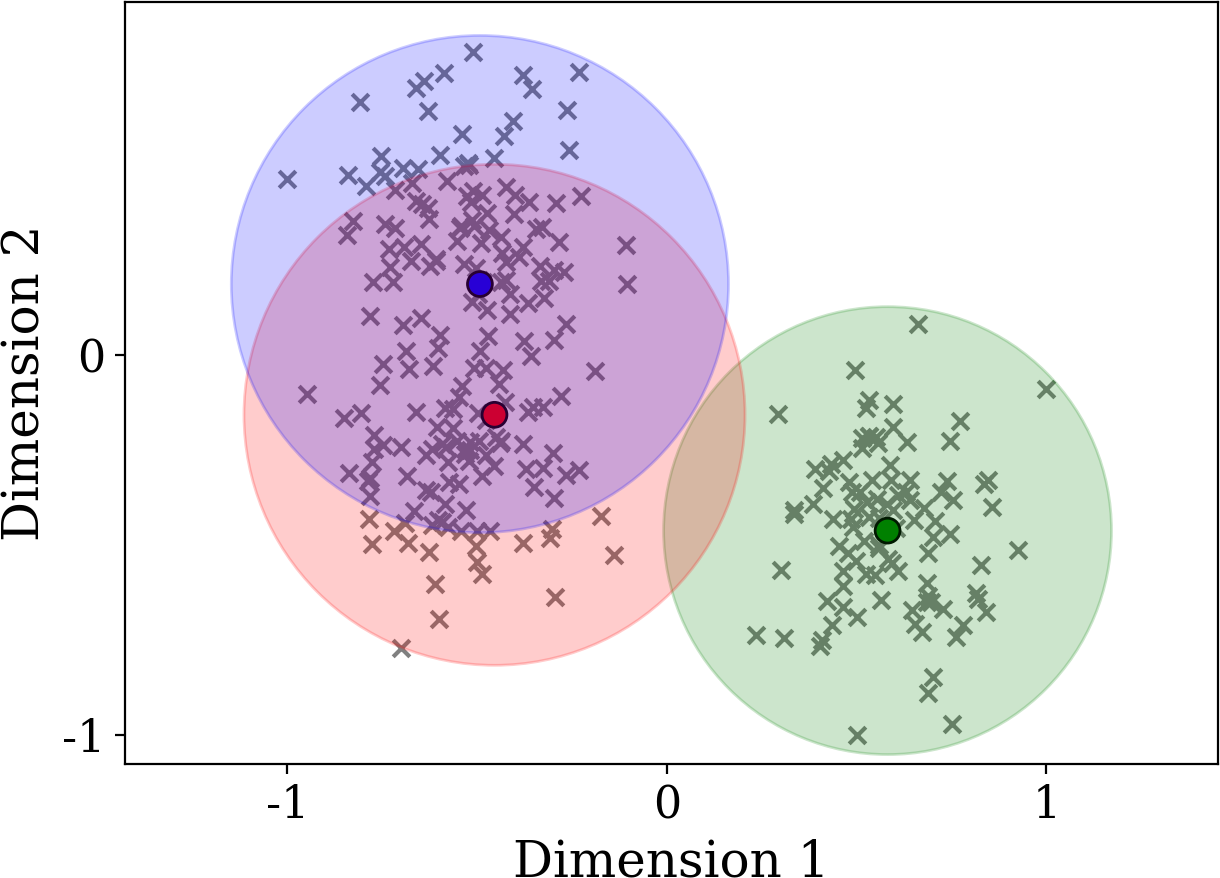}
		\label{fig_c3_cII}}
	\hfil
	\subfloat[Clusters formed in cIII]{\includegraphics[width=0.32\textwidth]{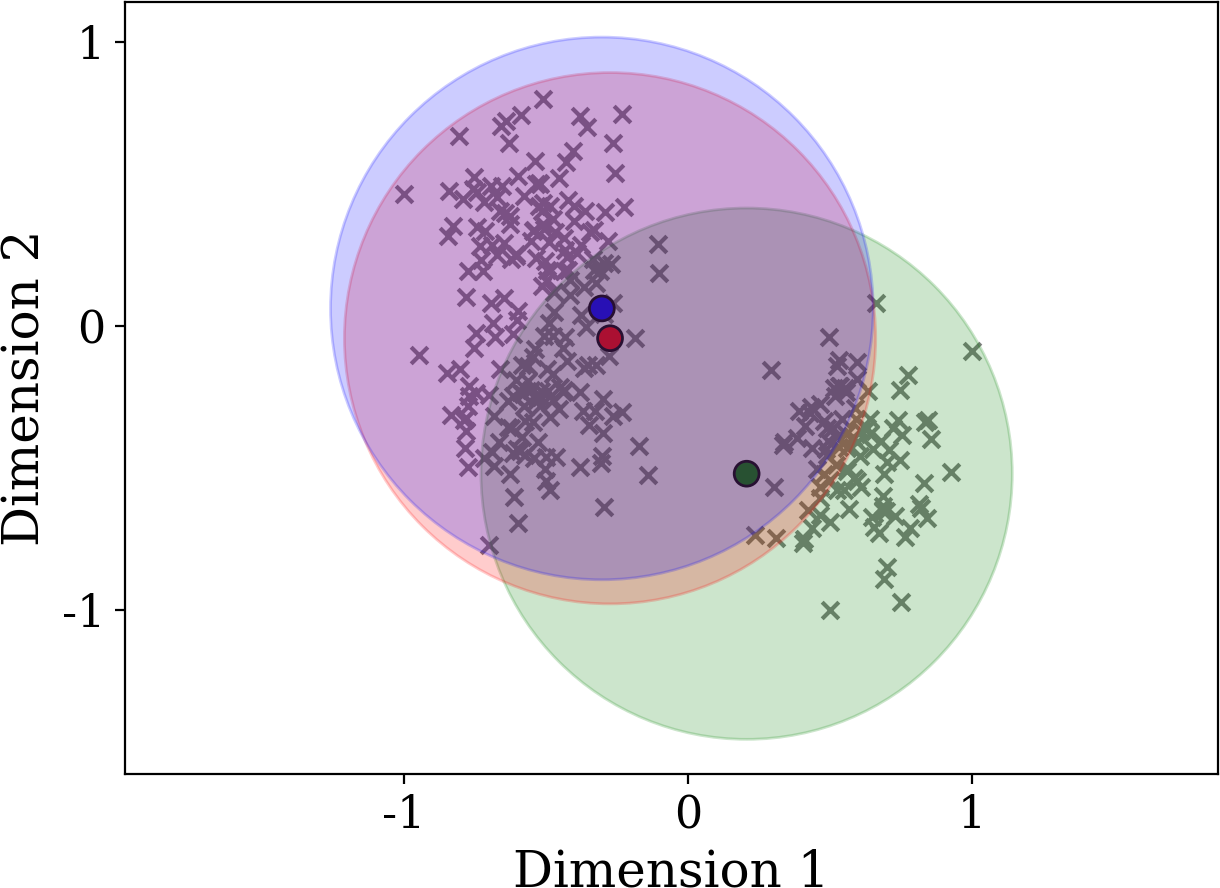}
		\label{fig_c3_cIII}}
	\hfil
	\subfloat[Clusters formed in cIV]{\includegraphics[width=0.32\textwidth]{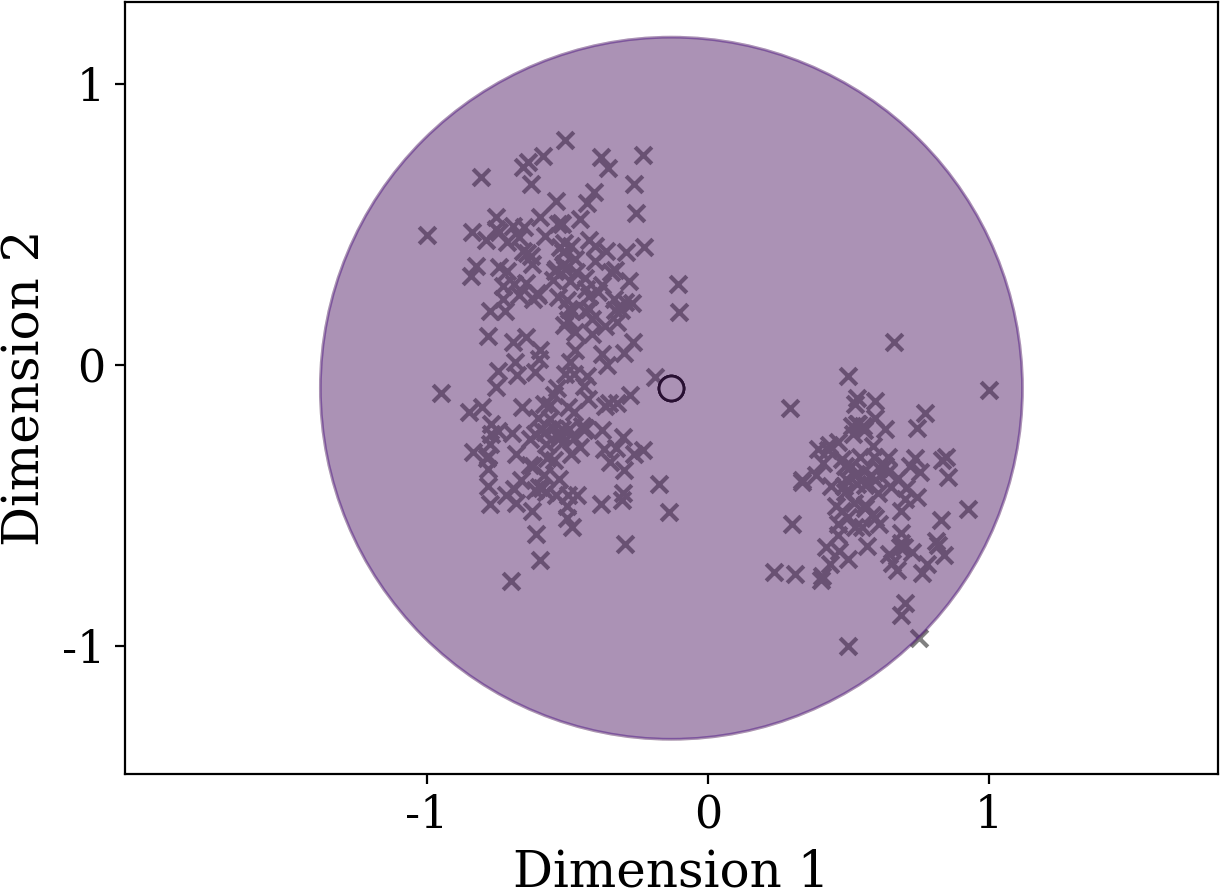}
		\label{fig_c3_cIV}}
	\caption{Clustering a synthetic dataset (a) using ECM-NSGA-II. The dataset contains two overlapped clusters \textbf{A} and \textbf{B}, and a third well-separated cluster \textbf{C}. At the top left corner of the Pareto front in (b), $f_1$ is minimized creating compact clusters with low overlap (c). At the bottom right corner, $f_2$ is maximized by minimizing $-f_2$, leading to more overlapped clusters as shown in (f). Across the Pareto front clusters formed have different levels of fuzziness, see (d) and (e).}
	\label{fig_fuzziness_across_pf}
\end{figure*}

In ECM, we propose the following two objective functions to be optimized in an MOO framework,

\begin{subequations}
\label{eqn_two_objs}
\begin{align}
\text{\emph{minimize }} & f_1 = \sum\limits_{i=1}^{N} \sum\limits_{j=1}^{c} \mu_{ij}||x_i - v_j||^2 \label{eqn_obj_f1}\ , \\
\text{\emph{maximize }} & f_2 = - \sum\limits_{i=1}^{N} \sum\limits_{j=1}^{c} \mu_{ij}log(\mu_{ij}) \label{eqn_obj_f2}\ .
\end{align}
\end{subequations}
One aim of CBC methods is to place the cluster centers in areas having high density of data points. Minimizing the sum of cluster-wise \emph{variances} ensures the formation of compact clusters. The objective function $f_1$ given in \eqref{eqn_obj_f1} generalizes the idea of variance by weighting the distance between data points and cluster centers with the membership degrees of data points in each cluster. This objective of \emph{cluster compactness} is to minimize the function $f_1$ to form compact clusters.

We propose the use of the objective function $f_2$ in \eqref{eqn_obj_f2} to avoid a fixed level of fuzzification (specified by $m$ in FCM). $f_2$ is the \emph{entropy} of membership values \cite{dsr95}, which is maximized when all membership values are equal to $1/c$, where $c$ is the number of clusters. If a data point is close to a cluster center, the corresponding membership value will be greater than $1/c$. On the other hand, if a data point is far from a specific cluster center, its membership to that cluster will be less than $1/c$. Increasing the entropy brings all the membership values for a data point close to $1/c$. This essentially increases the level of fuzzification of the membership values.

\emph{\textbf{Remark.} The two objectives to minimize $f_1$ and to maximize $f_2$ in \eqref{eqn_two_objs} are contradicting.}

This can be observed easily. To minimize $f_1$, for all $i$ we set $\mu_{ij} = 1$ if $d_{ij} = ||x_i - v_j||$ is minimum, and $0$ for all other $j$. However, for such values of $\mu_{ij}$ we get $f_2 = 0$, which is the minimum value of $f_2$. Thus minimizing $f_1$ does not maximize $f_2$.

To maximize $f_2$, we first form the Lagrangian for the maximization of $f_2$, subject to the constraint $\sum_{j=1}^{c} \mu_{ij} = 1$, as:

\begin{equation} \label{eq_lagrangian_f_2}
\mathbb{L} = -\sum_{i=1}^{N} \sum_{j=1}^{c} \mu_{ij} log(\mu_{ij}) + \sum_{i=1}^{N} \lambda_{i} (\sum_{j=1}^{c} \mu_{ij} - 1)\ .
\end{equation}

From the derivative of the Lagrangian, w.r.t. $\mu_{ij}$ and $\lambda_{i}$, we get $\mu_{ij} = 1/c$. This value of $\mu_{ij}$ maximizes $f_2$. However, as already observed, $f_1$ is minimized when $\mu_{ij} = 1$ if $d_{ij}$ is minimum, and $0$ for all other $j$. Hence, the values of $\mu_{ij}$ that maximize $f_2$ do not minimize $f_1$.

Thus the above remark shows that the two objectives of minimizing $f_1$ and maximizing $f_2$ are contradicting. Optimizing both in an MOO framework leads to a Pareto-optimal set of solutions representing the best compromises between the objectives. A wide Pareto front is formed due to the contradictory nature of the objectives, identifying clusterings with different levels of fuzziness.
Fig. \ref{fig_c3} shows a synthetic dataset containing three clusters $A$, $B$, and $C$, where $A$ and $B$ are close compared to $C$. The Pareto front obtained by solving ECM using the NSGA-II algorithm (to be discussed in Section \ref{sec_nsga2}) is shown in Fig. \ref{fig_c3_pf}. The clustering at end cI of the front has minimum value of $f_1$, with the estimated clusters being compact having little overlap as shown in Fig. \ref{fig_c3_cI}. As we move towards the end cIV along the front, the clusterings become increasingly fuzzy due to the increase in $f_2$ (and a consequent increase in $f_1$), characterized by increasing overlap between the estimated clusters. As fuzziness increases from cI to cIV, the estimated clusters corresponding to closer clusters $A$ and $B$ first become overlapped before getting overlapped with that of $C$, as expected (see Figs. \ref{fig_c3_cII}-\ref{fig_c3_cIV}).

\begin{figure}
\centering
	\subfloat[Clustering for $m=2$]{\includegraphics[width=0.23\textwidth]{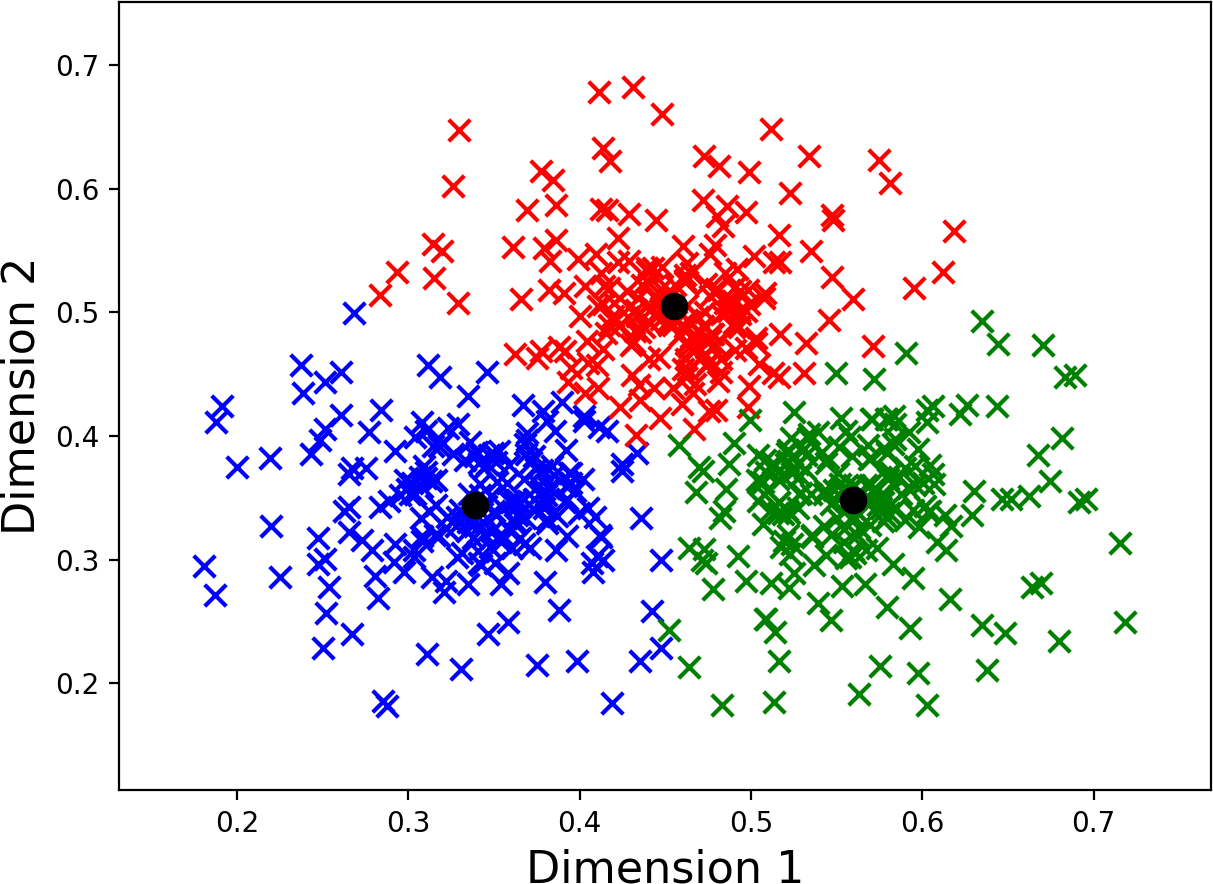}
		\label{fig_m2}}
	\hfil
	\subfloat[Clustering for $m=30$]{\includegraphics[width=0.23\textwidth]{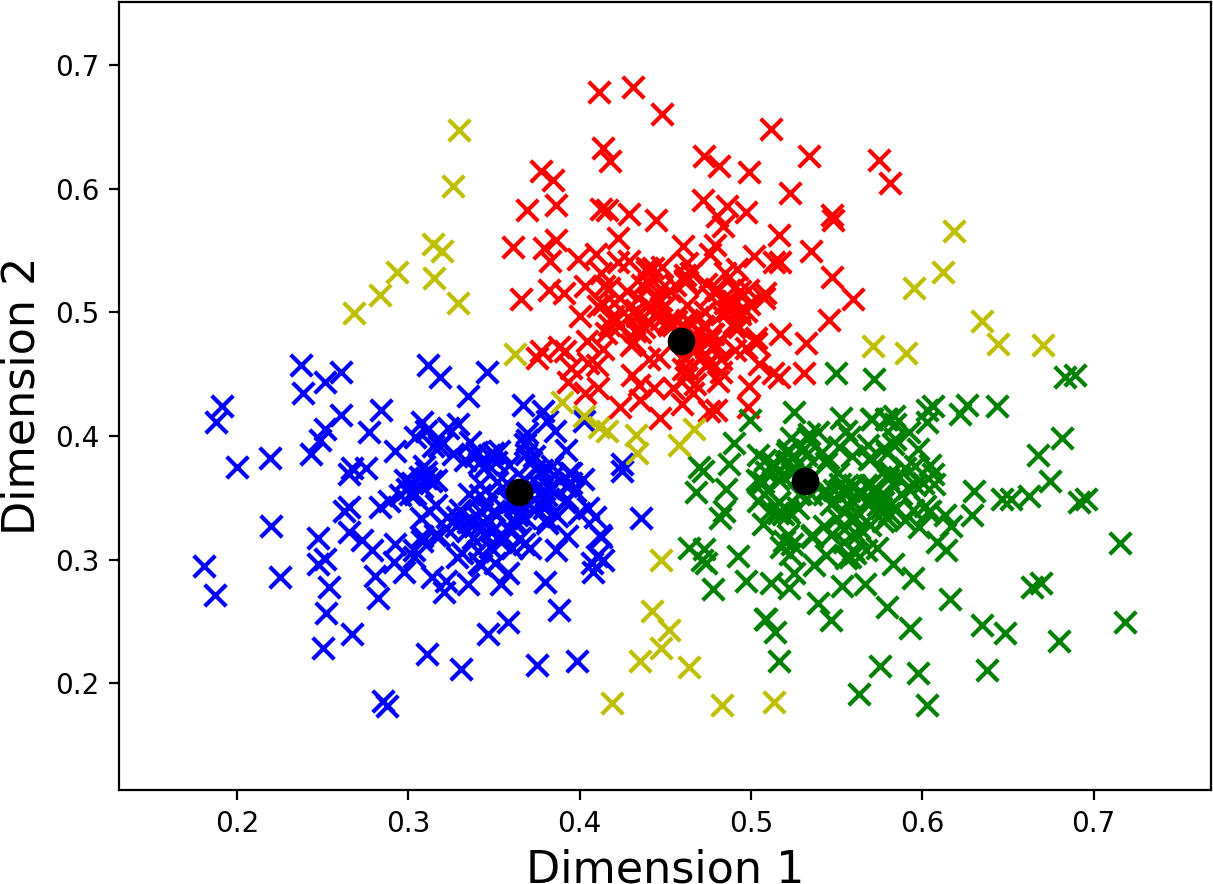}
		\label{fig_m30}}

	\subfloat[Clustering for $m=50$]{\includegraphics[width=0.23\textwidth]{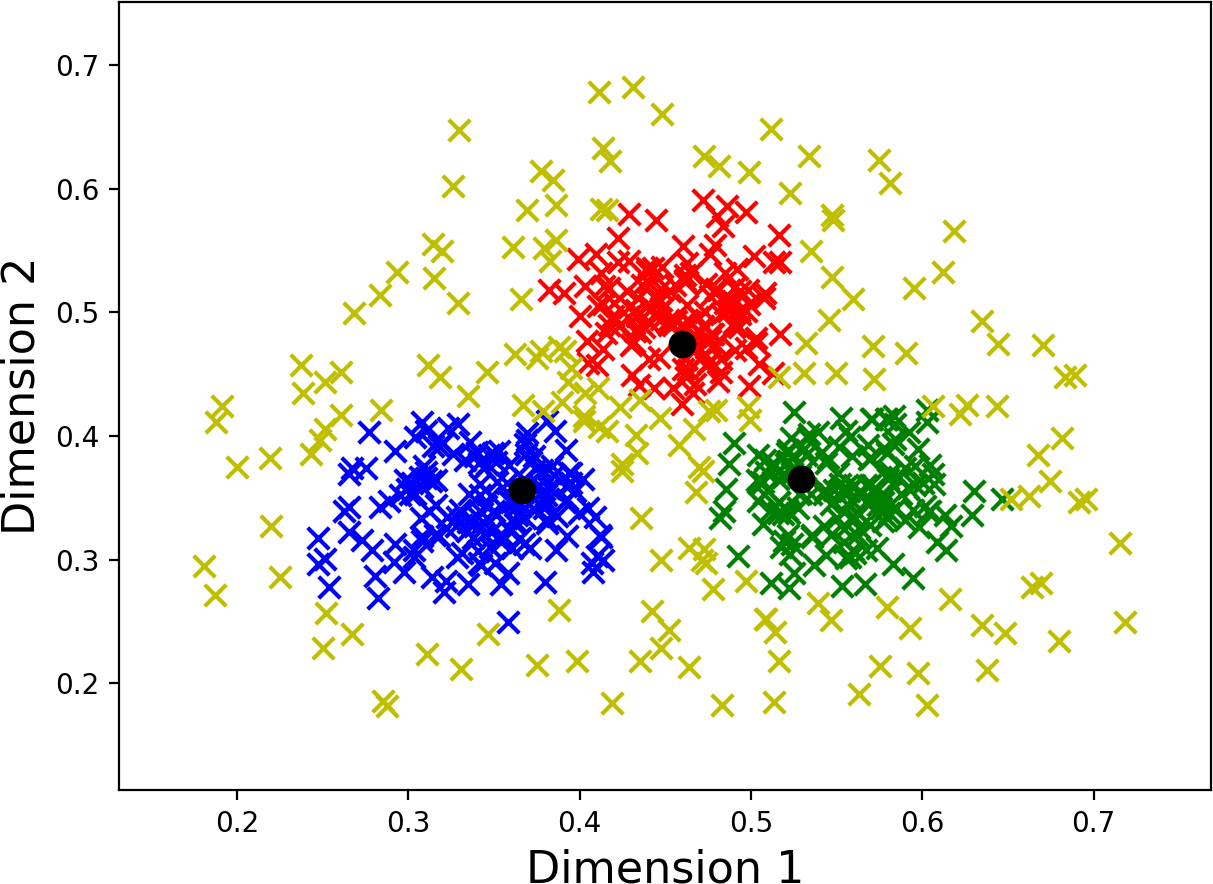}
		\label{fig_m50}}
        \hfil
	\subfloat[Clustering for $m=100$]{\includegraphics[width=0.23\textwidth]{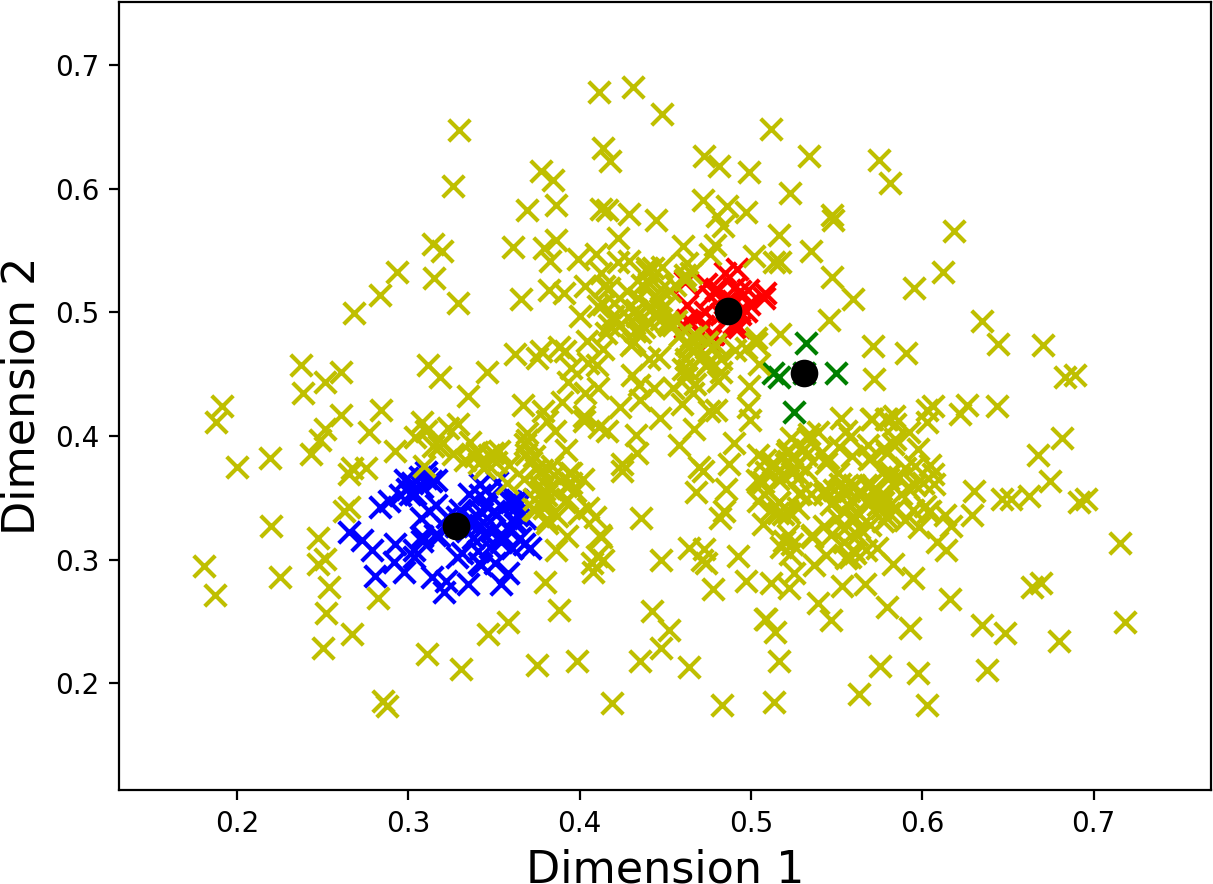}
		\label{fig_m100}}
	\caption{Clustering of a dataset with $3$ clusters for different levels of fuzziness $m = 2, 30, 50,$ and $100$. The points in different clusters are drawn in red, blue and green and the points in the regions of overlap between the clusters are drawn in yellow. For higher values of $m$, a large number of points in the regions of overlap do not contribute to the location of cluster centers.}
	\label{fig_m_figs}
\end{figure}

A simple alternative might be to run FCM for different valeus of $m$. However, the AO algorithm results in erroneous solutions for higher values of $m$. As shown in Figure \ref{fig_m_figs}, for higher values of $m$, there are data points that are assigned equal probability to all clusters. They do not contribute to the location of cluster centers, leading to convergence close to the inital center locations. The multi-objective method of ECM avoids this error. In Figure \ref{fig_c3_cIV} we observe that in the end of high fuzzification shown in clustering cIV, all the centers converge to the center of the dataset, as should occur at high levels of fuzzifiness due to the equal contribution of every points to all clusters. Another close attempt by MECA \cite{karayiannis_94} identifies crisp clusters using a linear combination of the cluster compactness and the entropy of memberships, weighted by a fuzzification parameter $\alpha$. MECA starts with $\alpha$ close to 1 where it considers maximum entropy, and over the iterations $\alpha$ goes close to 0 to identify crisp clusters at convergence. However, MECA does not have a stopping criteria to identify fuzzy clusters. MECA also requires specifying initial and final values to the fuzzification parameter, as well as how to decrease its value over the number of iterations. In general, on can note a renewed interest in the possible uses of the entropy of cluster memberships \cite{zhi_13,choy_17,yang_17}. The merit of our method of ECM is that by optimizing both objectives in a multi-objective setting, ECM identifies Pareto-optimal fuzzy clusters at different levels of fuzziness, from which an optimal fuzzy clustering can be selected for the application at hand.

\subsection{Multi-Objective Optimization Methods for ECM}

We use two popular evolutionary MOO methods, NSGA-II \cite{nsga2} and MOEA/D \cite{zhang07} \cite{zhang_09}, to solve the MOO problem posed by ECM. Both methods are briefly described next, accompanied with pseudo-codes to provide an overview of how they find the Pareto-optimal clusterings. For both methods, the solution vectors are formed by appending all cluster centers into a single vector.

% ~~~~~~~~~~~~~~~~~~~~~~~~~~~~~ %

\subsubsection{ECM-NSGA-II}\label{sec_nsga2}

\begin{figure}[b]
{
\footnotesize
\begin{algorithmic}
	\STATE \hrule
	\STATE \textbf{Algorithm I : The algorithm for ECM-NSGA-II}
	\STATE \hrule
	\STATE \textbf{Input:} Number of clusters $c$; Population size $pop$; number of function evaluations (\emph{FE}).
	\STATE \textbf{Output:} A set of $pop$ non-dominated clusterings.
    \vspace{+1mm}
	\hrule
    \vspace{+1mm}
	\STATE Randomly initialize all $pop$ chromosomes as vectors of $c$ candidate cluster centers.
	\REPEAT
		\FOR {each chromosome}
		\STATE Evaluate the two objective functions in \eqref{eqn_two_objs}.
		\STATE Perform non-dominated sorting to compute the rank.
		\STATE Compute the crowding distance.
		\ENDFOR
		\STATE Perform crossover and mutation to generate a new population.
		%\STATE Combine the previous and new population.
		\STATE Use binary tournament selection to select $pop$ chromosomes from the previous and new population, for the next iteration.
	\UNTIL{\emph{FE} function evaluations are performed.}
    \vspace{+1mm}
	\STATE \hrule
\end{algorithmic}
}
\end{figure}

In each iteration of ECM-NSGA-II, a set of chromosomes is maintained called a \emph{population}, so that multiple possible clusterings can be considered simultaneously. For each chromosome, the membership values can be computed using the centers according to the update expression derived from the Lagrangian of the cost function \eqref{eq_cost_mei} of MEI:

\begin{equation}
\mu_{ij} = e^{-d_{ij}^2} \Big/ \sum_{j=1}^{c} e^{-d_{ij}^2}  \ .
\end{equation}

The operations of \emph{crossover} or \emph{mutation} are performed to search for chromosomes yielding better objective function values. In crossover, two chromosomes can be combined to form two new chromosomes. In mutation, a single value in a chromosome can be altered. Chromosomes in a population are assigned ranks using the \emph{non-dominated sorting}, where all chromosomes in a Pareto front are assigned the same rank. Each chromosome on a Pareto front is assigned a \emph{crowding distance}, which is high for chromosomes with less chromosomes around it. \emph{Elitist} selection of the chromosomes is undertaken for the next iteration using a \emph{binary tournament selection}, where (i) higher ranked chromosomes have a higher probability of selection, and (ii) between chromosomes with the same rank, those with higher crowding distance have a higher probability of getting selected. A pseudo-code for ECM-NSGA-II is given in Algorithm I.

% ~~~~~~~~~~~~~~~~~~~~~~~~~~~~~ %

\subsubsection{ECM-MOEA/D}

ECM-MOEA/D decomposes an MOO problem with $n$ objective functions into $pop$ number of single objective optimization problems by using the Tchebycheff approach \cite{zhang07}. In each iteration, the population contains the $pop$ best solutions that have been found. Each subproblem is then optimized using information from its neighbouring subproblems. \emph{Elitism} is maintained by periodically adding newly generated non-dominated solutions to an \emph{External Population} (EP) and discarding solutions from it that are no longer non-dominated. A pseudo-code for ECM-MOEA/D is provided in Algorithm II.

\begin{figure}[b]
{
\footnotesize
\begin{algorithmic}
	\STATE \hrule
	\STATE \textbf{Algorithm II : The algorithm for ECM-MOEA/D}
	\STATE \hrule
	\STATE \textbf{Input:} Number of clusters $c$; number of subproblems $pop$; coefficient vectors with uniform spread $\lambda_1,\hdots,\lambda_{pop}$; number of neighbours $T$; number of function evaluations \emph{FE}.
	\STATE \textbf{Output:} The external population (EP) containing non-dominated clusterings.
    \vspace{+1mm}
	\hrule
    \vspace{+1mm}
	\STATE \textbf{Initialization : }
	\STATE Find the $T$ nearest neighbours for each coefficient vector.
	\STATE Initialize and evaluate $x_1,\hdots,x_{pop}$ of solution vectors each formed of $c$ candidate cluster centers.
	\STATE On all $x_i$ evaluate the two objective functions in \eqref{eqn_two_objs} and store the best values as $z_1$ and $z_2$.
	\STATE \textbf{Iteration : }
	\REPEAT
		\FOR{Each coefficient vector $\lambda_i$}
			\STATE Randomly select two out of its $T$ neighbours $\lambda_j$, $\lambda_l$, and retrieve the corresponding solutions $x_j$, $x_l$.
			\STATE Apply Differential Evolution mutation and crossover \cite{zhang_09} on $x_i, x_j$ and $x_l$ to form a new solution $y$.
			\STATE If $f_1(y) < z_1$ (and/or $f_2(y) > z_2$), then update $z_1$ (and/or $z_2$).
			\STATE If $y$ improves upon the current best solution of $\lambda_i$ or any of its $T$ neighbours, then replace the corresponding solution with $y$.
			\STATE Remove all vectors from the EP that are dominated by $y$, and add $y$ to the EP if none of the existing members dominate it.
		\ENDFOR
	\UNTIL{\emph{FE} function evaluations are performed.}
    \vspace{+1mm}
	\STATE \hrule
\end{algorithmic}
}
\end{figure}

% ~~~~~~~~~~~~~~~~~~~~~~~~~~~~~ %

\subsection{Computation Complexity of ECM}

The time complexity of NSGA-II is $O(n(pop)^2)$ \cite{nsga2}, whereas that of MOEA/D is $O(n(pop)T)$ \cite{zhang07}. The evaluation of both objective functions requires the computation of the memberships of $N$ data points to $c$ clusters. This takes $O(Nc)$ time. As $n=2$ for ECM, every iteration of ECM-NSGA-II takes overall $O(\max\{2(pop)^2, Nc\})$ time, and that of ECM-MOEA/D takes $O(\max\{2(pop)T, Nc\})$ time. In real-world scenarios, usually $N>>c$ as well as $N>>(pop)^2$, leading to an $O(Nc)$ time complexity per iteration for both variants of ECM.

% ~~~~~~~~~~~~~~~~~~~~~~~~~~~~~ %

\section{Experiments and Results}
\label{sec_exp_results}

We conduct various experiments to evaluate the clustering performance of ECM-NSGA-II and ECM-MOEA/D \footnote{MATLAB implementations available at \url{https://github.com/Avisek20/ecm}.}. For convenience of implementation, we minimize $-f_2$ instead of maximizing $f_2$ as per \eqref{eqn_obj_f2}. The clustering performances of ECM-NSGA-II and ECM-MOEA/D are compared with the conventional AO algorithms for FCM and MEI. Further comparison is undertaken with the Multi-Objective Genetic Algorithm (MOGA) \cite{mogaclust06} and MOGA based Support Vector Machines (MOGA-SVM) \cite{mogasvm09}, which are two existing MOO methods used to find clusterings when the number of clusters is known. Both MOGA and MOGA-SVM use $J_2$ ($J_m$ as in eqn. \eqref{eq_fcm_cost} with $m=2$) and the Xie-Beni index \cite{xie91} to find a fixed number of clusters. However, since the Xie-Beni index simply scales $J_m$ by the distance between the closest pair of cluster centers, these two functions are not generally contradicting. Hence, the use of these two objectives is not likely to give rise to a large variety of trade-off solutions (see Fig. \ref{fig_pareto_comp}).

In Section \ref{sec_mock_datasets} and \ref{sec_real_datasets}, we compare the performance of all methods on synthetic and real datasets. We also propose a method to select a clustering from the Pareto Front of ECM in Section \ref{sec_tradeoff}. In section \ref{sec_pf} we compare the Pareto fronts obtained by the contending MOO methods.

% ~~~~~~~~~~~~~~~~~~~~~~~~~~~~~ %

\subsection{Synthetic Datasets} \label{sec_mock_datasets}

\begin{figure}
\centering
	\subfloat[\emph{proximity1}]{\includegraphics[width=0.15\textwidth]{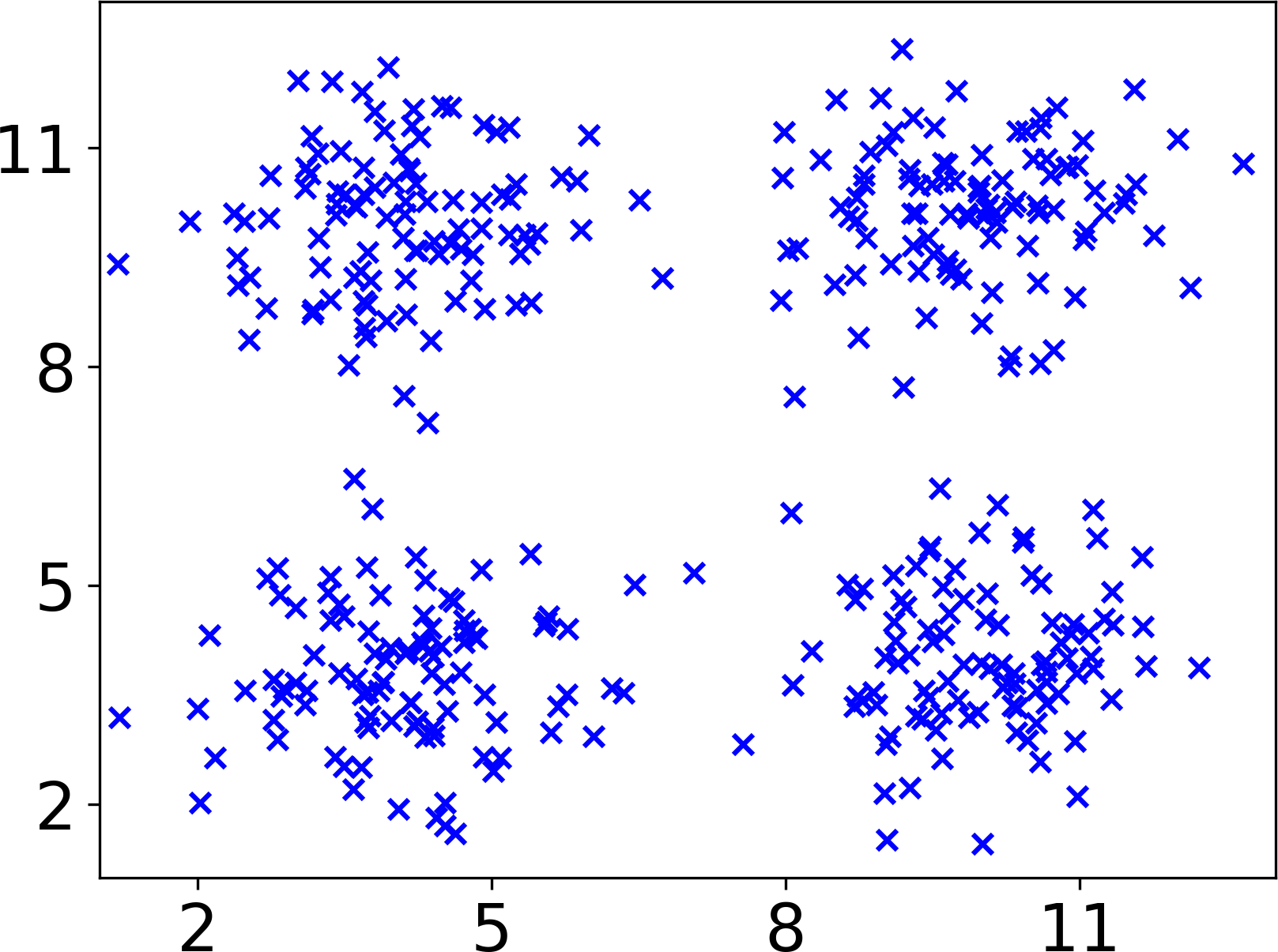}
		\label{fig_proximity1}}
	\hfil
    \subfloat[\emph{proximity2}]{\includegraphics[width=0.15\textwidth]{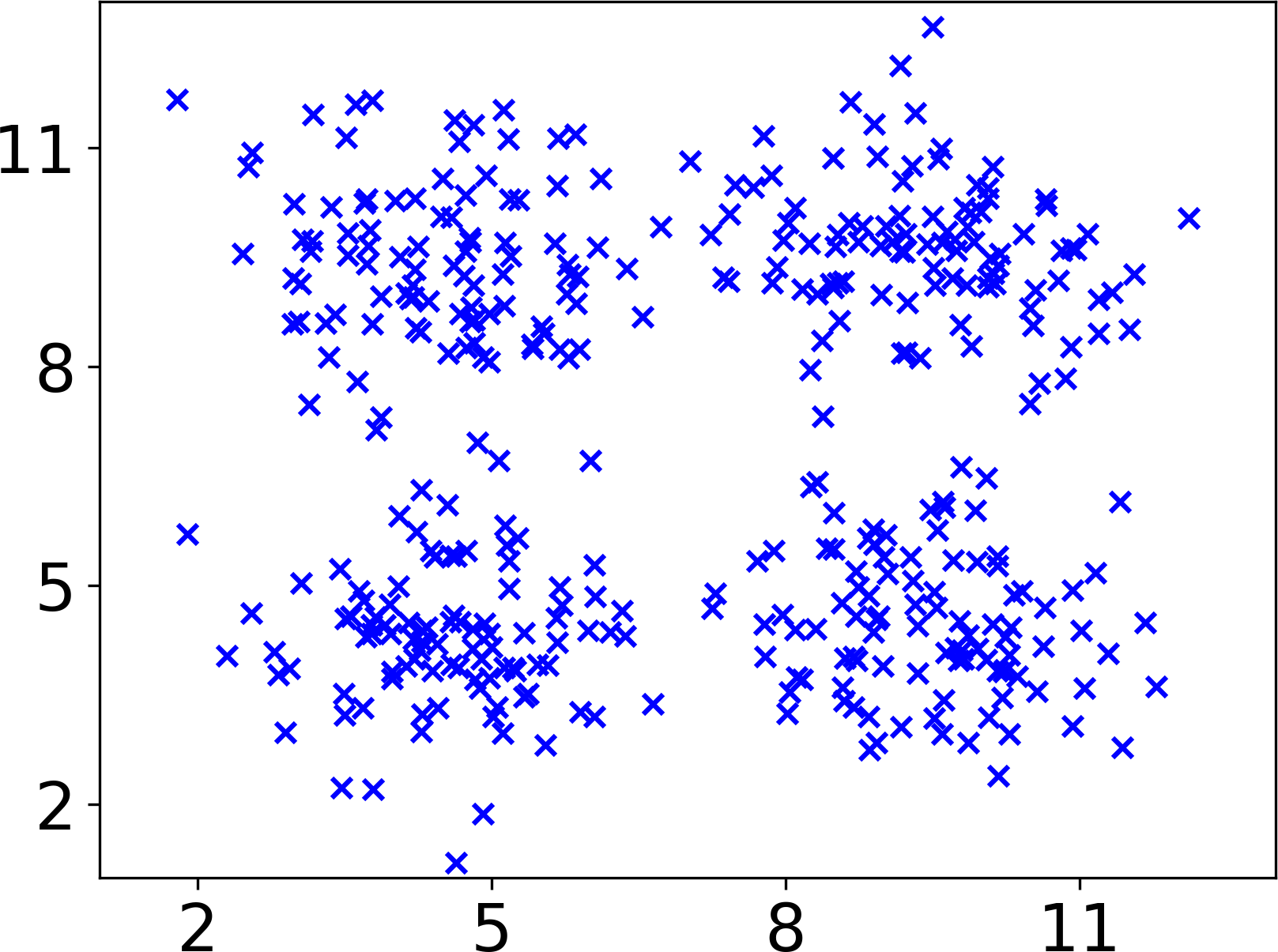}
		\label{fig_proximity2}}
	\hfil
	\subfloat[\emph{proximity3}]{\includegraphics[width=0.15\textwidth]{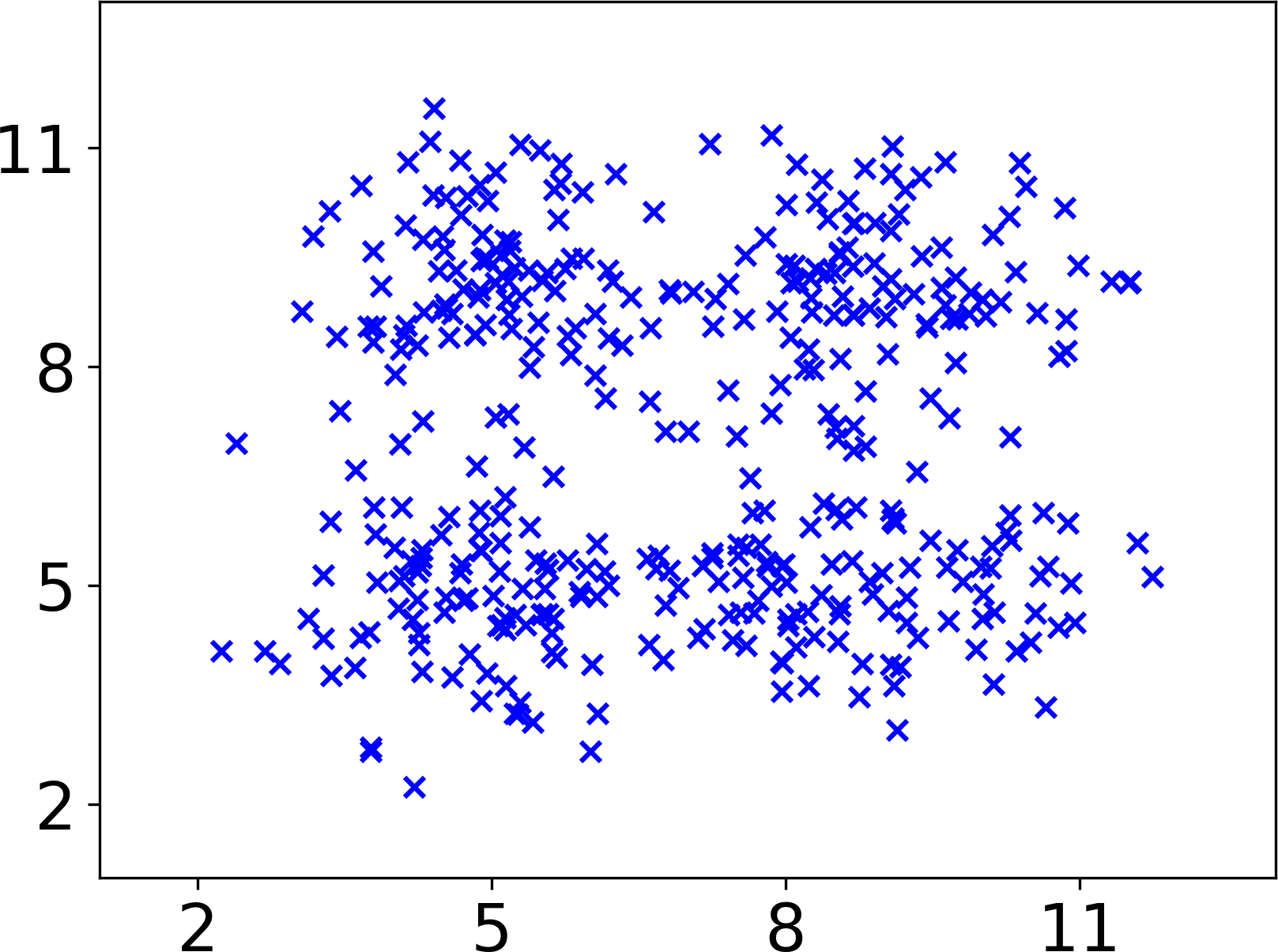}
		\label{fig_proximity3}}

    \subfloat[\emph{proximity4}]{\includegraphics[width=0.15\textwidth]{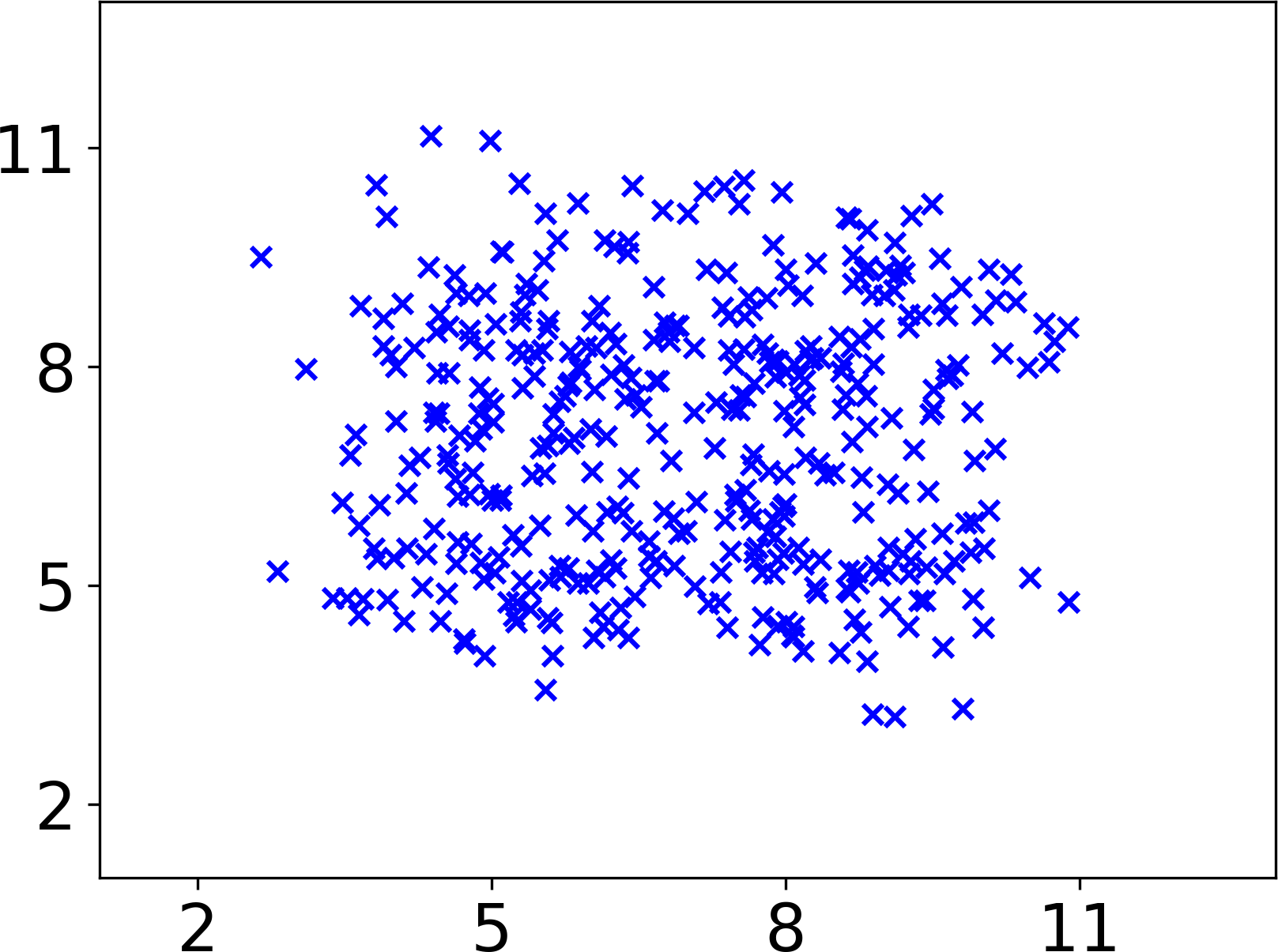}
        \label{fig_proximity4}}
    \hfil
    \subfloat[\emph{proximity5}]{\includegraphics[width=0.15\textwidth]{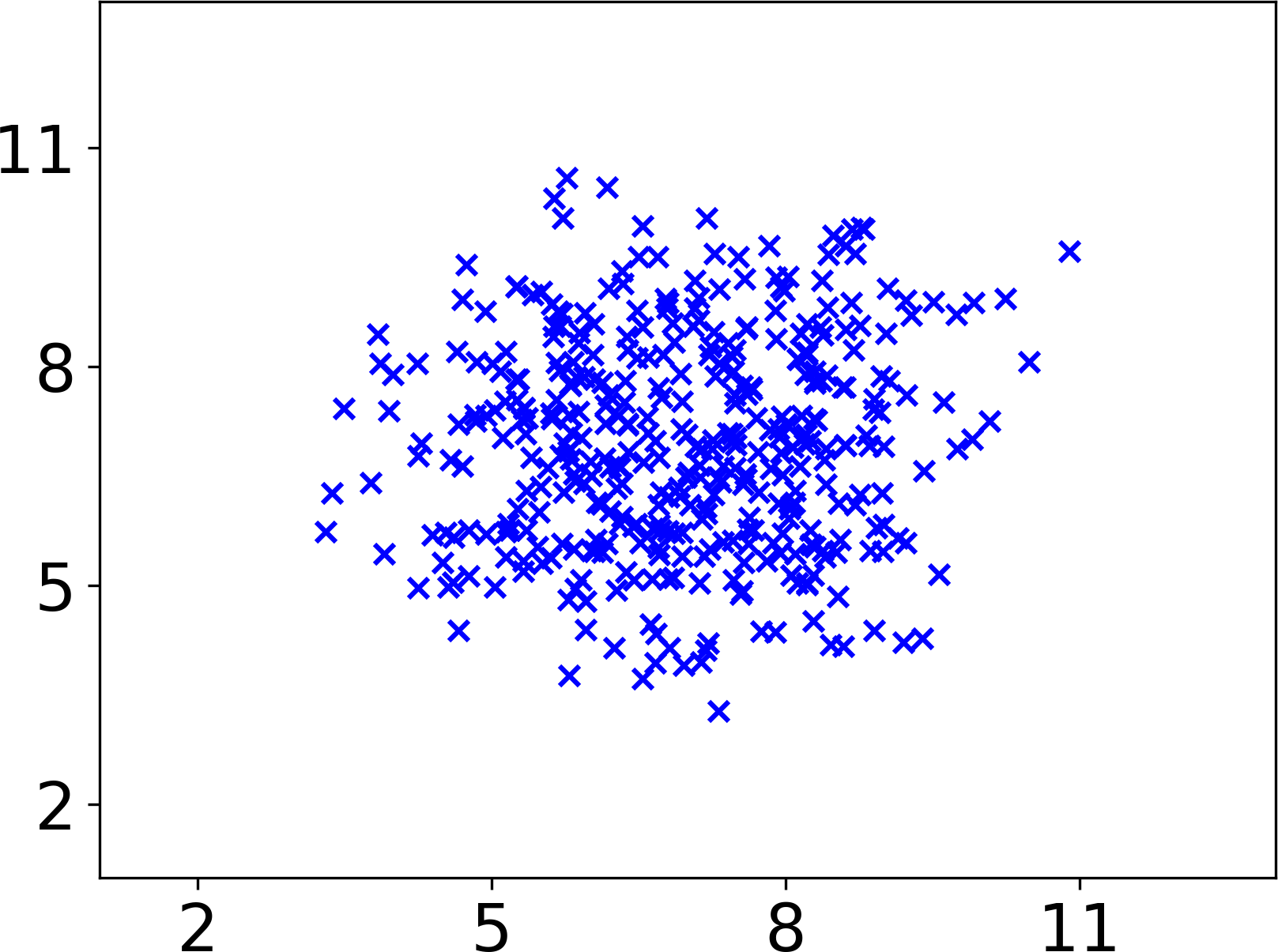}
		\label{fig_proximity5}}
	\hfil
	\caption{The synthetic \emph{proximity }datasets.}
	\label{fig_proximity_datasets}
\end{figure}

\begin{figure}
\centering
	\subfloat[\emph{spread1}]{\includegraphics[width=0.15\textwidth]{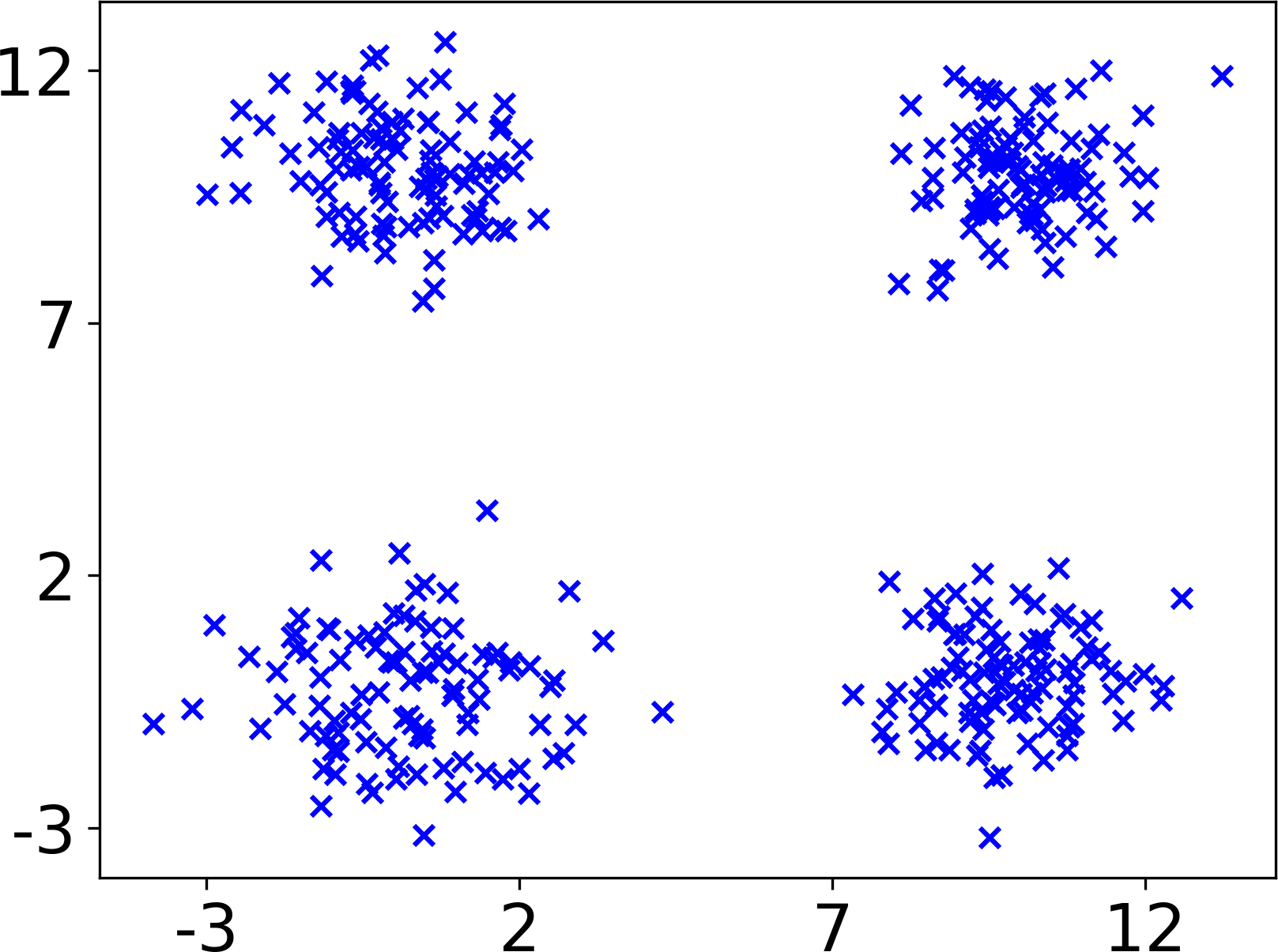}
		\label{fig_spread1}}
	\hfil
    \subfloat[\emph{spread2}]{\includegraphics[width=0.15\textwidth]{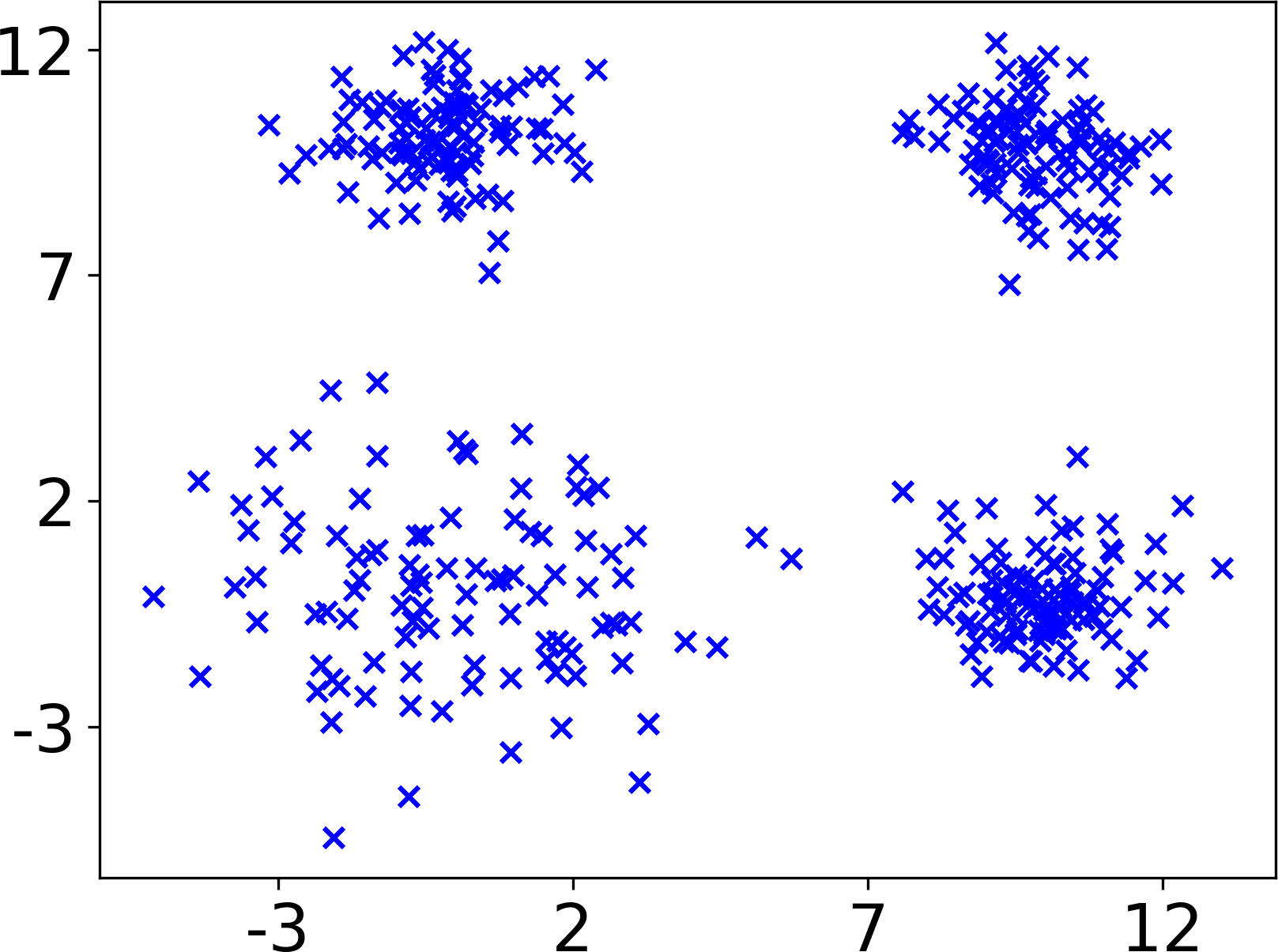}
		\label{fig_spread2}}
	\hfil
	\subfloat[\emph{spread3}]{\includegraphics[width=0.15\textwidth]{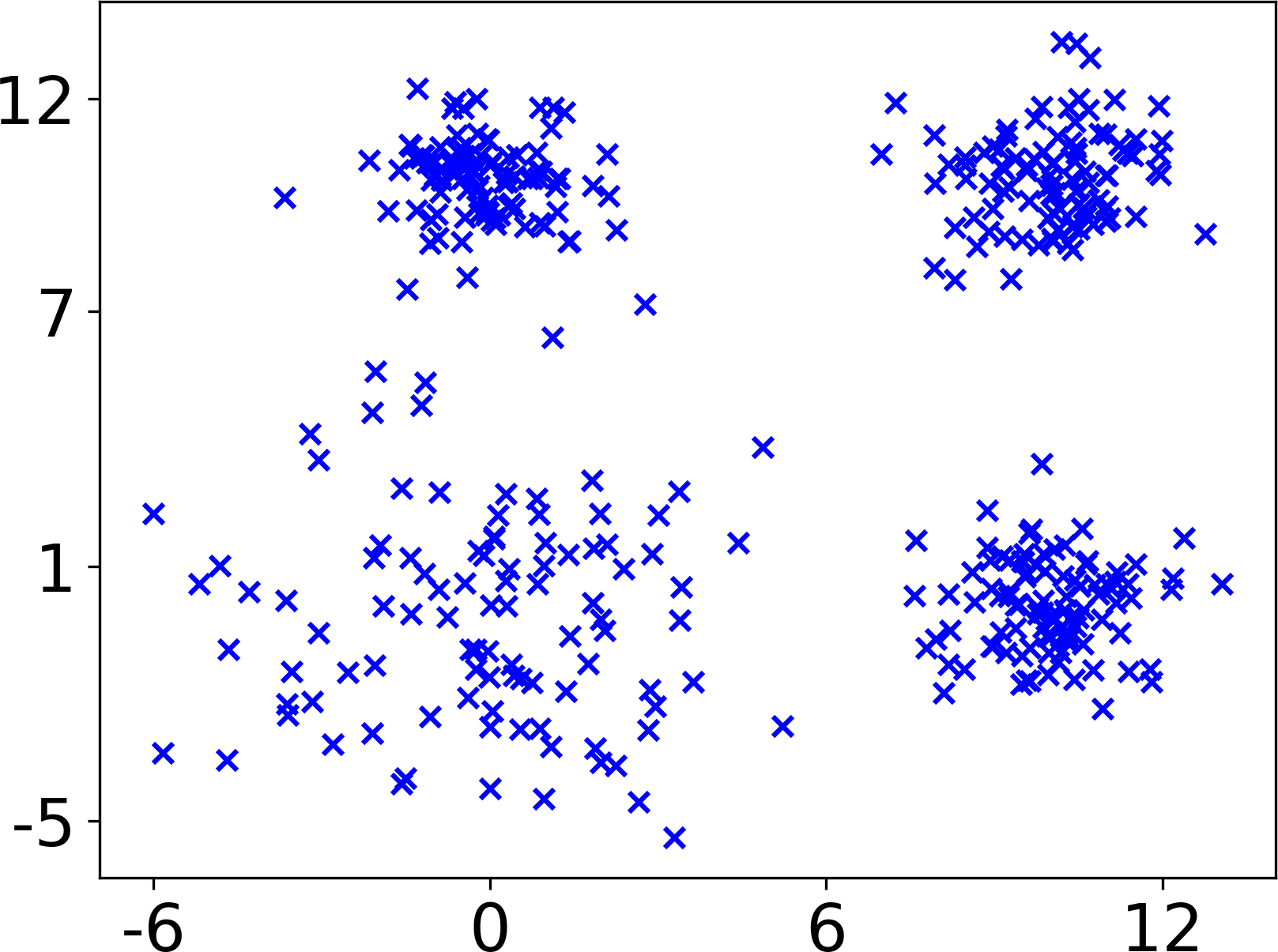}
		\label{fig_spread3}}

    \subfloat[\emph{spread4}]{\includegraphics[width=0.15\textwidth]{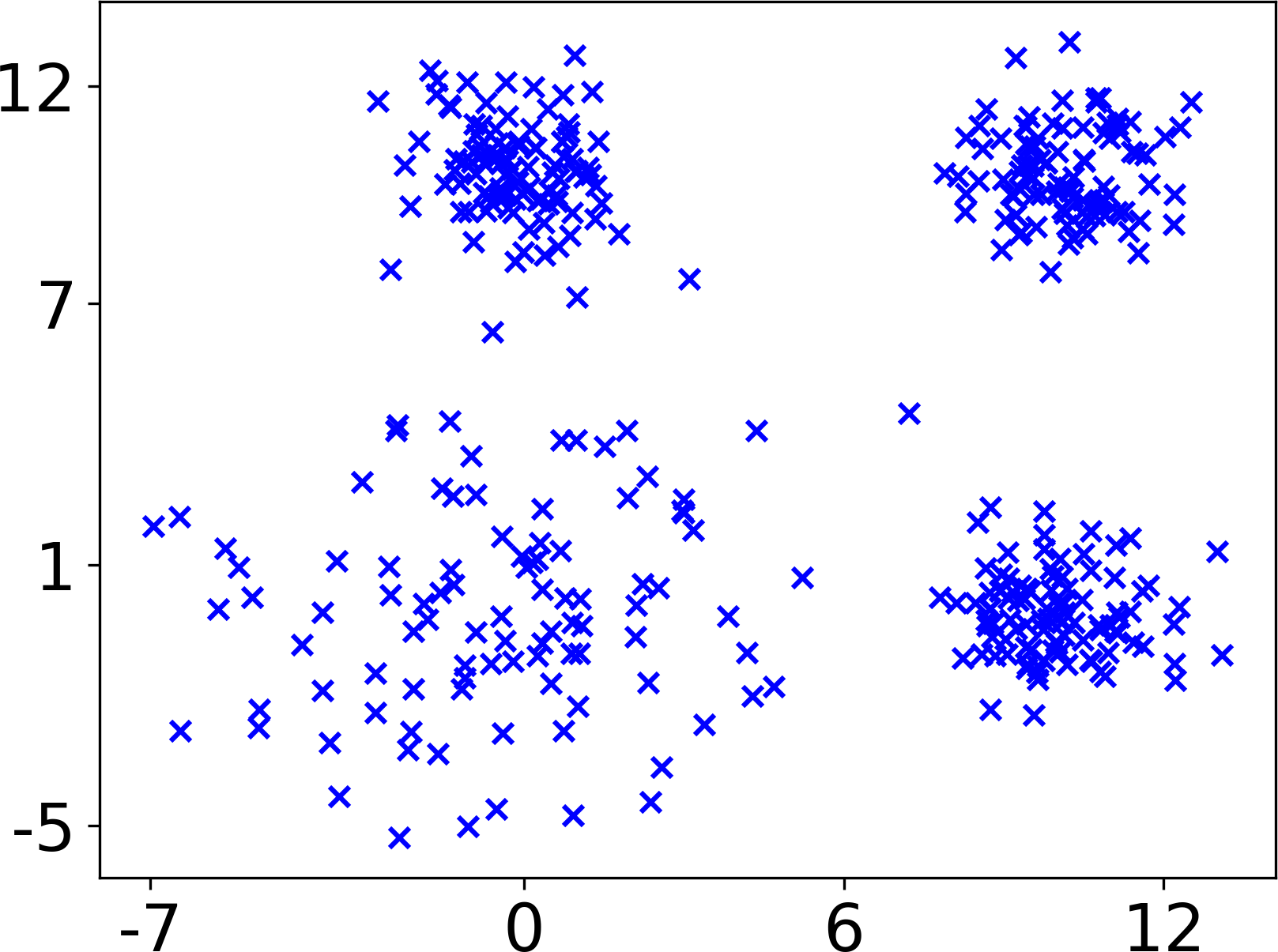}
    	\label{fig_spread4}}
    \hfil
	\subfloat[\emph{spread5}]{\includegraphics[width=0.15\textwidth]{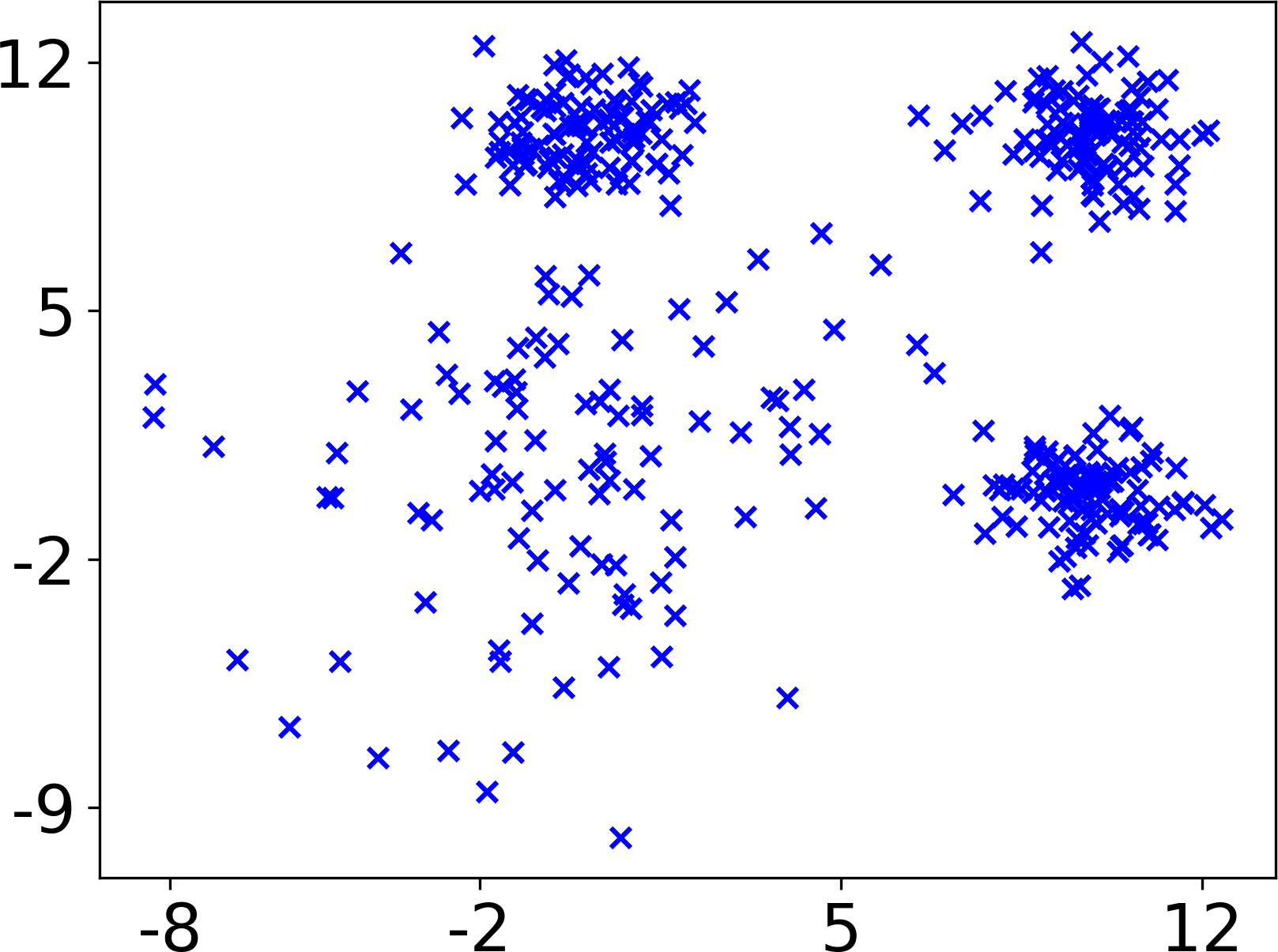}
		\label{fig_spread5}}
	\hfil
	\caption{The synthetic \emph{spread} datasets.}
	\label{fig_spread_datasets}
\end{figure}

We created 10 synthetic datasets to compare the performances of all fuzzy clustering methods. The \emph{proximity1-5} datasets test the clustering performance when clusters are drawn closer together, as shown in Fig. \ref{fig_proximity_datasets}. Datasets \emph{spread1-5}, on the other hand, test the performance when the spread of one cluster is progressively increased, as illustrated in Fig. \ref{fig_spread_datasets}. Table \ref{tab_data_gen} contains all information necessary to generate the synthetic \emph{proximity} and \emph{spread} datasets. We also use $20$ challenging datasets from the MOCK collection\footnote{Available at \url{http://personalpages.manchester.ac.uk/mbs/julia.handl/generators.html} (last accessed \today).} \cite{handl07}, having either $2$ or $10$ dimensions. The full specifications of the synthetic datasets are present in Table \ref{tab_synth_datasets}.

\begin{table}
    \centering
    {
    \footnotesize
    \caption{\label{tab_data_gen}Information to generate the synthetic datasets: the number of clusters $c$, dimension of the data $dim$, number of data points in each cluster $N_i$, the multidimensional Normal distribution $N(\mu, \sigma)$, where $\mu$ is the vector of cluster centers and $\sigma$ is the vector of the standard deviations in each dimension.}
    \begin{tabular}{l c c c p{5cm}}
        \hline
        Dataset & $c$ & $dim$ & $N_i$ & Clusters \\
        \hline
        \emph{proximity1} & 4 & 2 & 100 & $N([4,4],[1,1])$, $N([4,10],[1,1])$, $N([10,4],[1,1])$, $N([10,10],[1,1])$ \\
        \emph{proximity2} & 4 & 2 & 100 & $N([4.5,4.5],[1,1])$, $N([4.5,9.5],[1,1])$, $N([9.5,4.5],[1,1])$, $N([9.5,9.5],[1,1])$ \\
        \emph{proximity3} & 4 & 2 & 100 & $N([5,5],[1,1])$, $N([5,9],[1,1])$, $N([9,5],[1,1])$, $N([9,9],[1,1])$ \\
        \emph{proximity3} & 4 & 2 & 100 & $N([5.5,5.5],[1,1])$, $N([5.5,8.5],[1,1])$, $N([8.5,5.5],[1,1])$, $N([8.5,8.5],[1,1])$ \\
        \emph{proximity5} & 4 & 2 & 100 & $N([6,6],[1,1])$, $N([6,8],[1,1])$, $N([8,6],[1,1])$, $N([8,8],[1,1])$ \\
        \emph{spread1} & 4 & 2 & 100 & $N([0,0],[1,1])$, $N([0,10],[1,1])$, $N([10,0],[1,1])$, $N([10,10],[1,1])$ \\
        \emph{spread2} & 4 & 2 & 100 & $N([0,0],[1.5,1.5])$, $N([0,10],[1,1])$, $N([10,0],[1,1])$, $N([10,10],[1,1])$ \\
        \emph{spread3} & 4 & 2 & 100 & $N([0,0],[2,2])$, $N([0,10],[1,1])$, $N([10,0],[1,1])$, $N([10,10],[1,1])$ \\
        \emph{spread4} & 4 & 2 & 100 & $N([0,0],[2.5,2.5])$, $N([0,10],[1,1])$, $N([10,0],[1,1])$, $N([10,10],[1,1])$ \\
        \emph{spread5} & 4 & 2 & 100 & $N([0,0],[3,3])$, $N([0,10],[1,1])$, $N([10,0],[1,1])$, $N([10,10],[1,1])$ \\
        \hline
    \end{tabular}
    }
\end{table}

\begin{table}[t]
\centering
{\scriptsize
\caption{\label{tab_synth_datasets}Synthetic Datasets}
\begin{tabular}{l c r }
\hline
Dataset & Number of Clusters & Dataset Size\\
\hline
proximity1 & 4 & (400,2) \\
proximity2 & 4 & (400,2) \\
proximity3 & 4 & (400,2) \\
proximity4 & 4 & (400,2) \\
proximity5 & 4 & (400,2) \\
spread1 & 4 & (400,2) \\
spread2 & 4 & (400,2) \\
spread3 & 4 & (400,2) \\
spread4 & 4 & (400,2) \\
spread5 & 4 & (400,2) \\
2d-4c-no0 & 4 & (1572,2) \\
2d-4c-no1 & 4 & (1623,2) \\
2d-4c-no2 & 4 & (1064,2) \\
2d-4c-no3 & 4 & (1123,2) \\
2d-4c-no4 & 4 & (863,2) \\
2d-4c-no5 & 4 & (1638,2) \\
2d-4c-no6 & 4 & (1670,2) \\
2d-4c-no7 & 4 & (1028,2) \\
2d-4c-no8 & 4 & (1078,2) \\
2d-4c-no9 & 4 & (876,2) \\
10d-4c-no0 & 10 & (1289,10) \\
10d-4c-no1 & 10 & (958,10) \\
10d-4c-no2 & 10 & (838,10) \\
10d-4c-no3 & 10 & (1318,10) \\
10d-4c-no4 & 10 & (933,10) \\
10d-4c-no5 & 10 & (1139,10) \\
10d-4c-no6 & 10 & (977,10) \\
10d-4c-no7 & 10 & (1482,10) \\
10d-4c-no8 & 10 & (966,10) \\
10d-4c-no9 & 10 & (1183,10) \\
\hline
\end{tabular}
}
\end{table}

\begin{table}[t]
\centering
{\scriptsize
\caption{\label{tab_mock_results}Comparison of maximum ARI over synthetic datasets}
\begin{tabular}{l p{0.6cm} p{0.6cm} p{0.6cm} p{0.7cm} p{0.7cm} p{0.7cm}}
\hline
Dataset & FCM & MEI & MOGA & MOGA-SVM & ECM-NSGA-II & ECM-MOEA/D \\
\hline
proximity1 & \textbf{1.0000} & \textbf{1.0000} & \textbf{1.0000} & \textbf{1.0000} & \textbf{1.0000} & \textbf{1.0000} \\
proximity2 & 0.9474 & 0.9474 & 0.9474 & 0.9474  & \textbf{0.9476} & \textbf{0.9476}\\
proximity3 & 0.9050 & 0.9087 & 0.9049 & 0.9049 & 0.9087 & \textbf{0.9090} \\
proximity4 & 0.7376 & 0.7426 & 0.7430 & 0.7428 & 0.7459 & \textbf{0.7465} \\
proximity5 & 0.3475 & 0.3455 & 0.3504 & 0.3724 & 0.3682 & \textbf{0.3740} \\
spread1 & \textbf{1.0000} & \textbf{1.0000} & \textbf{1.0000} & \textbf{1.0000} & \textbf{1.0000} & \textbf{1.0000} \\
spread2 & 0.9867 & 0.9867 & 0.9867 & 0.9867 & \textbf{1.0000} & \textbf{1.0000} \\
spread3 & 0.9543 & 0.9543 & 0.9543 & 0.9543 & 0.9671 & \textbf{0.9735} \\
spread4 & 0.9479 & 0.9542 & 0.9479 & 0.9479 & \textbf{0.9670} & 0.9543 \\
spread5 & 0.8876 & 0.8818 & 0.8876 & 0.8876 & \textbf{0.9414} & 0.8876 \\
2d-4c-no0 & 0.8834 & 0.8691 & 0.8834 & \textbf{0.8851} & 0.8823 & 0.8727 \\
2d-4c-no1 & 0.8819 & 0.7836 & 0.7798 & 0.7870 & \textbf{0.8874} & 0.8517 \\
2d-4c-no2 & 0.8885 & 0.9143 & 0.8885 & 0.8917 & \textbf{0.9570} & 0.8454 \\
2d-4c-no3 & 0.9364 & 0.8696 & 0.9378 & 0.9378 & 0.9436 & \textbf{0.9601} \\
2d-4c-no4 & 0.7878 & 0.3571 & 0.7878 & 0.7878 & \textbf{0.8308} & 0.6353 \\
2d-4c-no5 & 0.9300 & 0.8364 & 0.9344 & 0.9363 & \textbf{0.9407} & 0.8700 \\
2d-4c-no6 & 0.9547 & 0.9546 & 0.9565 & 0.9547 & 0.9620 & \textbf{0.9759} \\
2d-4c-no7 & \textbf{0.9117} & 0.8275 & 0.7081 & 0.6761 & 0.8991 & 0.8659 \\
2d-4c-no8 & 0.9804 & 0.9152 & 0.9835 & \textbf{0.9838} & 0.9825 & 0.9179 \\
2d-4c-no9 & 0.8883 & 0.8841 & 0.8884 & 0.8884 & \textbf{0.9008} & 0.8779 \\
10d-4c-no0 & 0.9083 & 0.9473 & 0.7687 & 0.7687 & \textbf{0.9707} & 0.9432 \\
10d-4c-no1 & 0.9797 & 0.9722 & 0.9926 & 0.9906 & \textbf{0.9944} & 0.9917 \\
10d-4c-no2 & 0.8844 & 0.9335 & 0.9307 & 0.8844 & \textbf{0.9776} & 0.9662 \\
10d-4c-no3 & 0.9023 & 0.8790 & 0.8875 & 0.8875 & \textbf{0.9847} & 0.8640 \\
10d-4c-no4 & 0.9043 & 0.9147 & 0.8122 & 0.8122 & \textbf{0.9850} & 0.9655 \\
10d-4c-no5 & 0.7569 & \textbf{0.8749} & 0.7173 & 0.7304 & 0.8229 & 0.8494\\
10d-4c-no6 & 0.8727 & 0.8990 & 0.9670 & 0.9761  & \textbf{0.9976} & 0.8130\\
10d-4c-no7 & 0.9940 & 0.9940 & 0.9940 & 0.9940 & 0.9960 & \textbf{0.998} \\
10d-4c-no8 & 0.9603 & 0.9334 & 0.9660 & 0.9602 & \textbf{0.9710} & 0.9676 \\
10d-4c-no9 & 0.9577 & 0.9836 & 0.9863 & 0.9870 & \textbf{0.9964} & 0.9895 \\
\hline
Avg. Rank & 4.0833 & 4.4833 & 3.9167 & 3.8167 & \textbf{1.7500} & 2.9500 \\
ECM-NSGA-II & $H_1$ & $H_1$ & $H_1$ & $H_1$ & - & $H_1$ \\
ECM-MOEA/D & $H_0$ & $H_1$ & $H_0$ & $H_0$ & $H_1$ & - \\
\hline
\end{tabular}
}
\end{table}

For each dataset, we normalize the data so that each feature is scaled within the range $[-1,1]$. We run both FCM and MEI 50 times for a maximum of $5000$ iterations, with an error tolerance of $10^{-16}$. We run ECM-NSGA-II, ECM-MOEA/D, MOGA, and MOGA-SVM with a population size of $50$ for $5000$ fitness evaluations.

For FCM, we set $m=2$, as per convention. MEI, ECM-NSGA-II and ECM-MOEA/D use the admissible error radius parameter $\sigma$. In \cite{dsr95}, $\sigma$ was set to $0.7$. However, we have not found any general recommendations regarding the choice $\sigma$. Moreover, we have observed that the centers tend to converge if $\sigma$ is large when compared to the spread of the data. We therefore decided against using a constant value of $\sigma$ across all datasets. Instead, for each dataset we set $\sigma$ as the standard deviation over the squared Euclidean distances between the data points and the mean of the dataset.

In the presence of the original cluster labels, we can use the Adjusted Rand Index (ARI) \cite{Hubert1985} (calculated by assigning each data point to the cluster to which it has the maximum membership) to compare the performance of the contending methods. For each MOO method, the clustering having maximum ARI is selected for comparison. For FCM and MEI, we report the maximum ARI over the 50 runs. The results are detailed in Table \ref{tab_mock_results}. The outcomes of the  sign-rank test conducted with ECM-NSGA-II and ECM-MOEA/D as the control are also listed, alongside the average ranks.

\begin{figure}[b]
\centering
\includegraphics[width=0.48\textwidth]{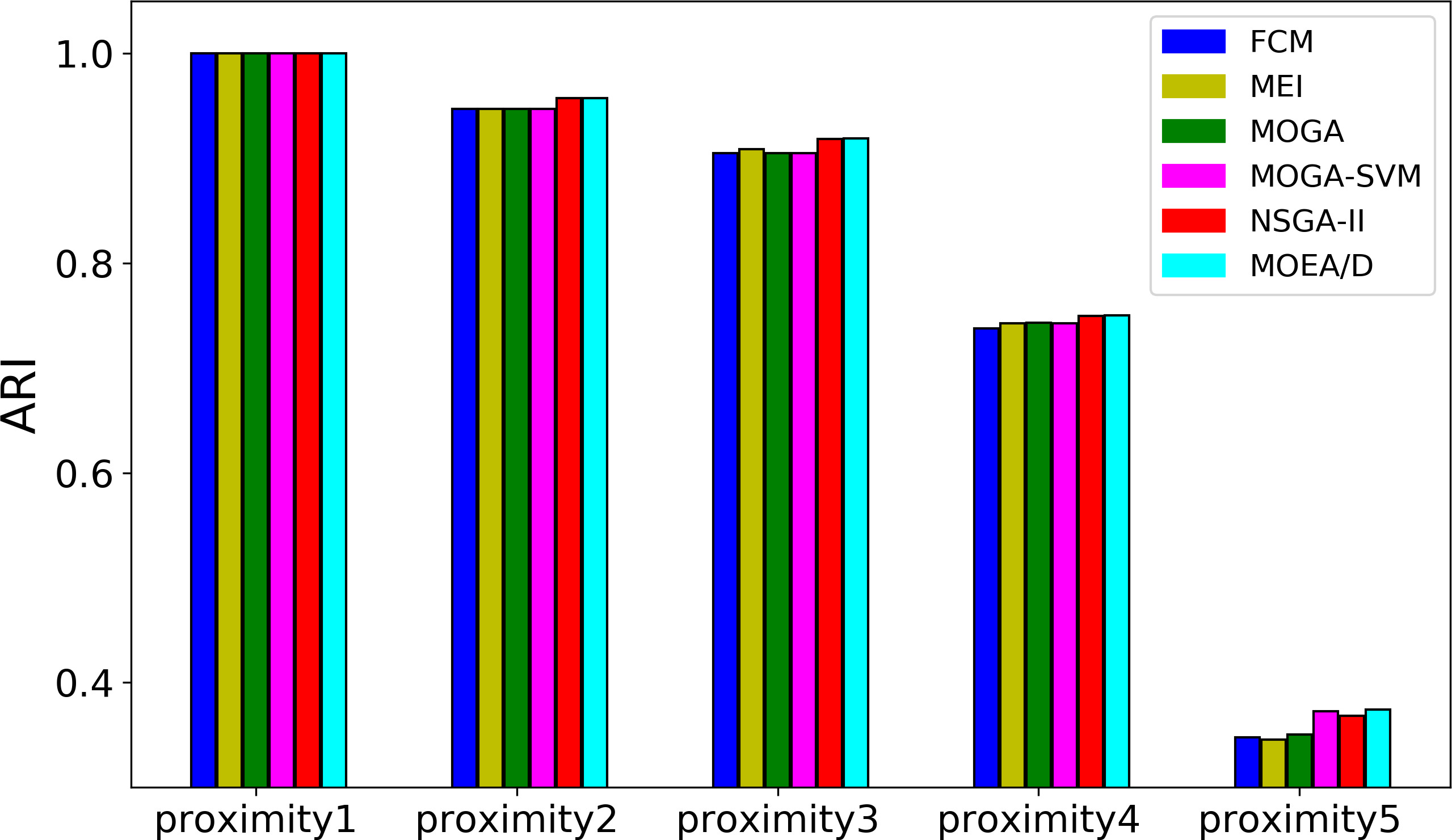}
\caption{Comparison of the maximum ARI achieved on the \emph{proximity} datasets}
\label{fig_proximity_ari}
\end{figure}

\begin{figure}[b]
\centering
\includegraphics[width=0.48\textwidth]{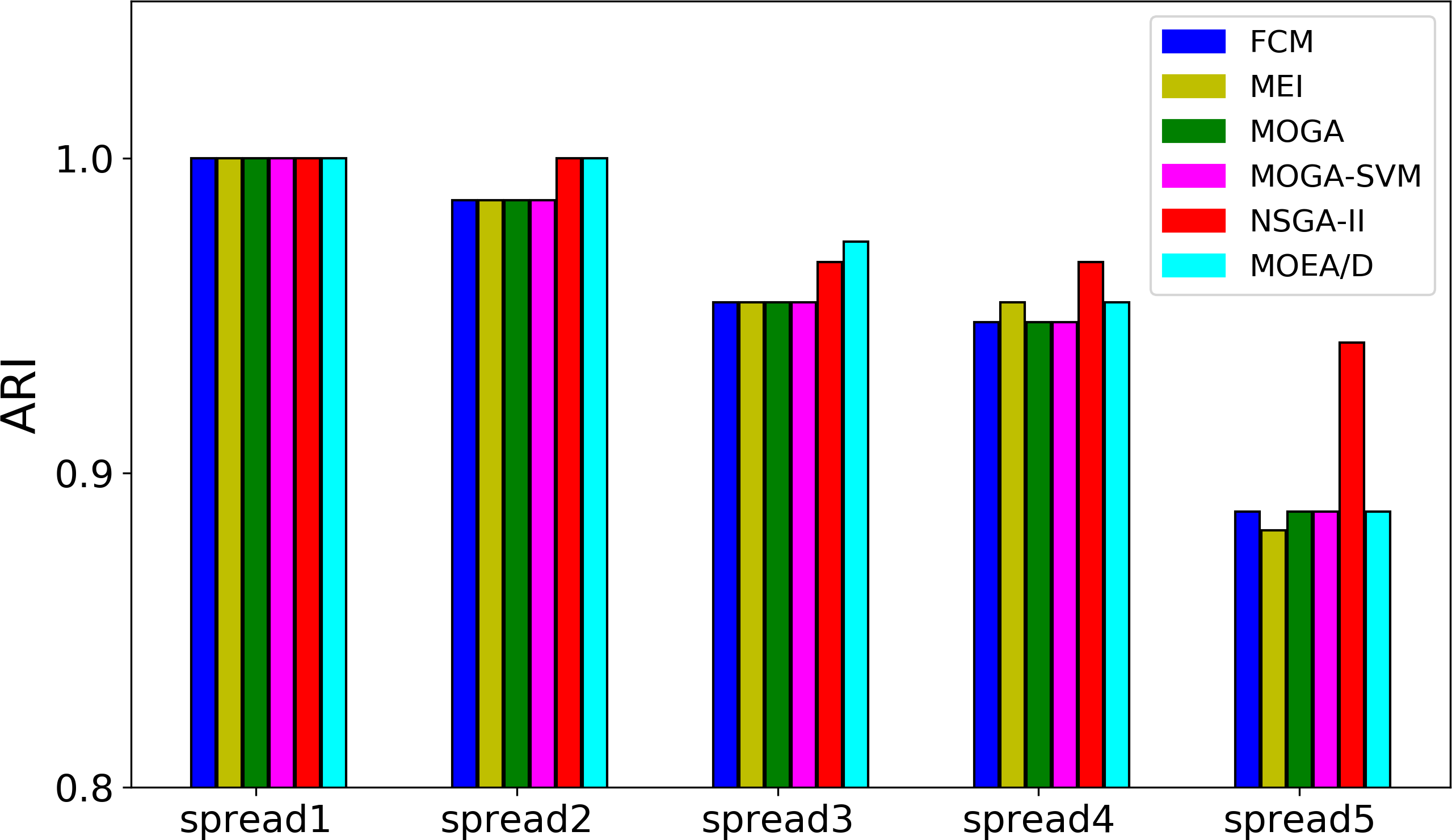}
\caption{Comparison of the maximum ARI achieved on the \emph{spread} datasets}
\label{fig_spread_ari}
\end{figure}

We observe that overall ECM-NSGA-II achieves the best ARI, which is also supported by the average rank as well as the sign-rank test. Even on the ten-dimensional data, ECM-NSGA-II generally produces higher ARI compared to the other methods. On the five \emph{proximity} datasets illustrated in Fig. \ref{fig_proximity_datasets}, ECM-MOEA/D performs the best as shown in Fig. \ref{fig_proximity_ari}. On the other hand, for the five \emph{spread} datasets illustrated in Fig. \ref{fig_spread_datasets}, ECM-NSGA-II generally achieves higher ARI values as shown in Fig. \ref{fig_spread_ari}. This suggests that ECM-MOEA/D is more resilient to closeness between clusters while the NSGA-II variant exhibits greater resilience to disparate spreads between the clusters. Further experiments on the tuning of the parameters of NSGA-II and MOEA/D are present in Section 2 of the supplementary document.

% ~~~~~~~~~~~~~~~~~~~~~~~~~~~~~ %

\subsection{Real Datasets}
\label{sec_real_datasets}

We compare the performance of the five methods on fifteen real datasets shown in Table \ref{tab_real}. Twelve of the datasets are from the UCI Machine Learning Repository \cite{Lichman:2013}. Three of the datasets are high-dimensional gene expression data sets, available at \url{http://www.stat.cmu.edu/~jiashun/Research/software/GenomicsData/}. We undertake the preprocessing of the datasets as per the discussion in Section \ref{sec_mock_datasets} and also retain the same parameter settings for the contending methods. We report the maximum ARI achieved by each method in Table \ref{tab_real_results}, along with average ranks and the results of the Wilcoxon sign-rank test\footnote{$H_1$ : Significantly different from the control.\newline $H_0$: Statistically comparable to the control.}. We observe that ECM-NSGA-II and ECM-MOEA/D respectively achieve the best ARI for nine and six out of the fifteen datasets. The rest of the methods do not achieve the best ARI for any of the datasets. This suggests the efficacy of the ECM formulation in general and ECM-NSGA-II in particular.

\begin{table}
\centering
{\scriptsize
\caption{\label{tab_real}Real datasets}
\begin{tabular}{l l p{1.2cm}}
\hline
Dataset & Dataset Size & Number of clusters\\
\hline
Balance Scale (B. Scale) & (625,4) & 3 \\
Breast Tissue (B. Tissue) & (106,9) & 6 \\
Breast Cancer Wisconsin (wdbc) & (683,9) & 2 \\
banknote authentication (banknote) & (1372,4) & 2\\
Echocardiogram (echo) & (106,9) & 2 \\
Ecoli & (336,7) & 8 \\
Iris & (150,4) & 3 \\
magic & (19020,10) & 2 \\
seeds & (210,7) & 3 \\
sonar & (208,60) & 2 \\
User Knowledge Modeling (ukm) & (258,5) & 4 \\
wine & (178,13) & 4 \\
colon cancer & (62,2000) & 2 \\
lung cancer & (181,12533) & 2 \\
prostate cancer & (102,6033) & 2 \\
\hline
\end{tabular}
}
\end{table}

\begin{table}
\centering
{\scriptsize
\caption{\label{tab_real_results}Comparison of maximum ARI over real datasets}
\begin{tabular}{p{1.59cm} p{0.6cm} p{0.6cm} p{0.8cm} p{0.7cm} p{0.95cm} p{0.8cm}}
\hline
Dataset & FCM & MEI & MOGA & MOGA-SVM & ECM-NSGA-II & ECM-MOEA/D \\
\hline
B. scale & 0.2800 & 0.1600 & 0.3320 & 0.2865 & 0.3020 & \textbf{0.3880} \\
B. Tissue & 0.3100 & 0.3360 & 0.3095 & 0.3432 & \textbf{0.3880} & 0.2627 \\
wdbc & 0.8300 & 0.8520 & 0.8300 & 0.8300 & \textbf{0.8800} & 0.8464 \\
banknote & 0.0450 & 0.0220 & 0.1050 & 0.1029 & 0.3450 & \textbf{0.4072} \\
echo & 0.0854 & 0.0670 & 0.0854 & 0.0854 & \textbf{0.3787} & 0.3588 \\
Ecoli & 0.3766 & 0.4288 & 0.3864 & 0.4280 & \textbf{0.6600} & 0.6365 \\
Iris & 0.7287 & 0.6898 & 0.7287 & 0.7565 & \textbf{0.8857} & 0.7484 \\
magic & 0.0577 & 0.0281 & 0.0868 & 0.0818 & \textbf{0.1423} & 0.065 \\
seeds & 0.7266 & 0.7048 & 0.7265 & 0.7266 & \textbf{0.8110} & 0.7606 \\
sonar & 0.0064 & 0.0085 & 0.0085 & 0.0253 & 0.0064 & \textbf{0.0733} \\
ukm & 0.1770 & 0.1957 & 0.2354 & 0.2140 & 0.2880 & \textbf{0.3620} \\
wine & 0.8498 & 0.8685 & 0.8649 & 0.8804 & \textbf{0.8975} & 0.8203 \\
colon cancer & 0.0001 & 0.0044 & -0.0064 & 0.0014 & \textbf{0.1632} & -0.0064 \\
lung cancer & 0.0019 & 0.0001 & 0.0433 & 0.1812 & 0.0121 & \textbf{0.3090} \\
prostate cancer & 0.0044 & 0.0001 & 0.0044 & 0.0044 & 0.0288 & \textbf{0.1032} \\
\hline
Avg. Rank & 4.77 & 4.63 & 3.90 & 3.23 & \textbf{1.83} & 2.63 \\
ECM-NSGA-II & $H_1$ & $H_1$ & $H_1$ & $H_1$ & - & $H_0$ \\
ECM-MOEA/D & $H_1$ & $H_1$ & $H_1$ & $H_1$ & $H_0$ & - \\
\hline
\end{tabular}
}
\end{table}

% ~~~~~~~~~~~~~~~~~~~~~~~~~~~~~ %

\subsection{Selection of a Suitable Clustering from the Pareto Set}
\label{sec_tradeoff}

For unsupervised applications, due to the absence of cluster labels, the best solution may be chosen from the Pareto set found by ECM based on some internal \cite{wli15,bez16} or multi-criterion decision making indices \cite{qu_16}. Alternatively, one can select a suitable trade-off clustering by inspecting the Pareto front.

We present a method to select a suitable clustering from the Pareto fronts identified by ECM-NSGA-II and ECM-MOEA/D. One can observe that starting with the extreme Pareto optimal solution having the lowest value of cluster compactness $f_1$, if the points along the Pareto front fall below the line joining the two ends of the front, then there has been a greater increase in entropy $f_2$ compared to the increase in the value of cluster compactness $f_1$. This means that some of the cluster centers have moved slightly closer to each other (evident from a slight increase in cluster compactness $f_1$) but resulting in a large increase in the entropy $f_2$. This is only possible if the clusters in question are truly overlapped because the points in the region of overlap facilitate a large increase in entropy $f_2$. The knee-point from this region of the Pareto front provides an optimal trade-off solution identifying clusters with the appropriate level of overlap.

On the other hand, if the clusterings along the Pareto front were to move above the said line, there would be a larger increase in cluster compactness compared to the increase in entropy. This is only possible if the true clusters are well-separated, and the identified cluster centers have moved closer to each other and away from the true cluster centers. Hence, a deviation above and away from the line joining the end-points indicates that the true clusters are well-separated. In this scenario, the clustering with minimum value of cluster compactness is the best choice.

Based on these insights, we propose the following method for selecting a clustering from the Pareto front.

\begin{itemize}
    \item If the first three points\footnote{Three points are considered to observe the general trend of the front.} do not lie above the line joining the endpoints, traverse the Pareto front till it touches/ crosses the line joining the endpoints. Choose the clustering corresponding to the point lying at maximum distance from the line within traversed region.
    \item Otherwise, choose the clustering corresponding to the minimum value of cluster compactness $f_1$.
\end{itemize}

For example, let us consider the dataset in Figure \ref{fig_fuzziness_across_pf}. We observe from the Pareto front in Figure \ref{fig_c3_pf} that starting from the clustering cI, the Pareto front dips below the line joining point cI and cIV. Therefore, we traverse the Pareto front until it crosses the red line joining the endpoints. Within this region, the clustering cII lies at maximum distance from the red line. Hence using the above method, cII is chosen as the appropriate trade-off clustering. Let us also consider a dataset with three well-separated, equally spaced clusters, which is shown in the first row of Table \ref{tab_more_figs1} along with the obtained Pareto front. As the Pareto front is observed to rise above the red line joining the extremities, the clustering having minimum value of cluster compactness $f_1$ is selected (marked with a red circle). The final cell of the first row of Table \ref{tab_more_figs1} shows the corresponding clustering. Further demonstrations of the effectiveness of this method over a number of synthetic datasets are shown in the rest of Table \ref{tab_more_figs1} and in Table \ref{tab_more_figs2}.

\begin{table}
    \vspace{-1mm}
    \scriptsize
    \caption{The selection of a suitable trade-off clustering across different datasets.}
    \label{tab_more_figs1}
    \begin{tabular}{| p{0.6cm} | p{2.1cm} | p{2.1cm} | p{2.1cm} |}
        \hline
        \vtop{\hbox{\strut Descri-}\hbox{\strut ption}} & Original Data & Pareto Front & Selected Clustering \\
        \hline
        \spheading{3 equally-spaced, well-separated clusters} & \includegraphics[width=0.12\textwidth]{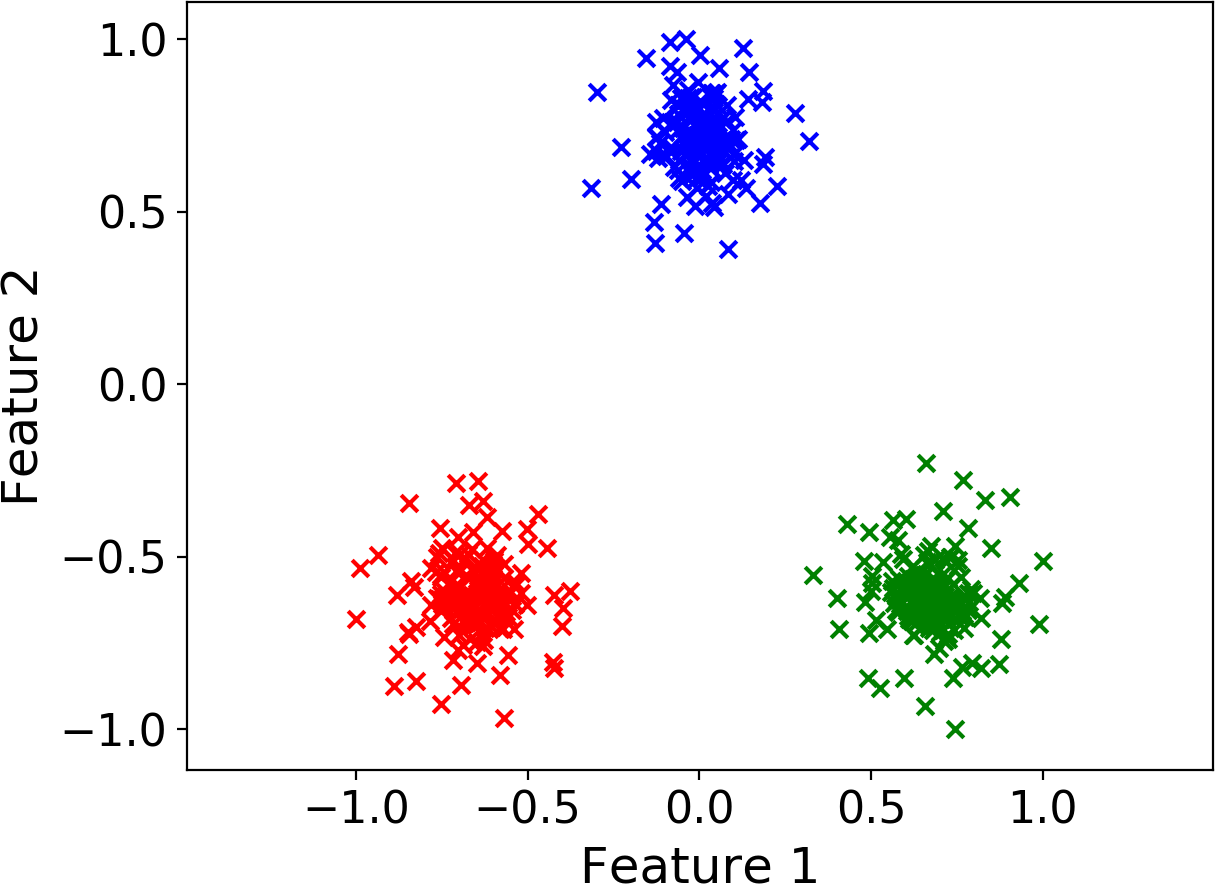} & \includegraphics[width=0.12\textwidth]{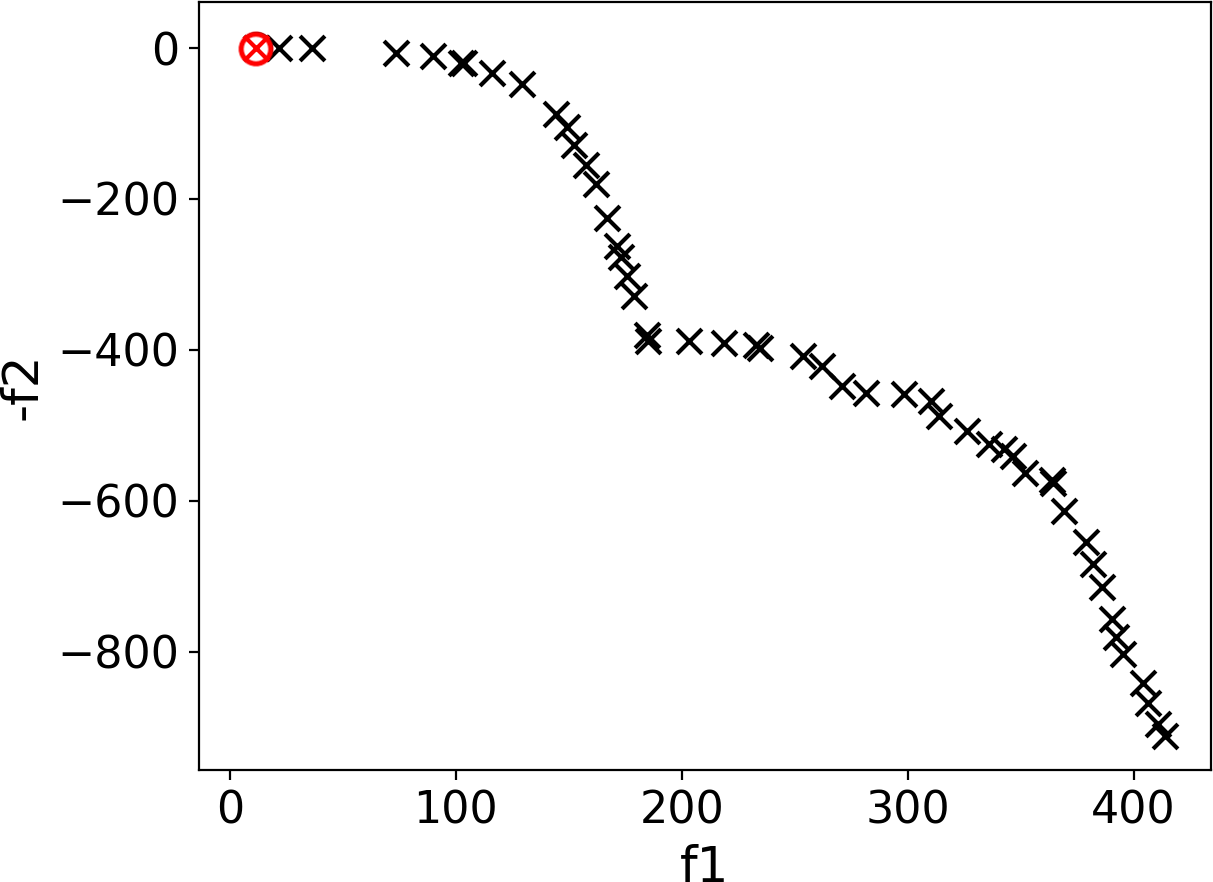} & \includegraphics[width=0.12\textwidth]{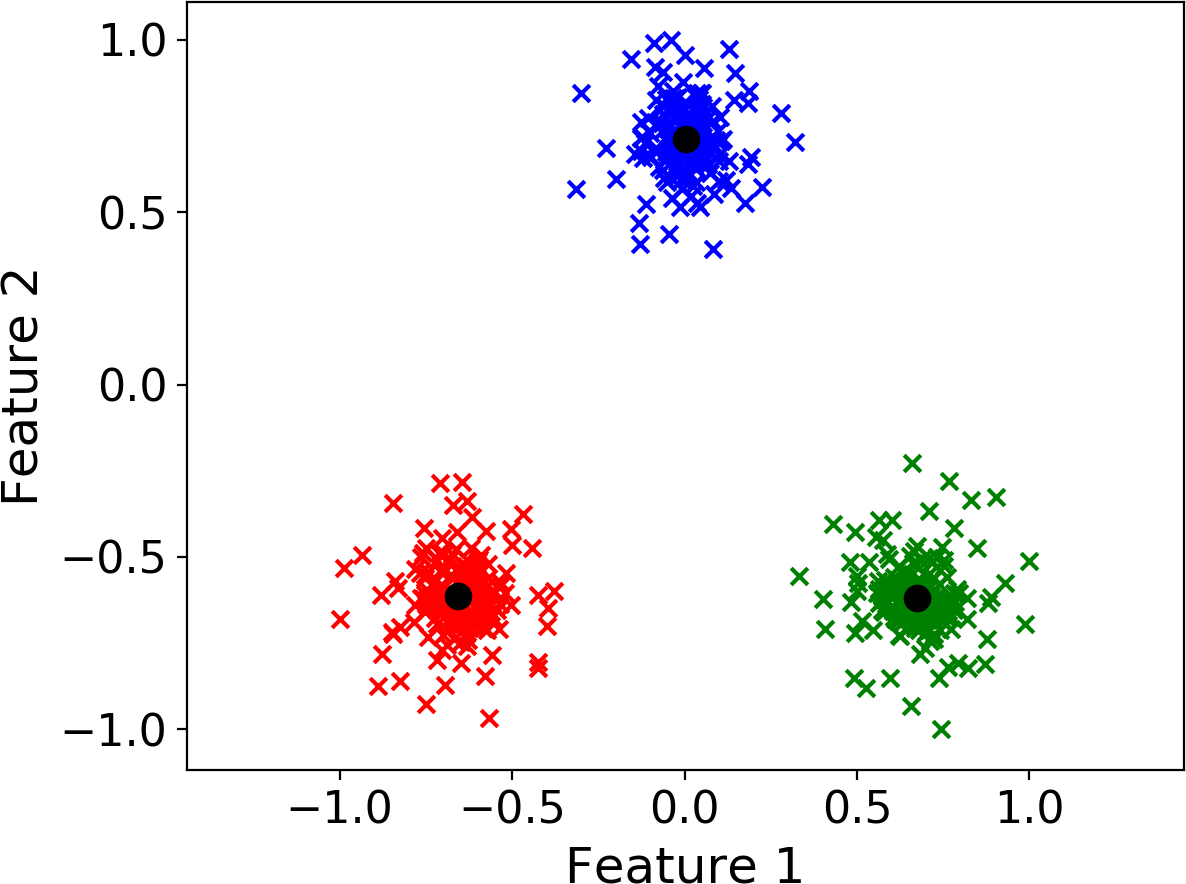} \\
        \hline
        \spheading{3 equally-spaced, slightly-overlapped clusters} & \includegraphics[width=0.12\textwidth]{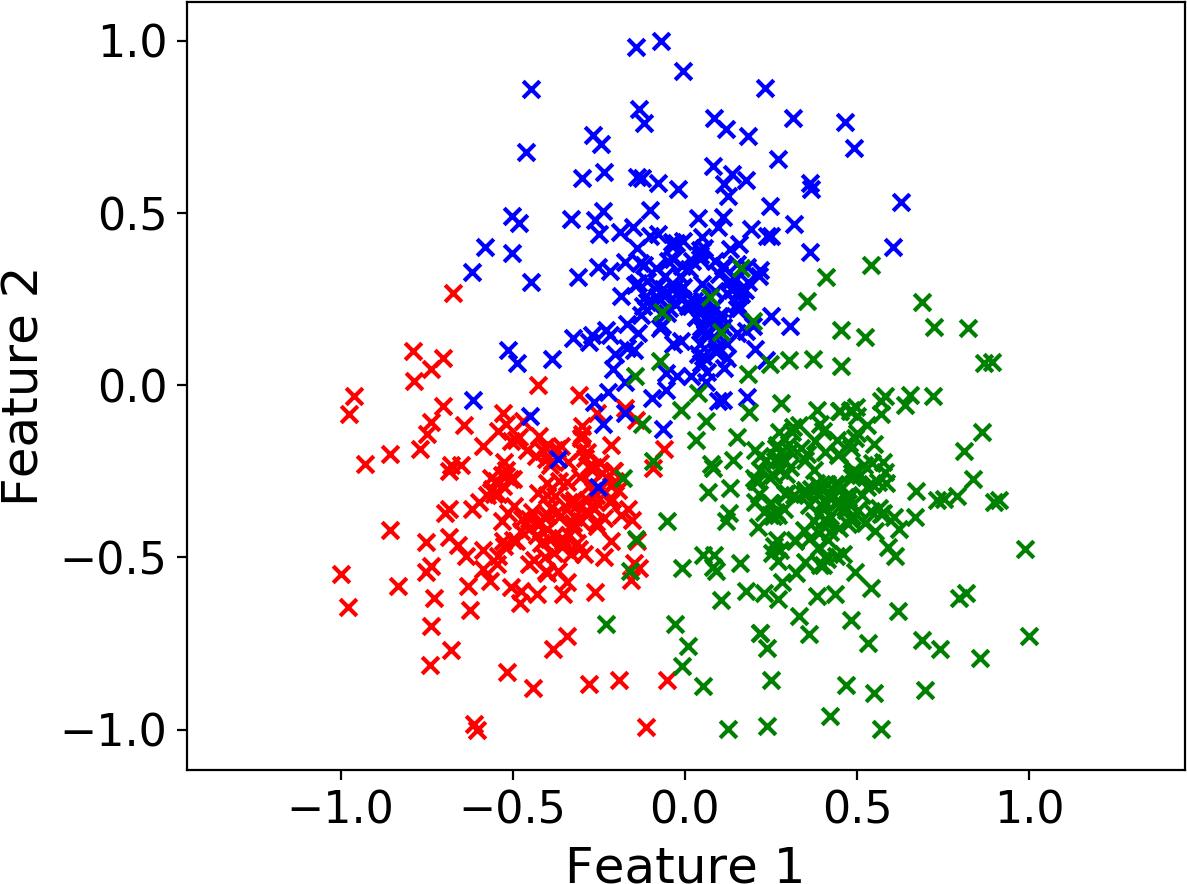} & \includegraphics[width=0.12\textwidth]{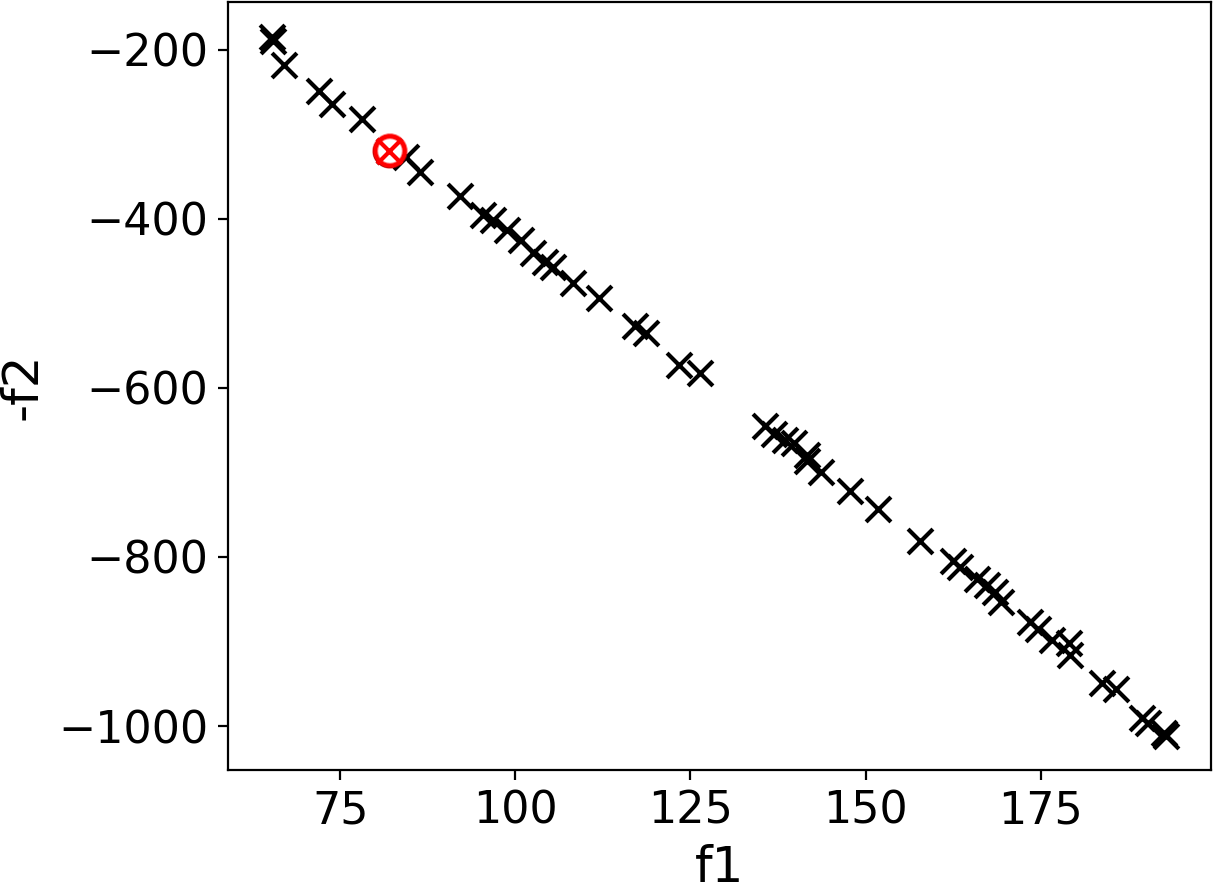} & \includegraphics[width=0.12\textwidth]{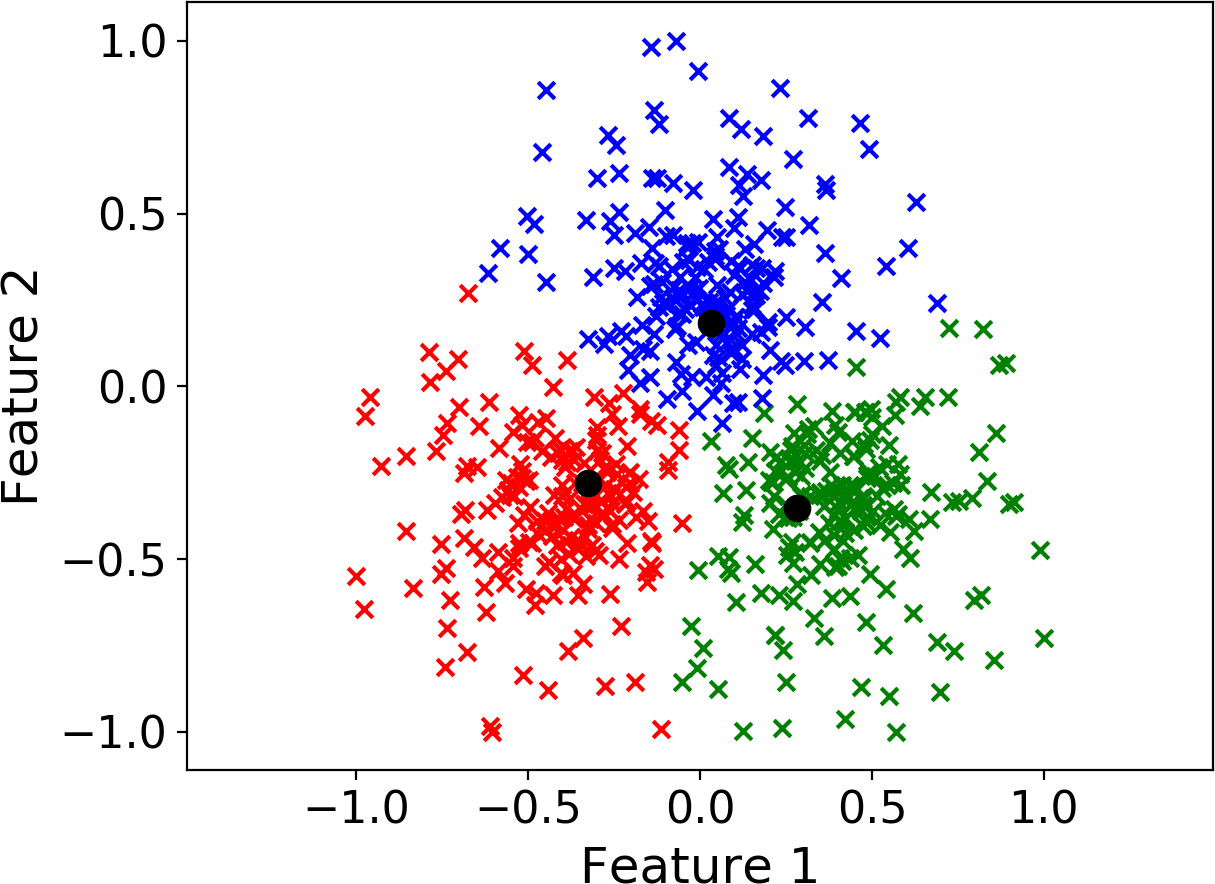} \\
        \hline
        \spheading{3 equally-spaced, highly overlapped clusters} & \includegraphics[width=0.12\textwidth]{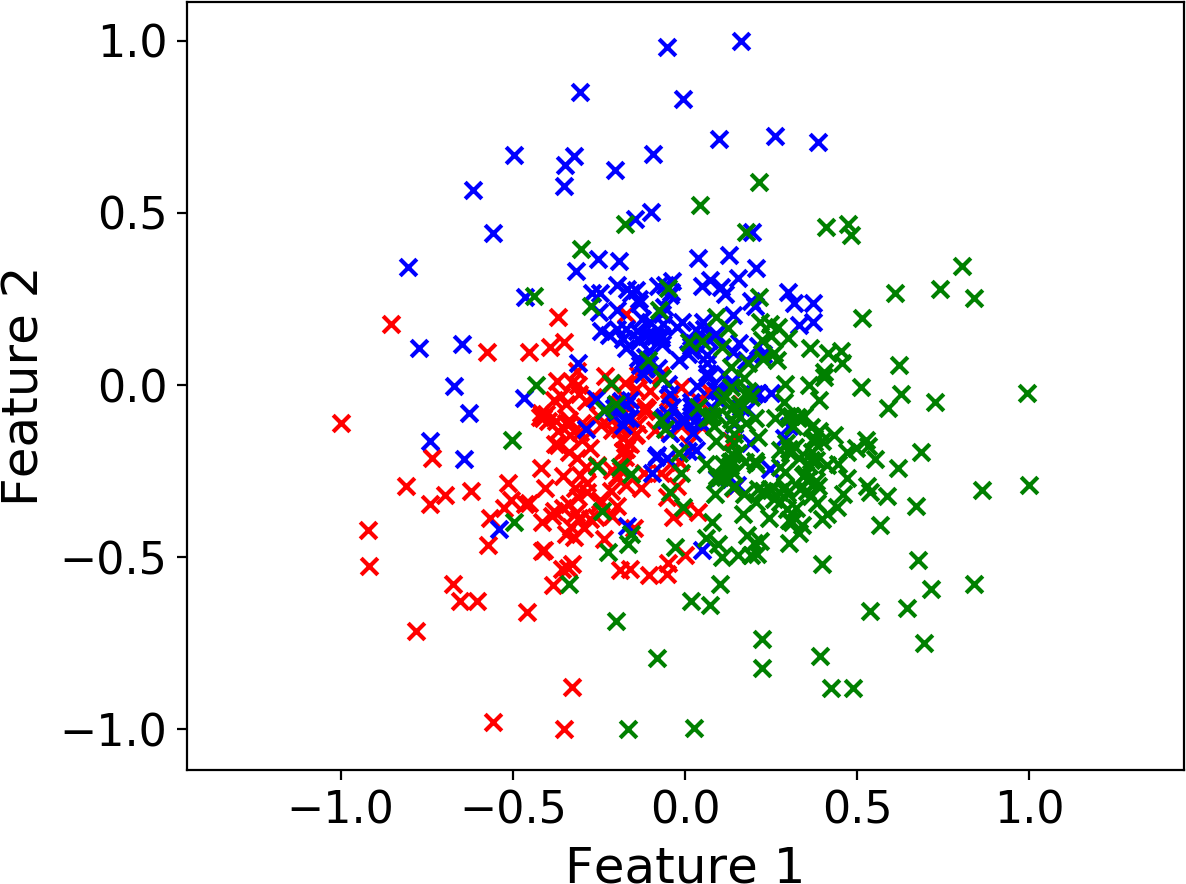} & \includegraphics[width=0.12\textwidth]{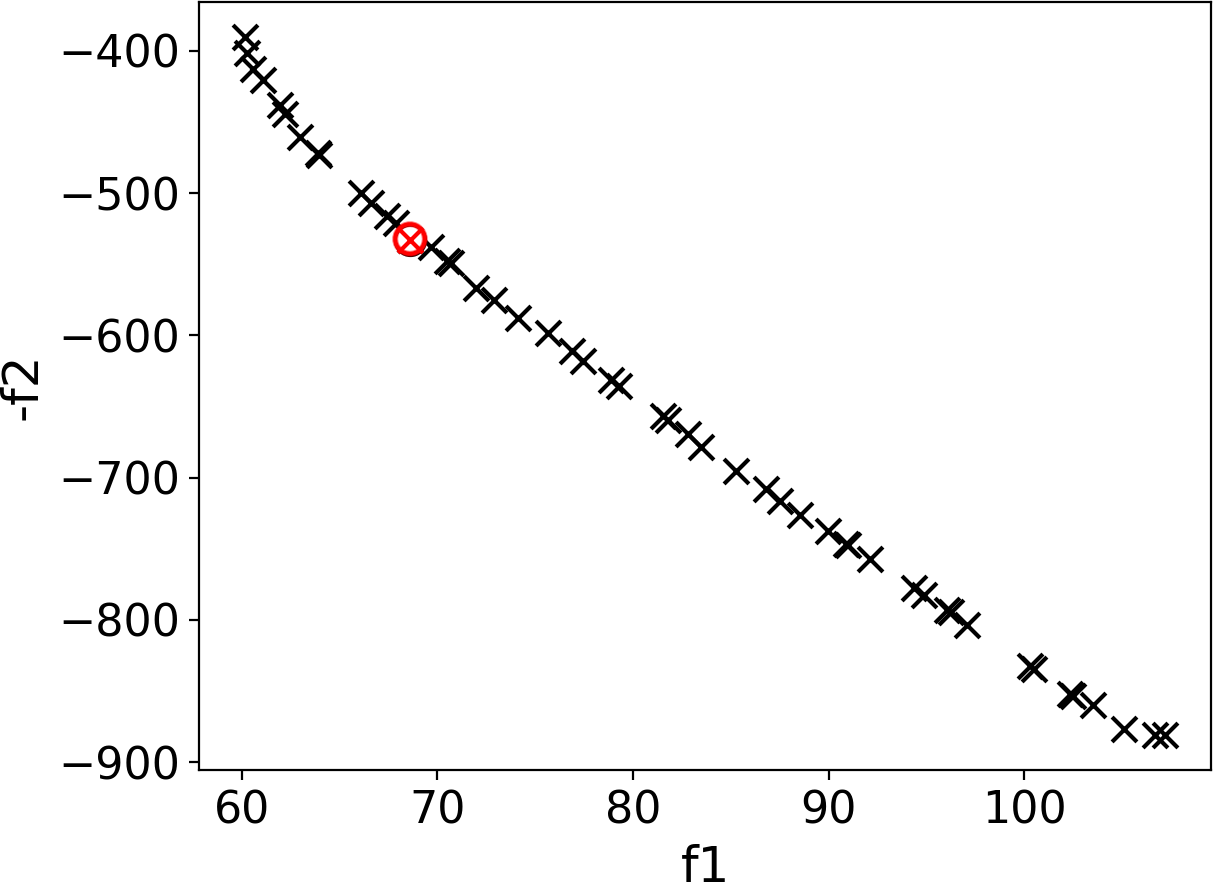}  & \includegraphics[width=0.12\textwidth]{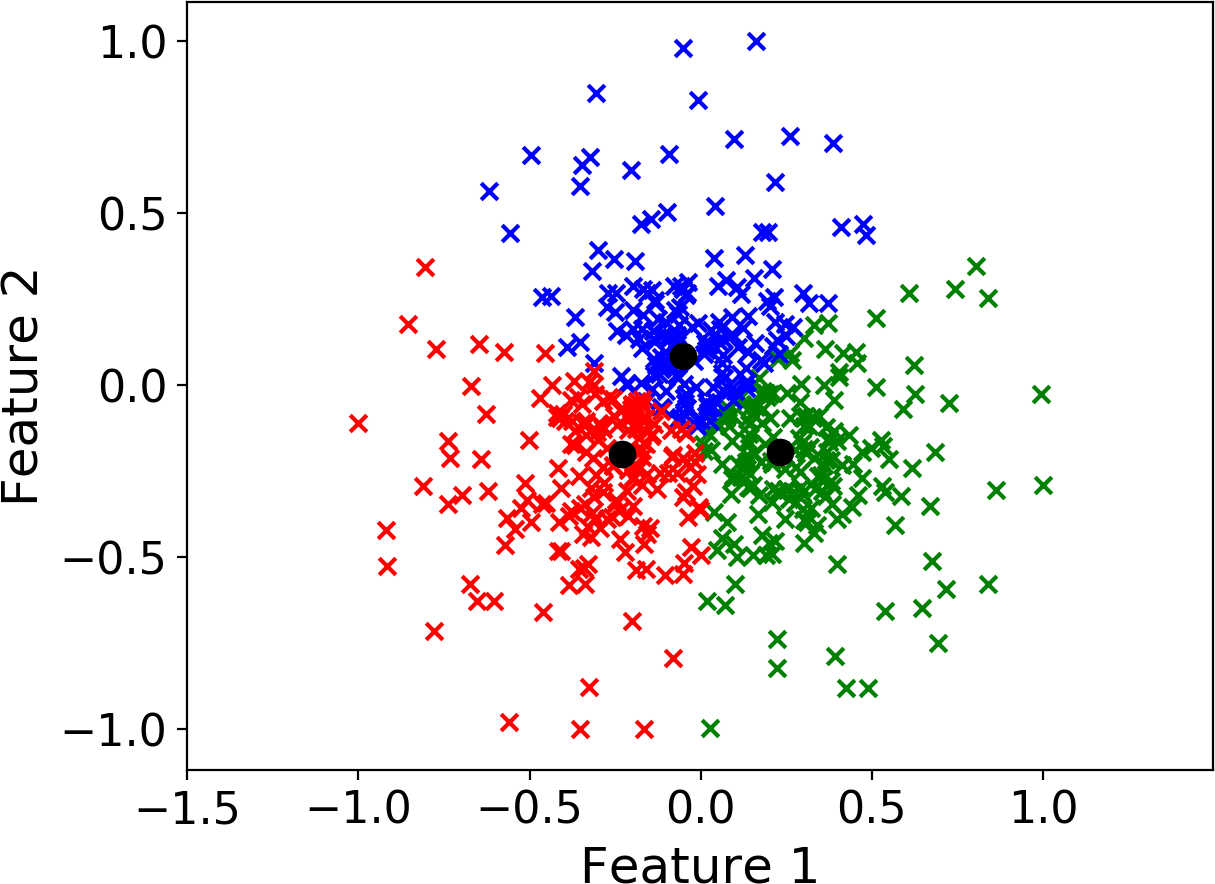} \\
        \hline
        \spheading{3 well-separated clusters} & \includegraphics[width=0.12\textwidth]{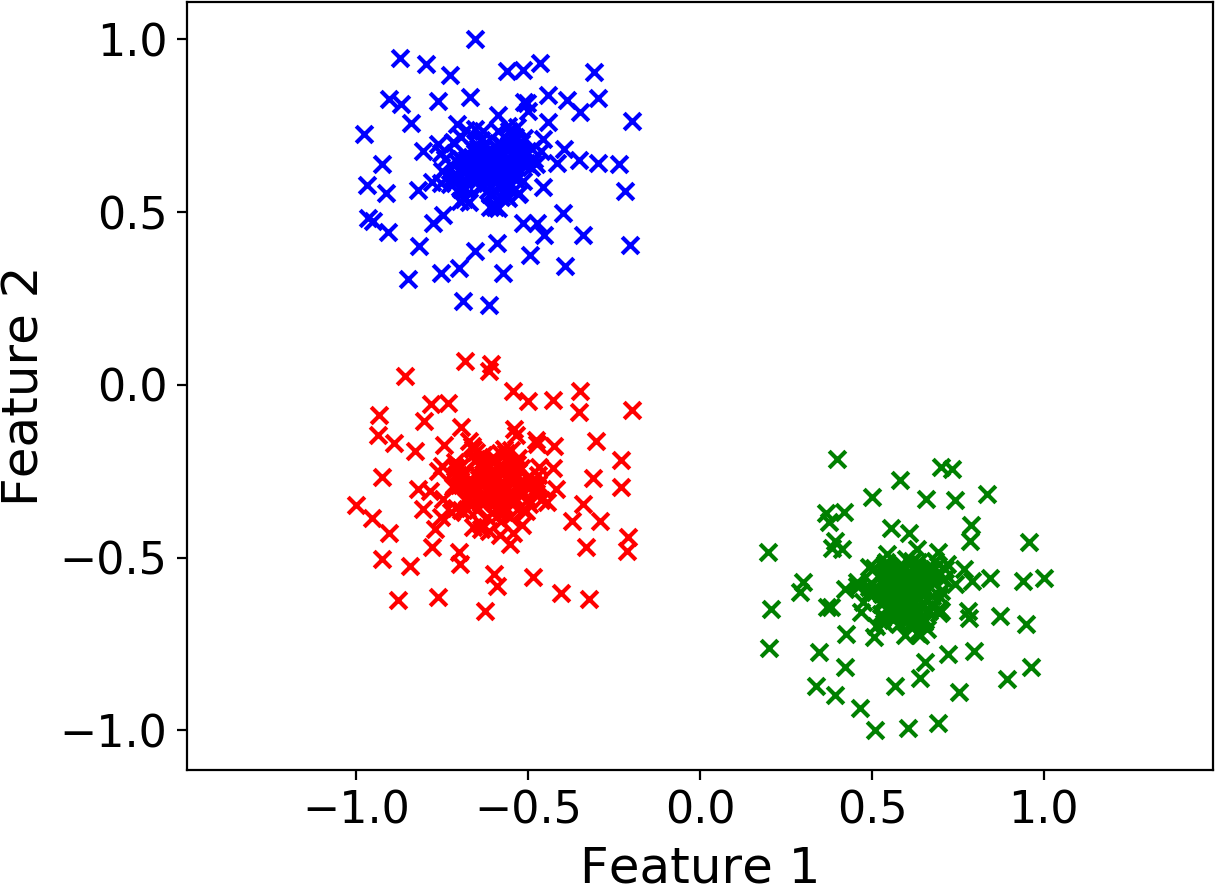} & \includegraphics[width=0.12\textwidth]{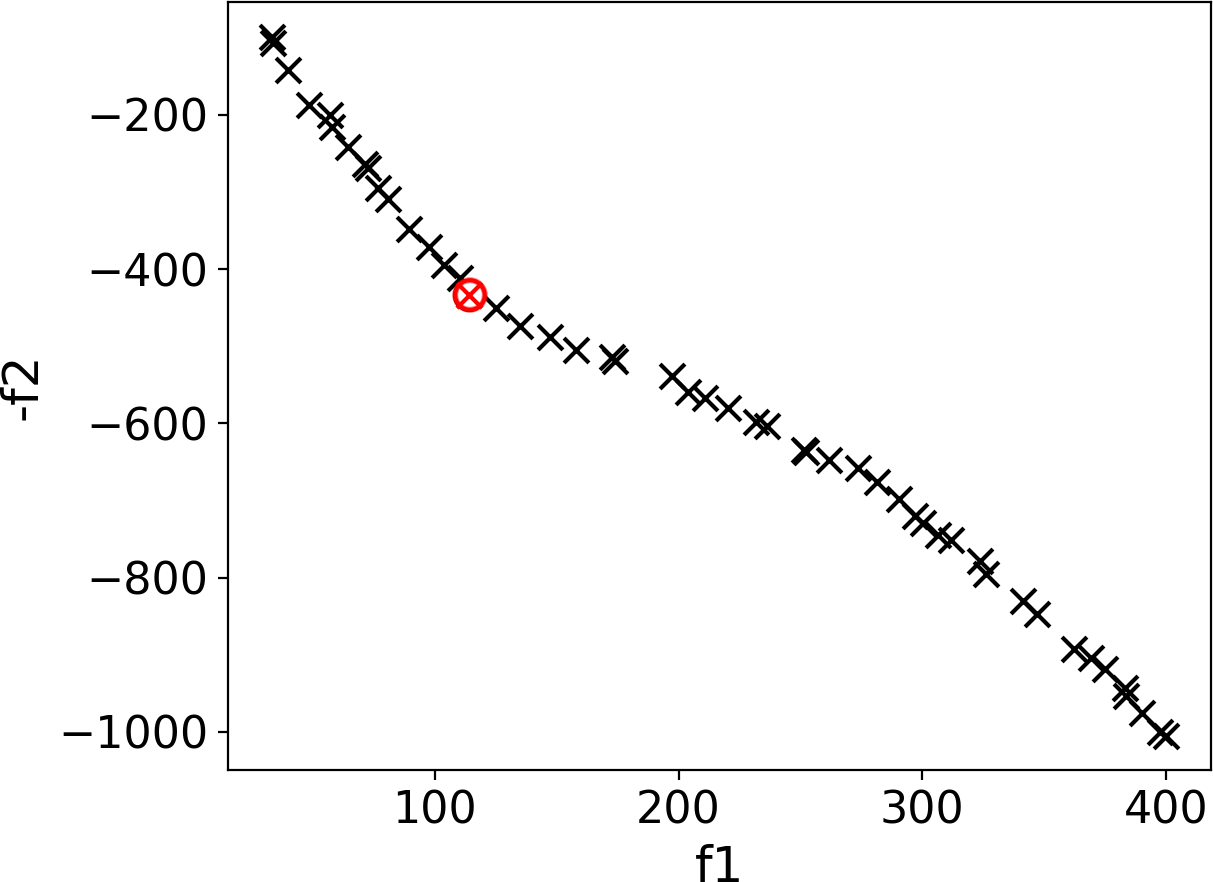} & \includegraphics[width=0.12\textwidth]{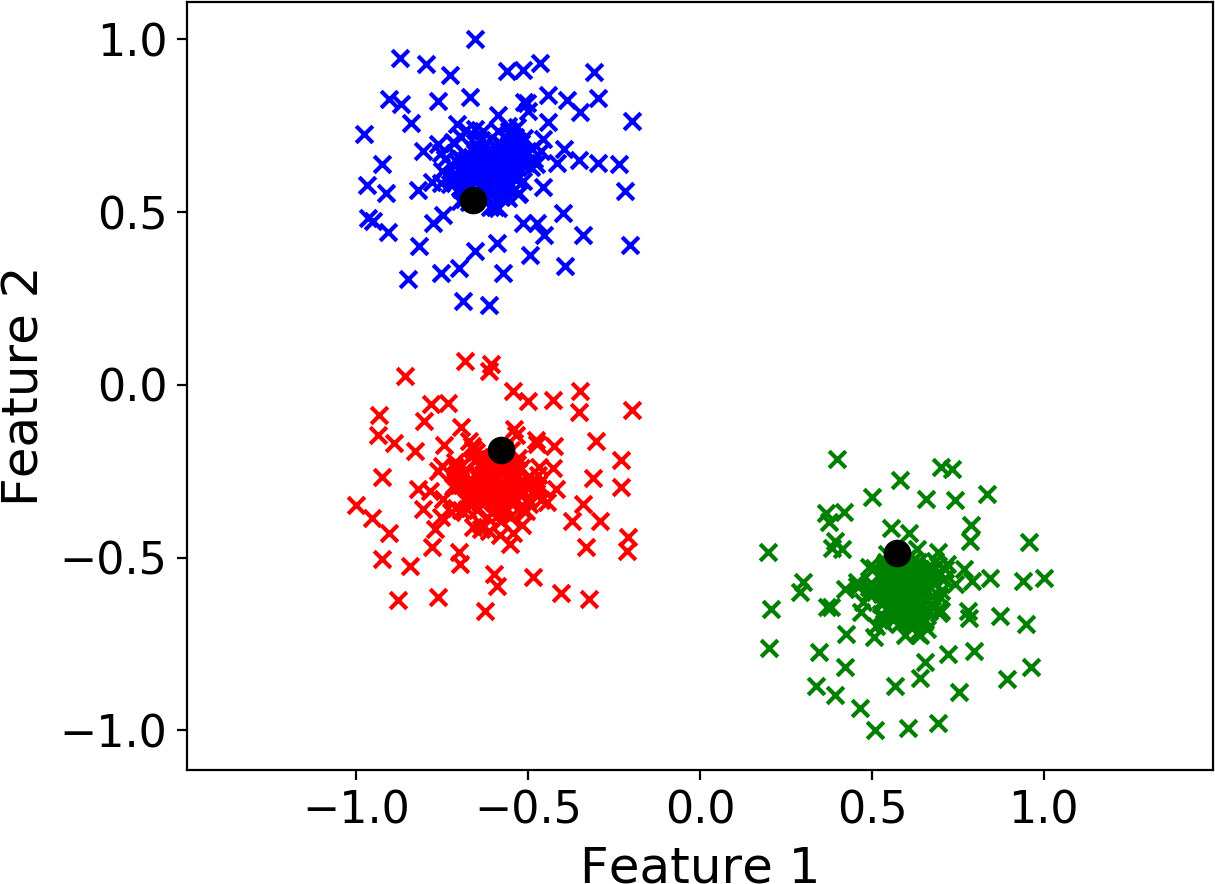} \\
        \hline
        \spheading{3 clusters, where 2 are slightly overlapped} & \includegraphics[width=0.12\textwidth]{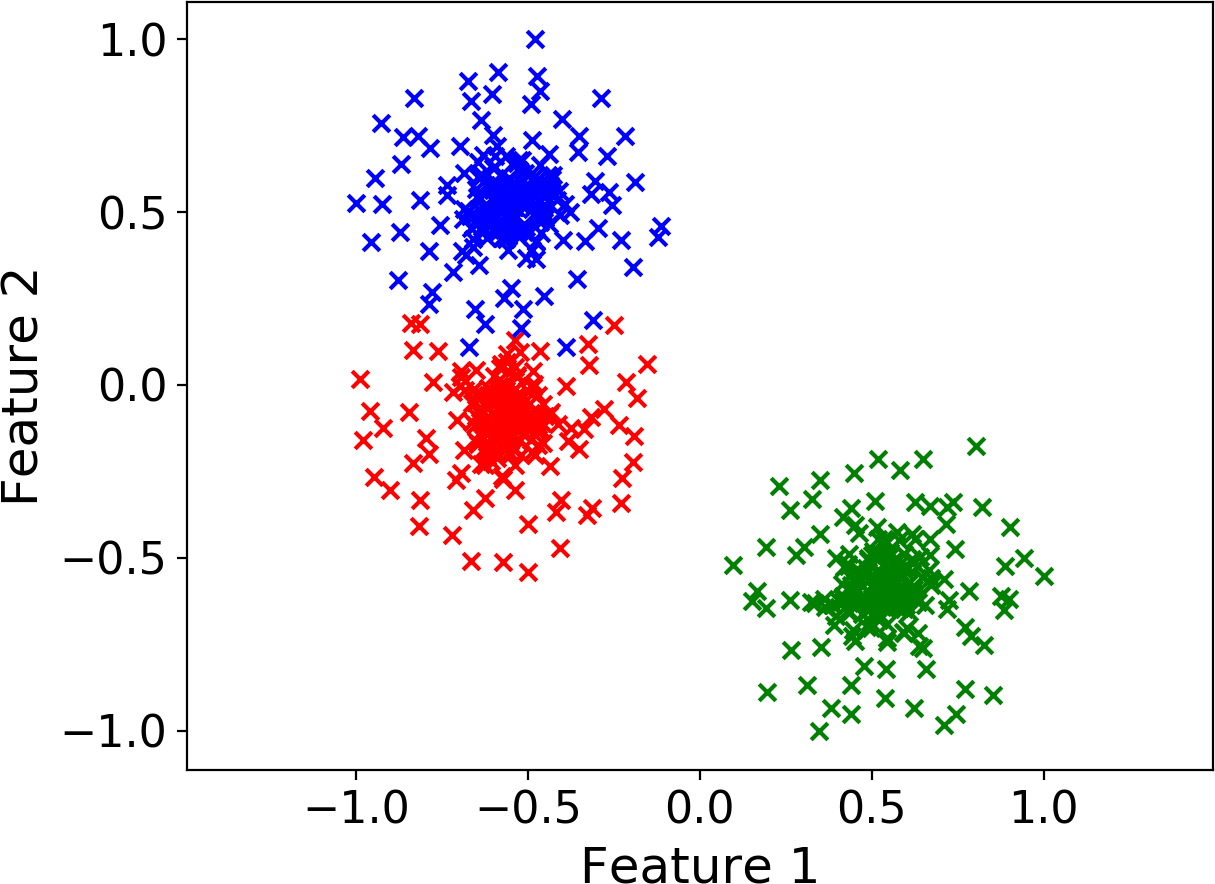} & \includegraphics[width=0.12\textwidth]{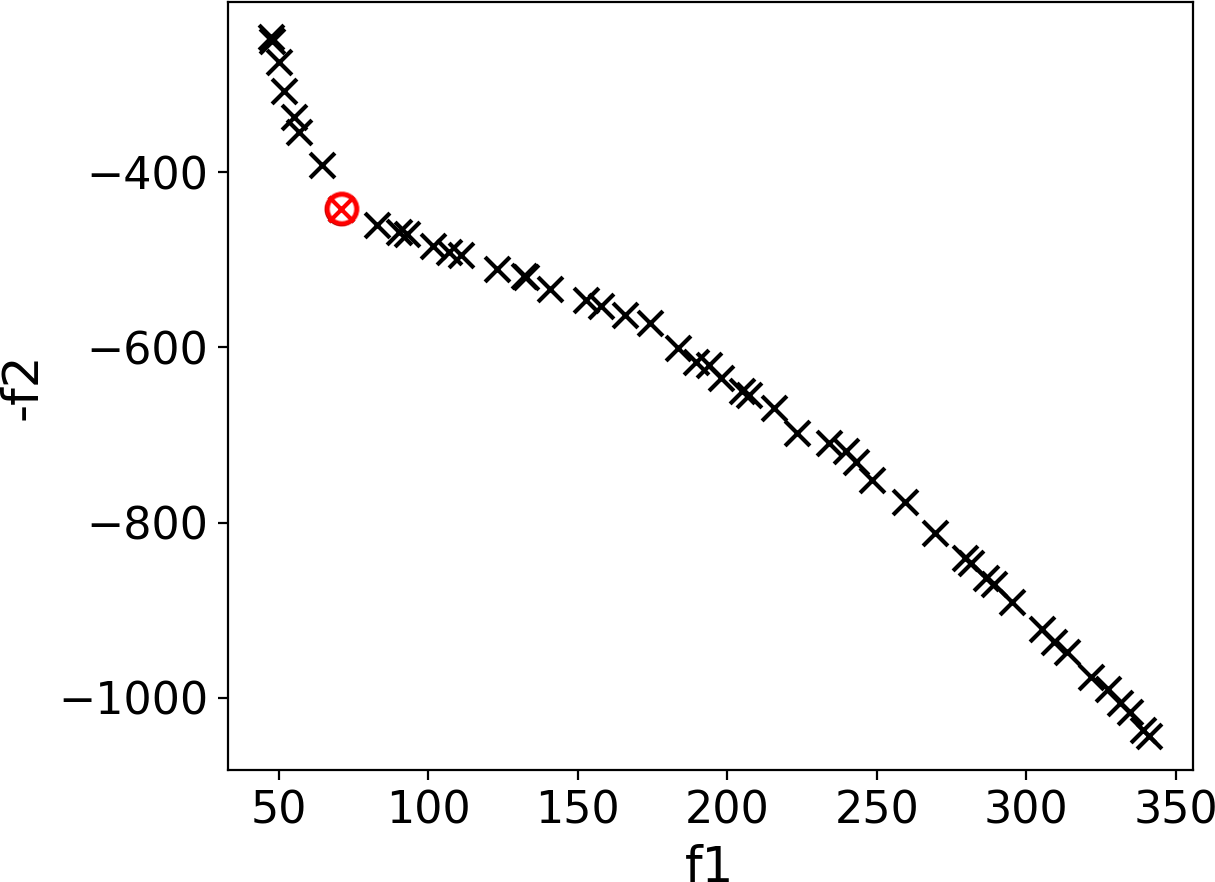} & \includegraphics[width=0.12\textwidth]{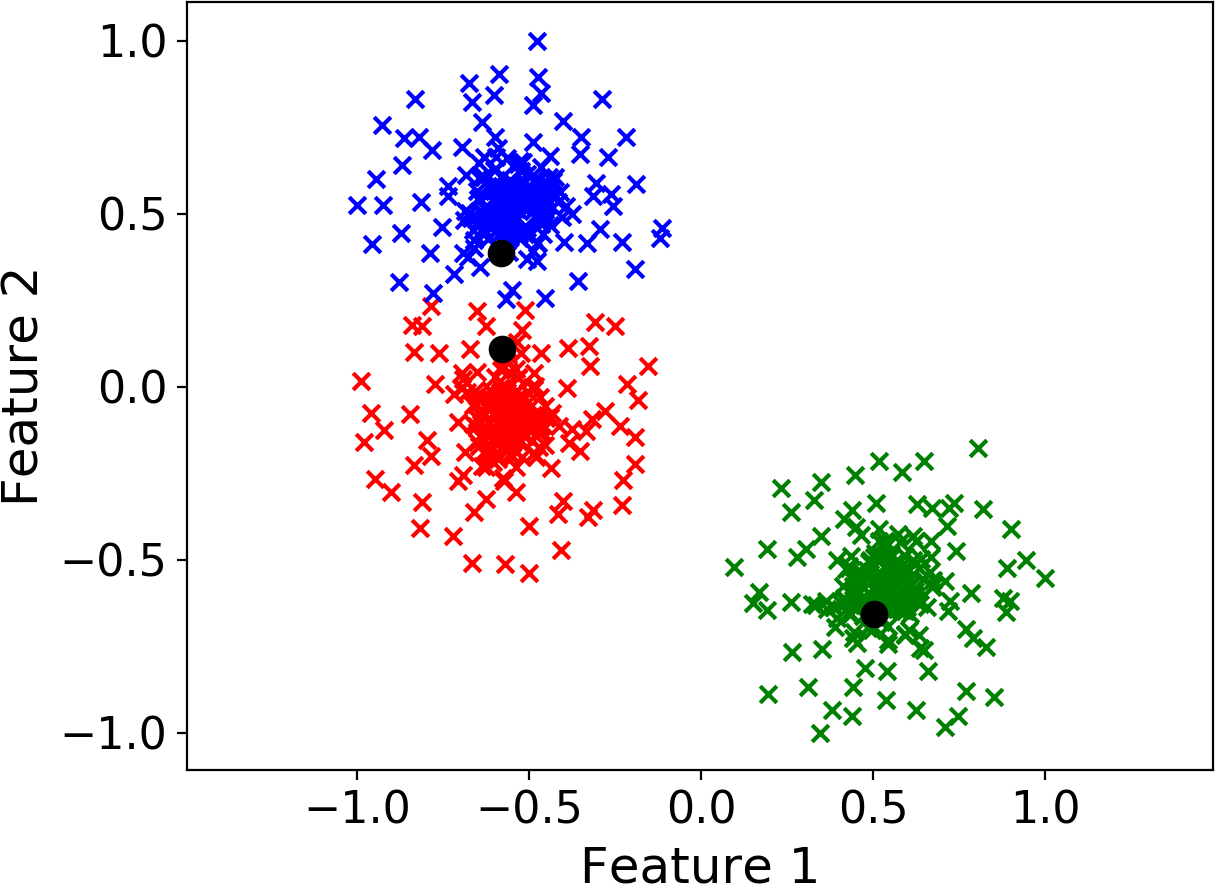} \\
        \hline
        \spheading{4 well-separated clusters} & \includegraphics[width=0.12\textwidth]{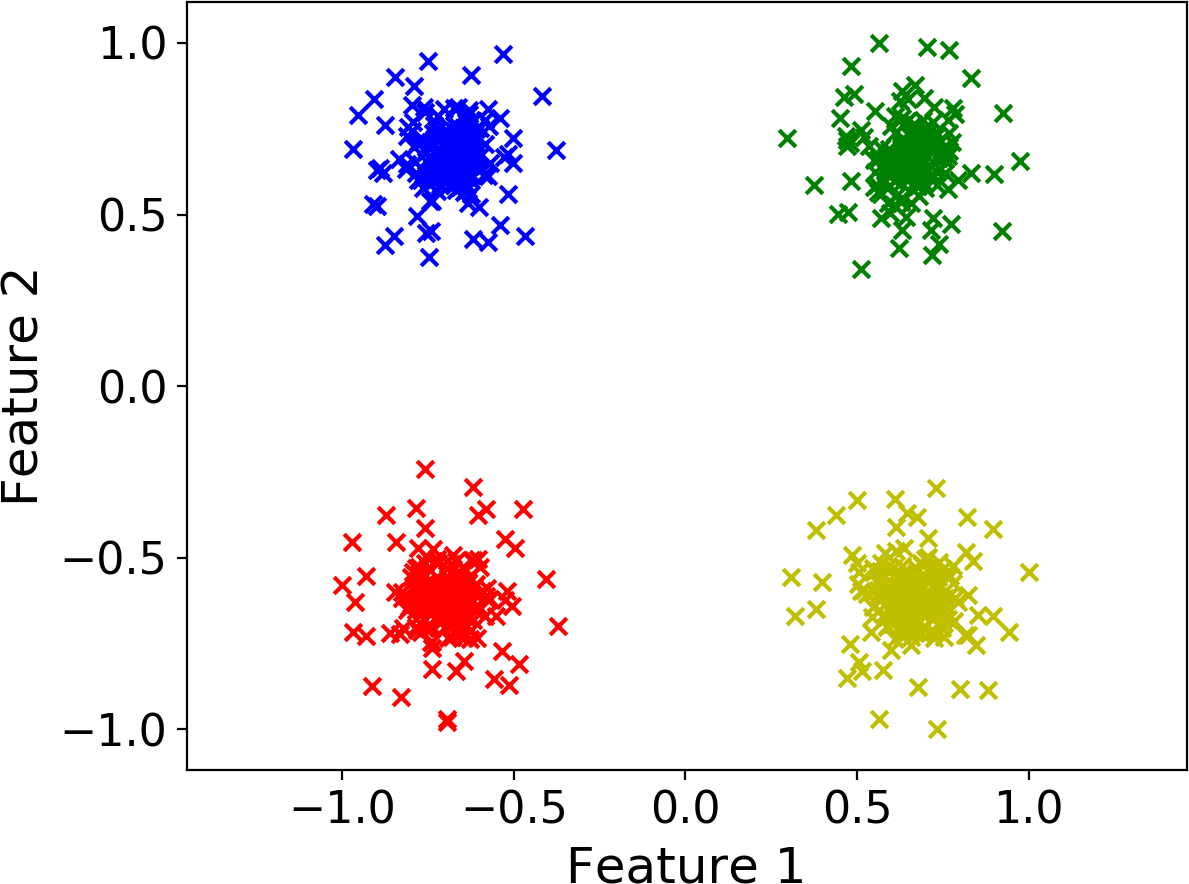} & \includegraphics[width=0.12\textwidth]{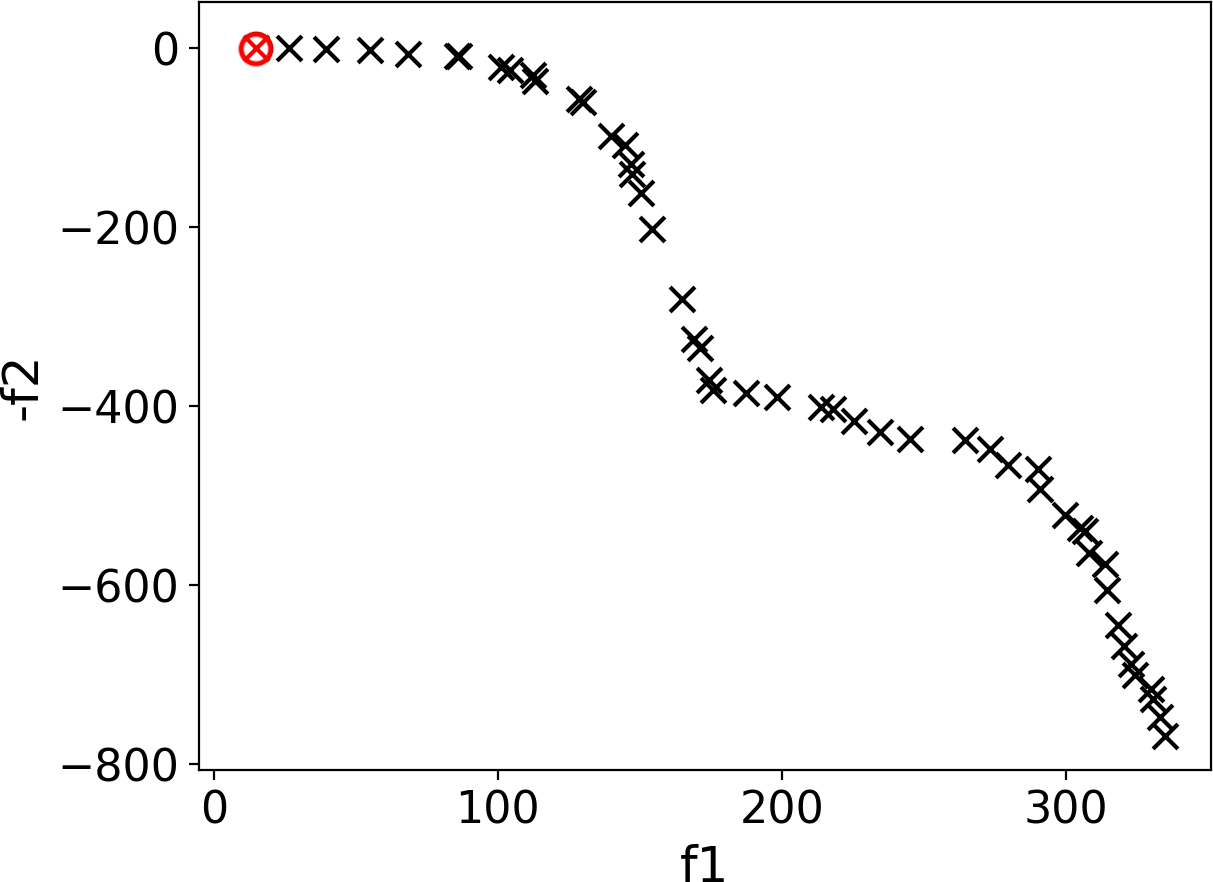} & \includegraphics[width=0.12\textwidth]{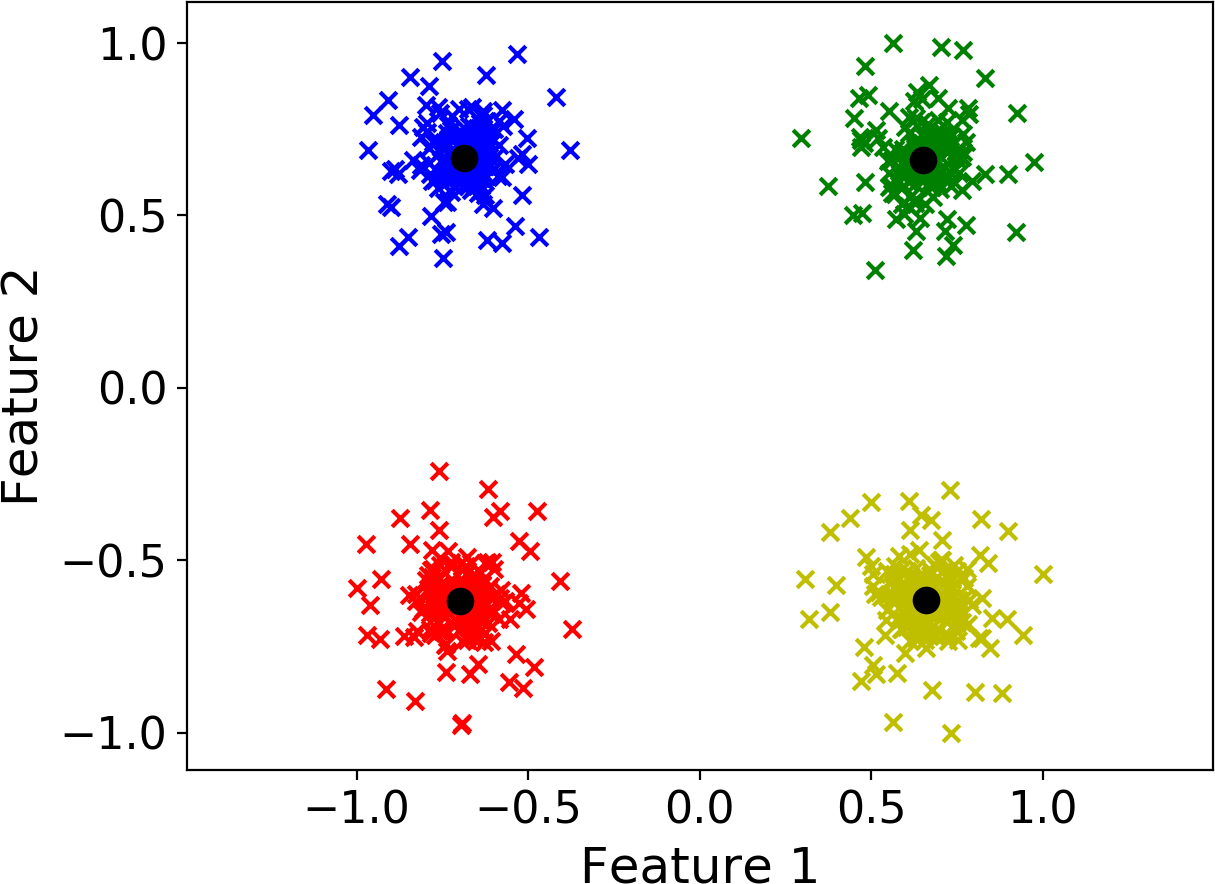} \\
        \hline
        \spheading{4 clusters, where 2 are slightly overlapped} & \includegraphics[width=0.12\textwidth]{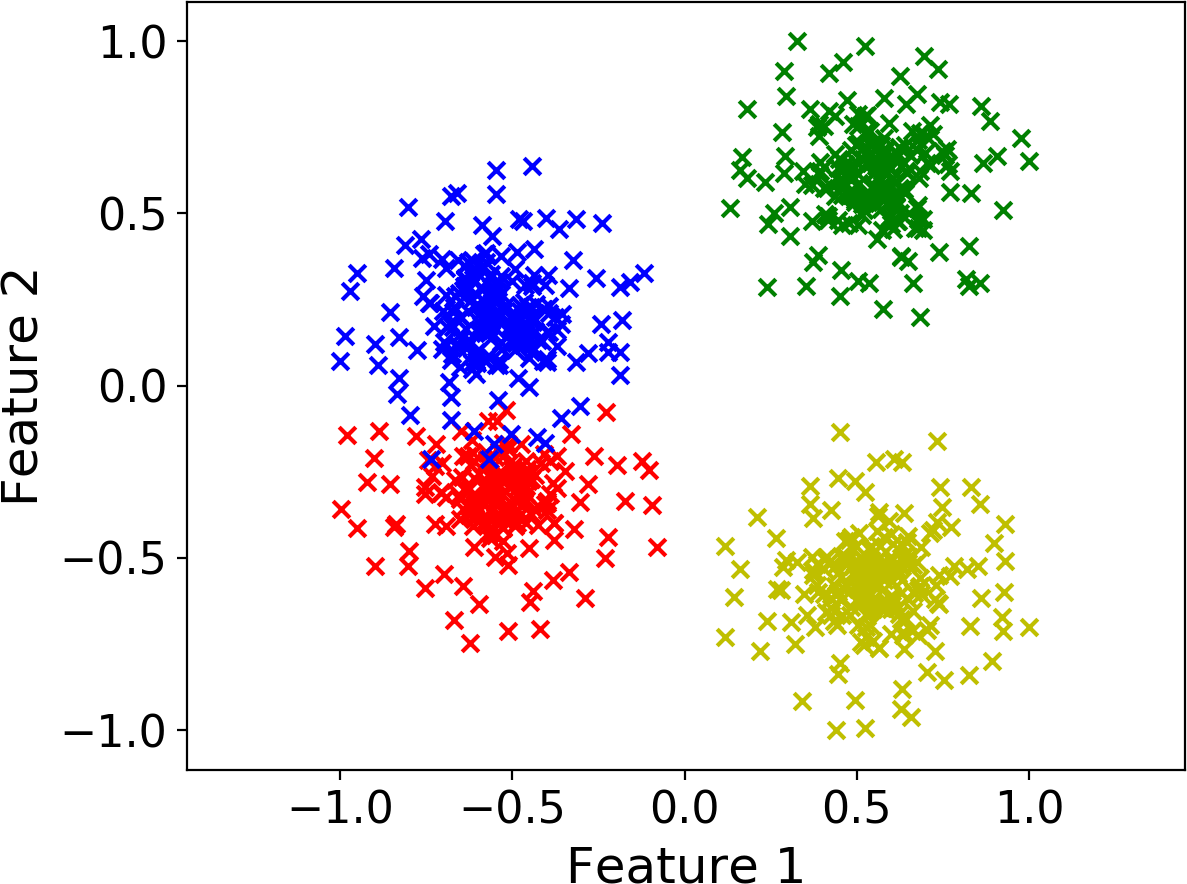} & \includegraphics[width=0.12\textwidth]{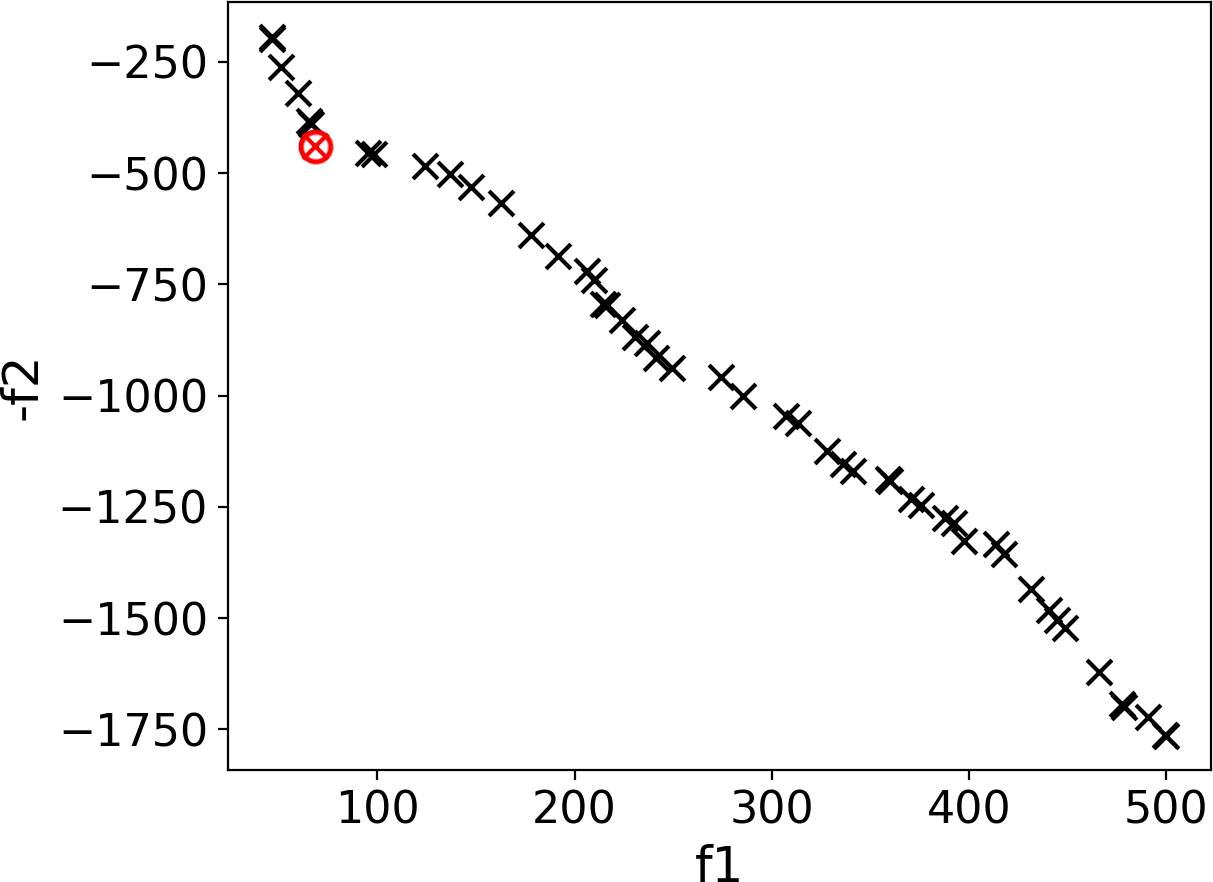} & \includegraphics[width=0.12\textwidth]{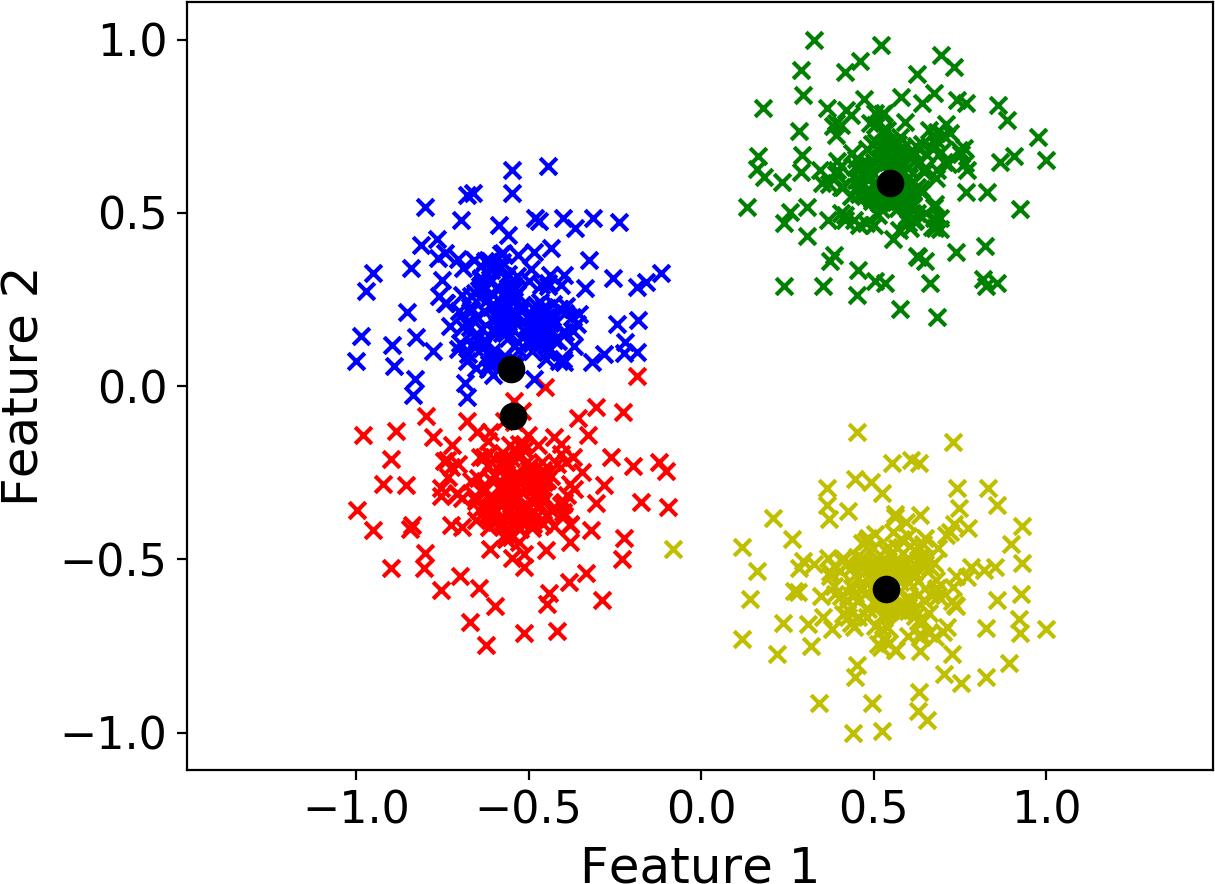} \\
        \hline
    \end{tabular}
\end{table}

\begin{table}
    \vspace{-1mm}
    \scriptsize
    \caption{(Contd. from Table VI) The selection of a suitable trade-off clustering across different datasets.}
    \label{tab_more_figs2}
    \begin{tabular}{| p{0.6cm} | p{2.1cm} | p{2.1cm} | p{2.1cm} |}
        \hline
        \vtop{\hbox{\strut Descri-}\hbox{\strut ption}} & Original Data & Pareto Front & Selected Clustering \\
        \hline
        \spheading{4 clusters, where 2 are highly overlapped} & \includegraphics[width=0.12\textwidth]{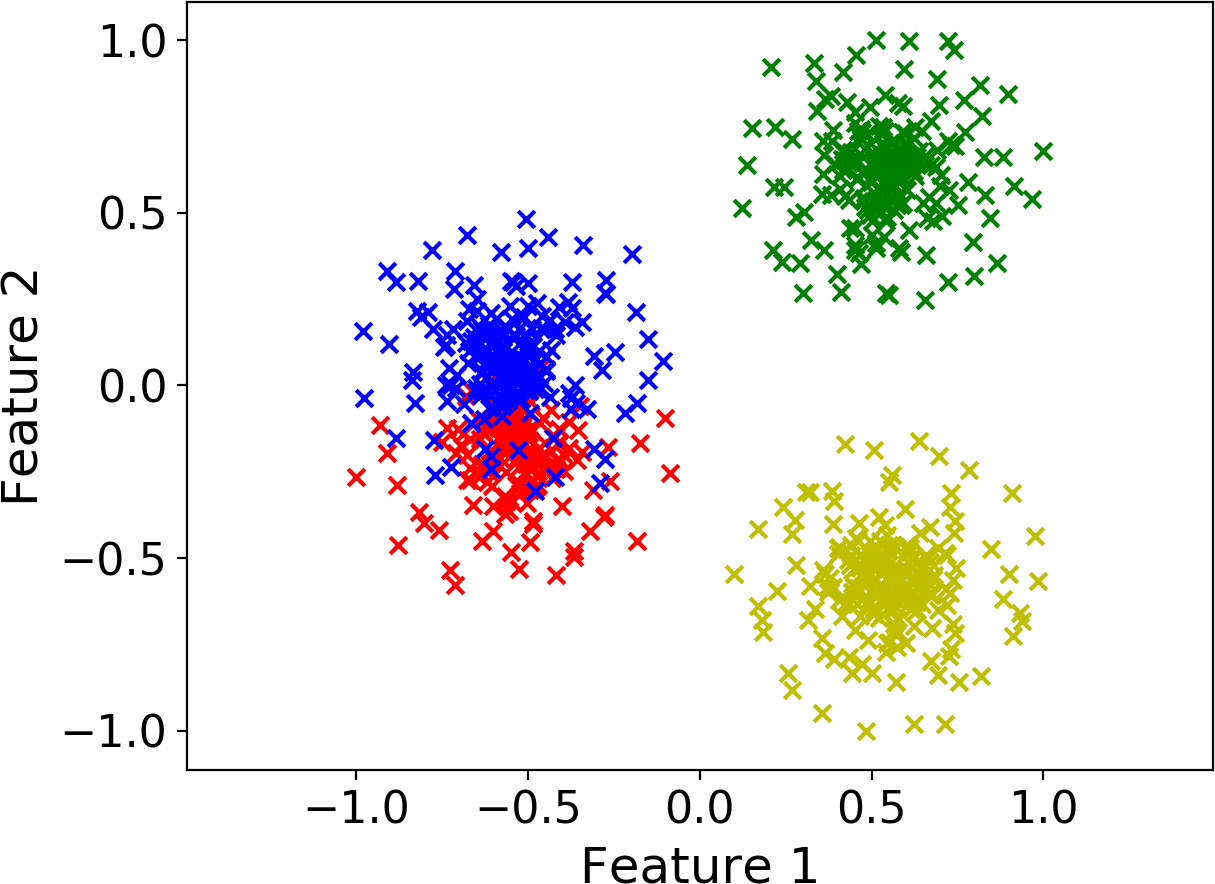} & \includegraphics[width=0.12\textwidth]{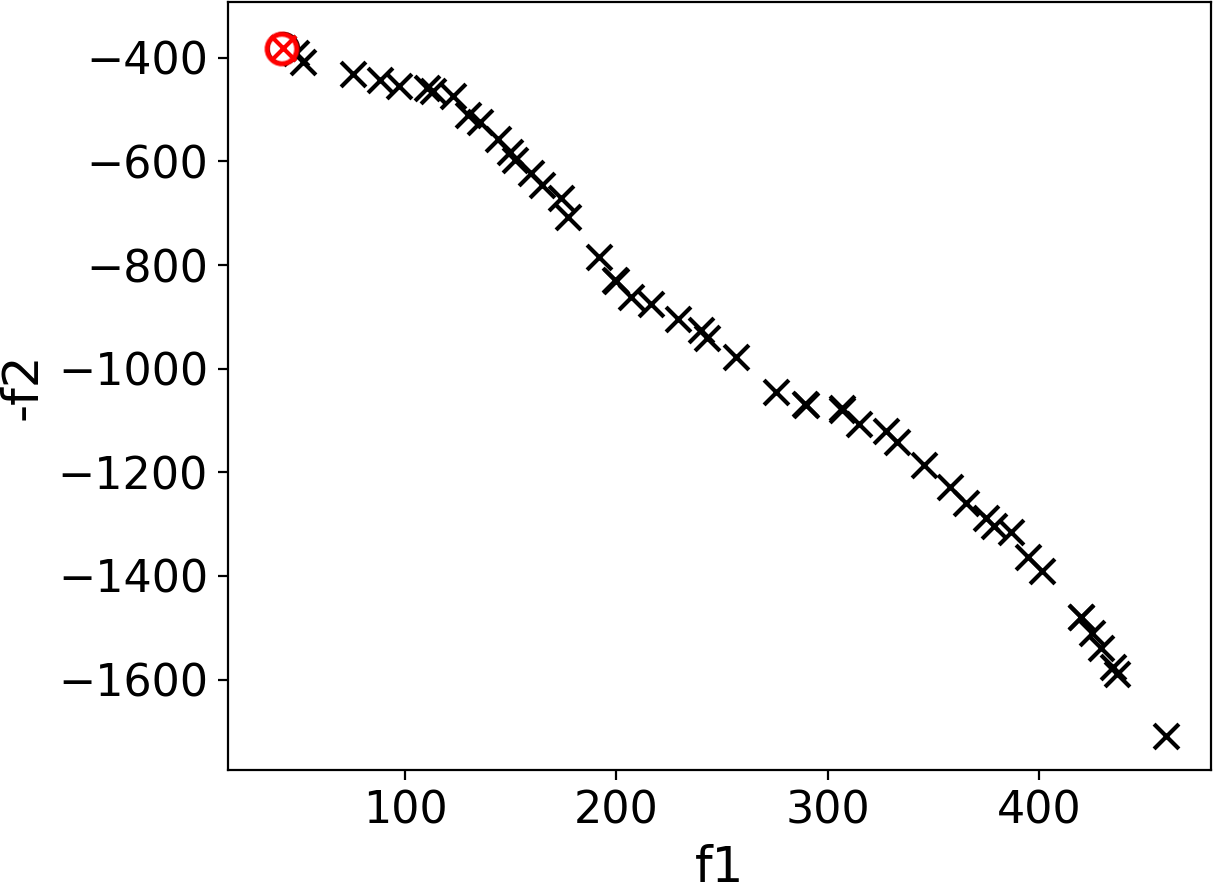} & \includegraphics[width=0.12\textwidth]{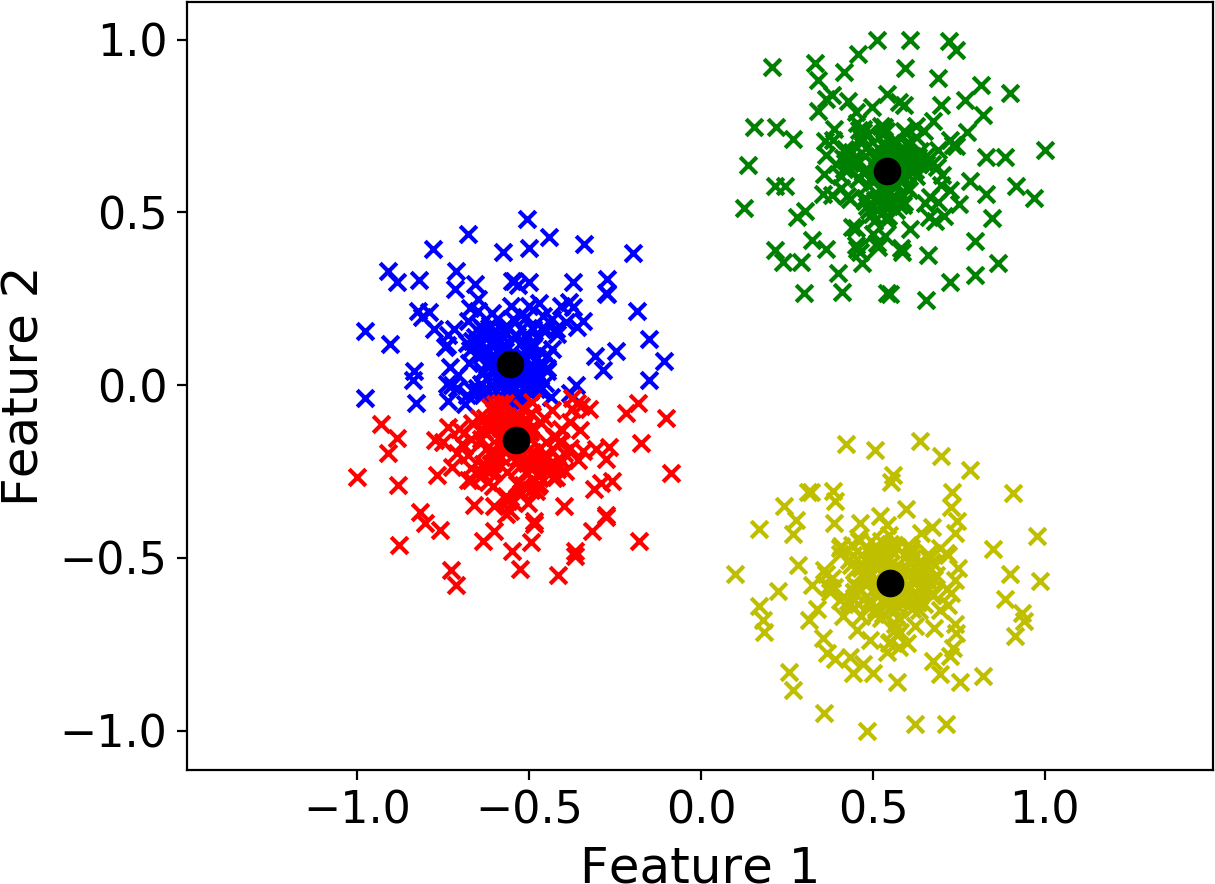} \\
        \hline
        \spheading{4 clusters, where 3 are slightly overlapped} & \includegraphics[width=0.12\textwidth]{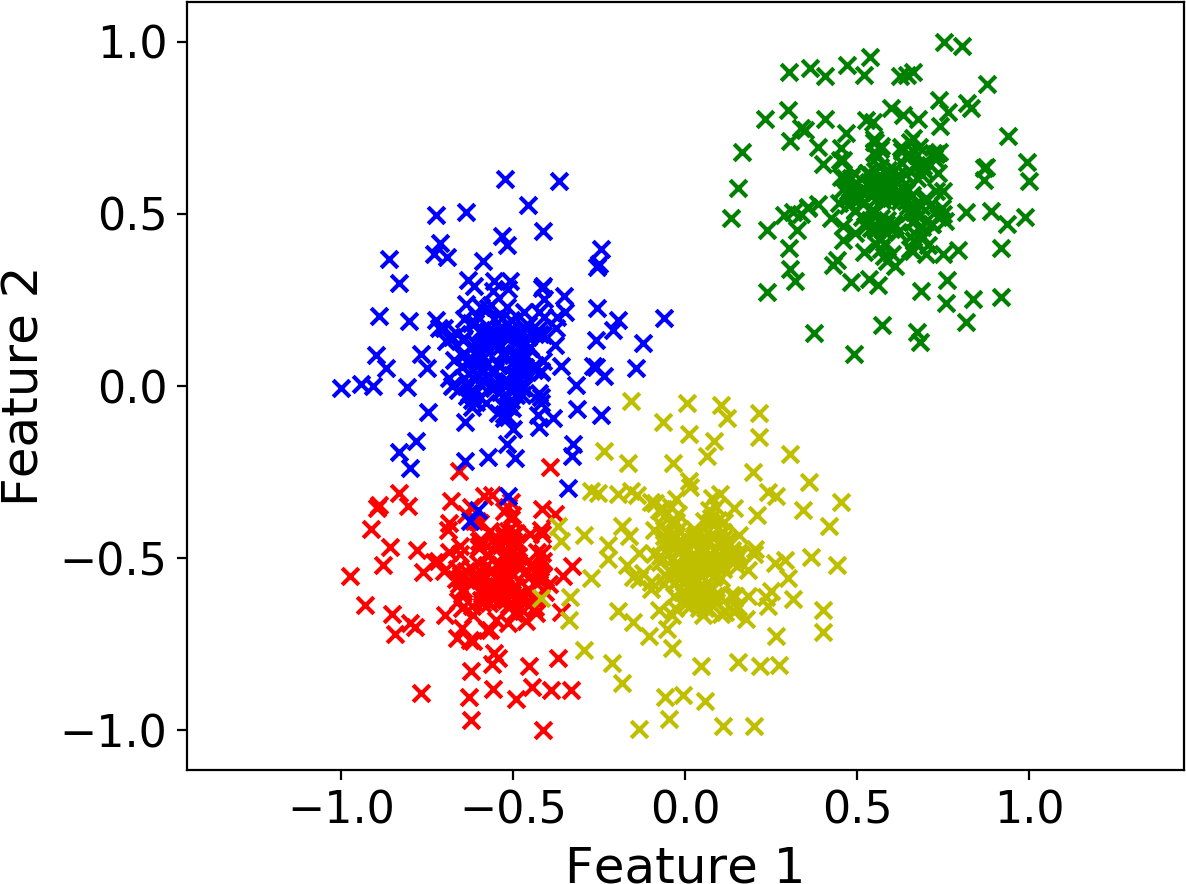} & \includegraphics[width=0.12\textwidth]{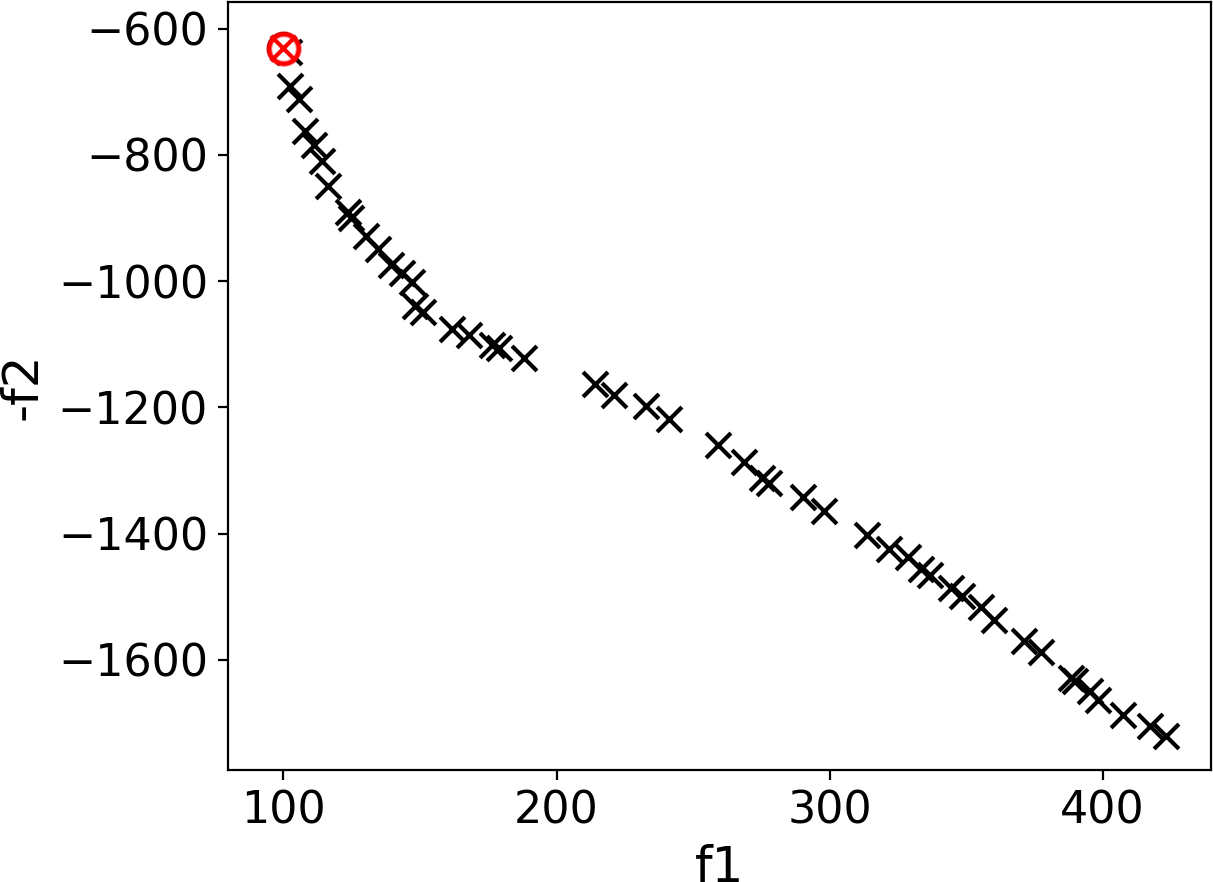} & \includegraphics[width=0.12\textwidth]{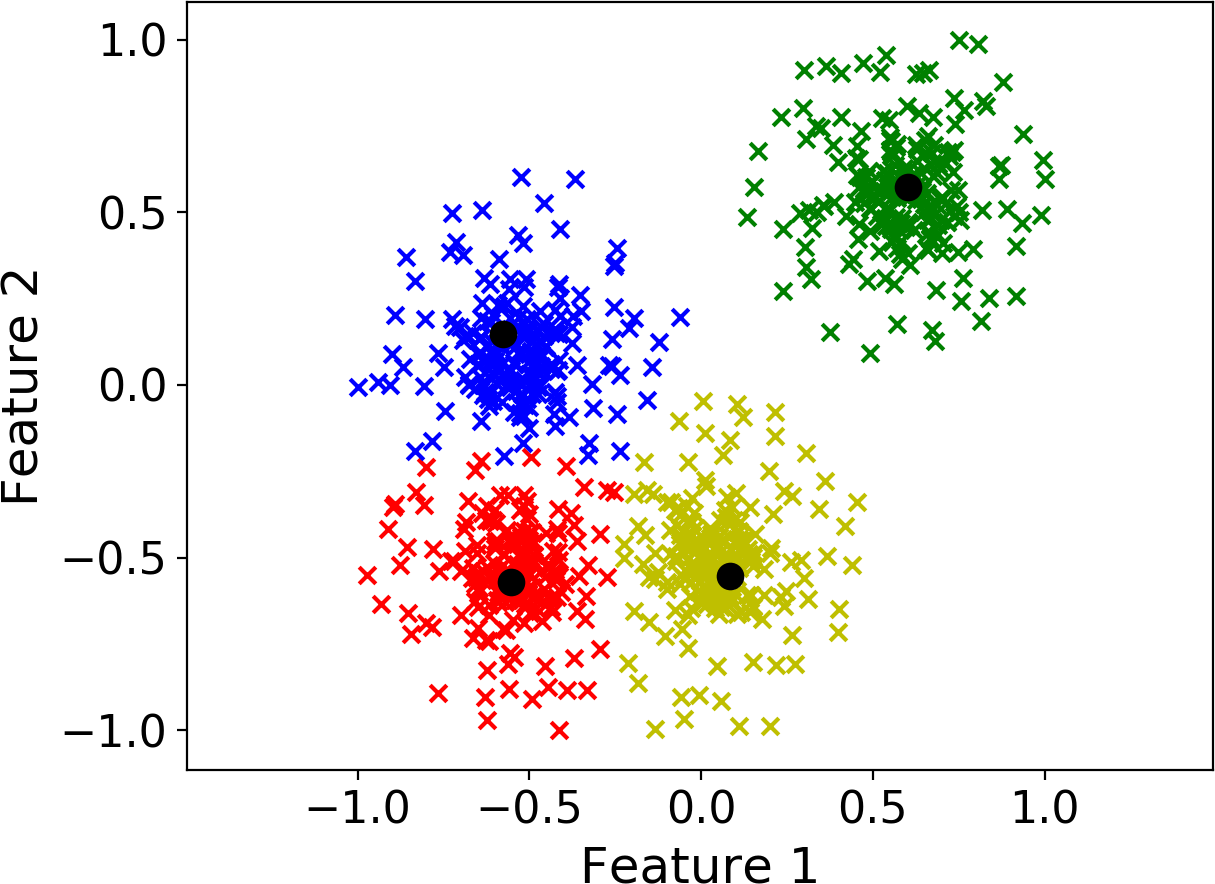} \\
        \hline
        \spheading{4 clusters, where 3 are highly overlapped} & \includegraphics[width=0.12\textwidth]{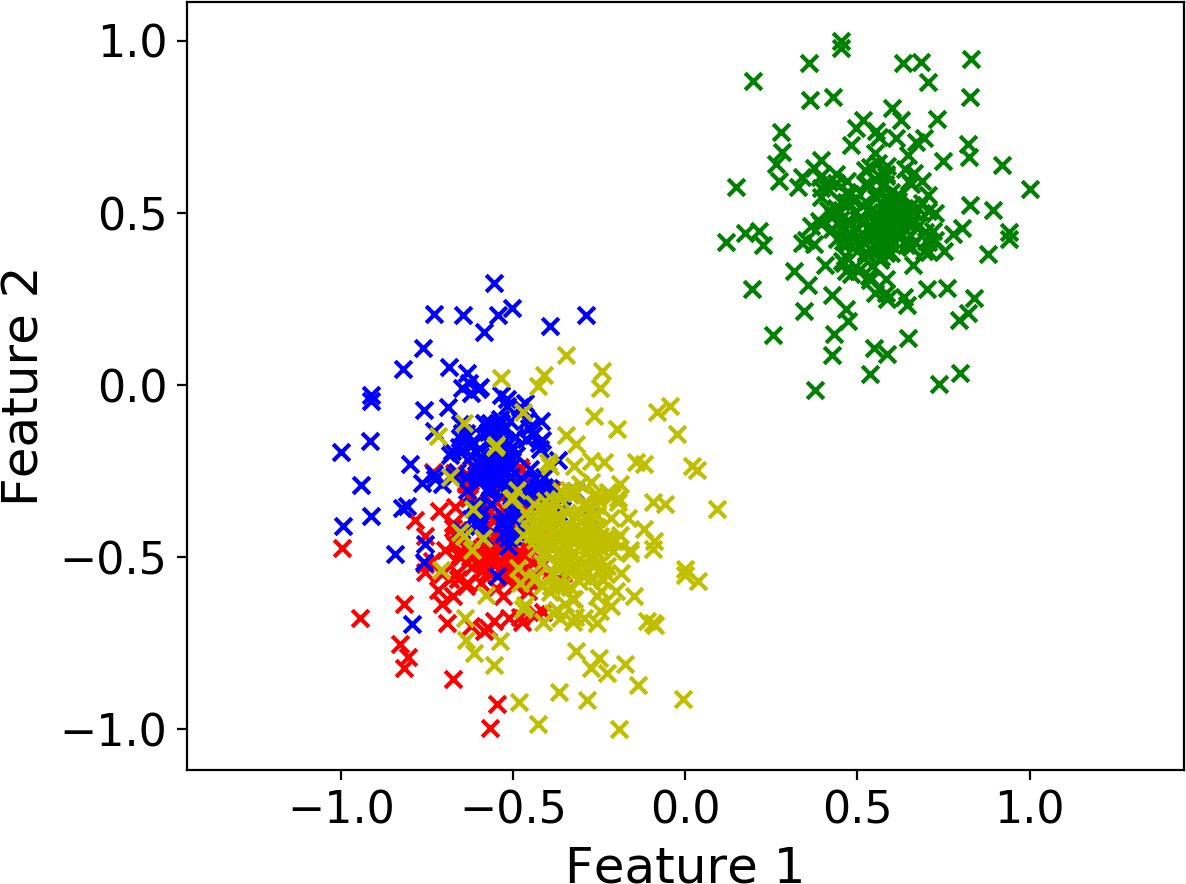} & \includegraphics[width=0.12\textwidth]{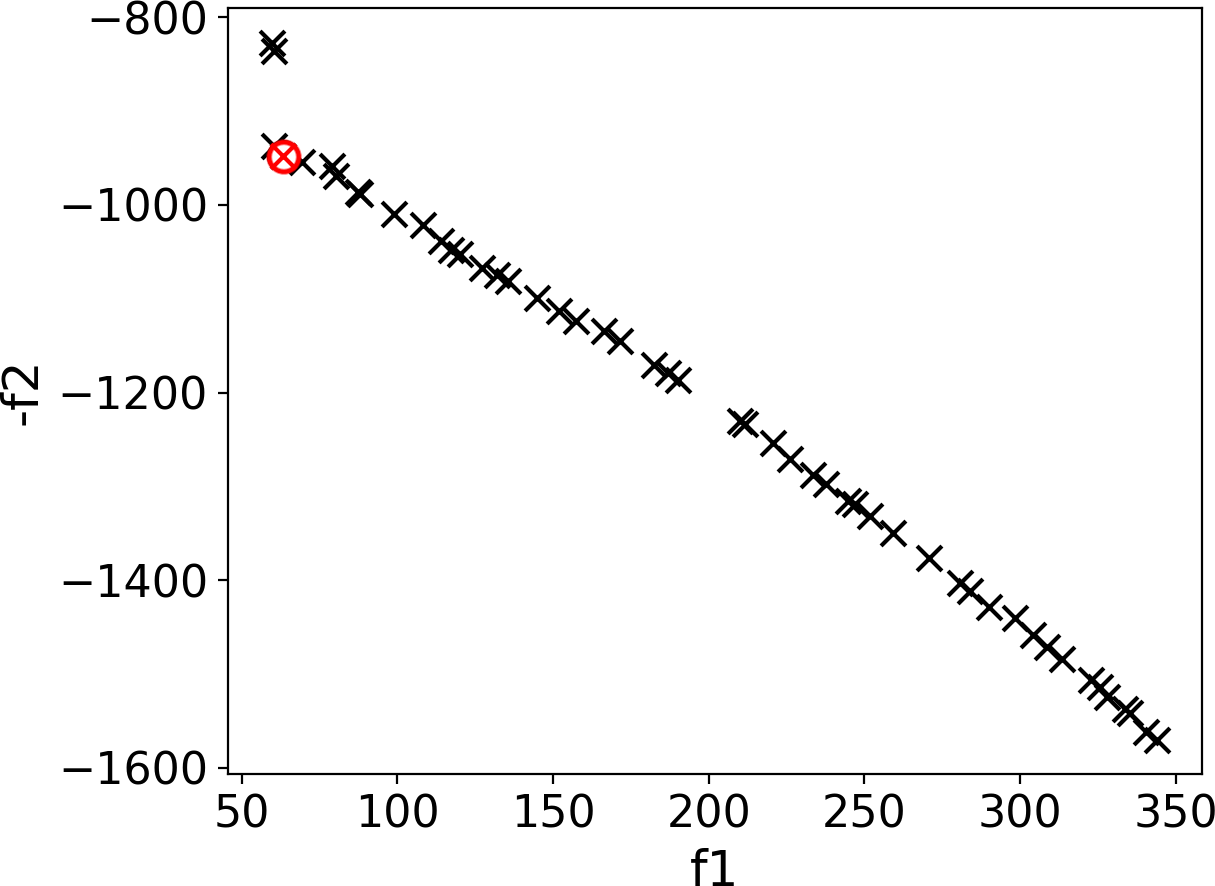} & \includegraphics[width=0.12\textwidth]{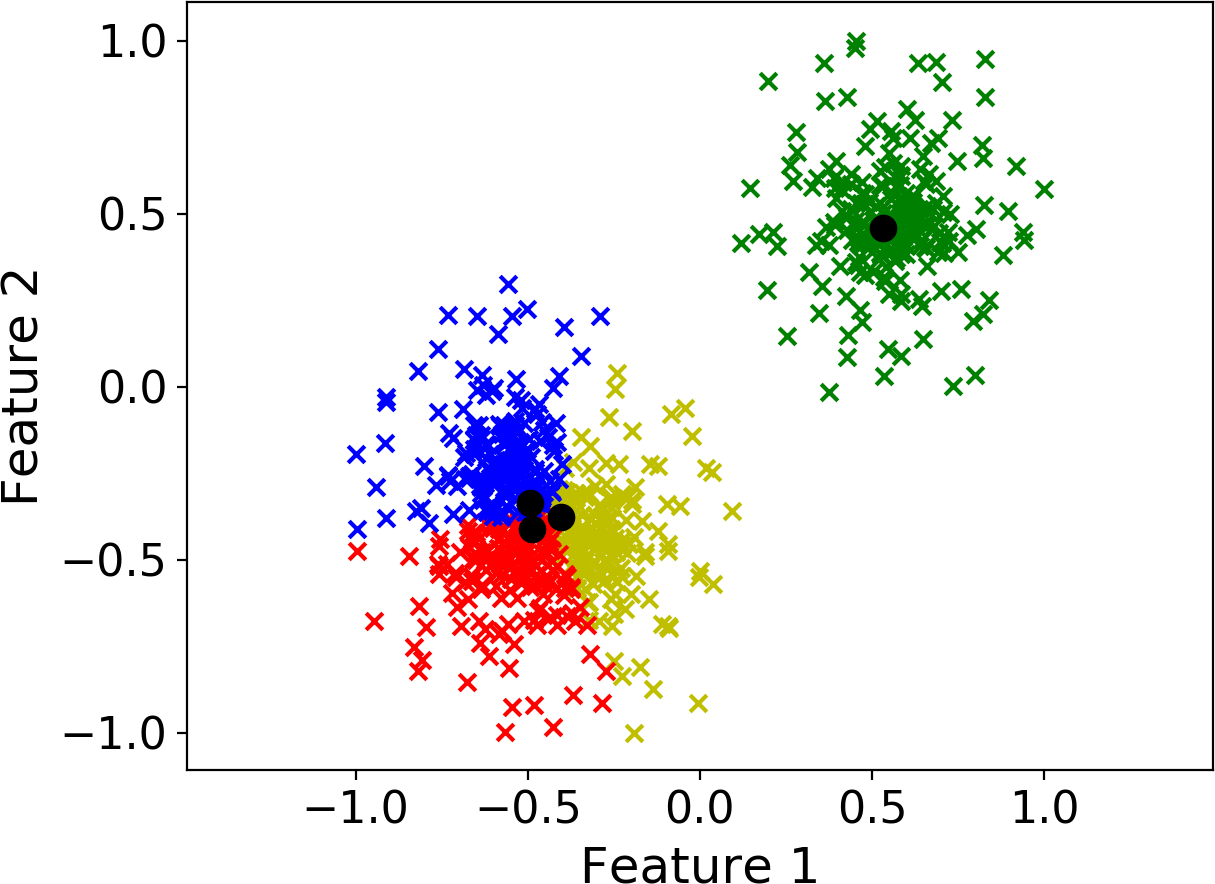} \\
        \hline
        \spheading{5 well-separated clusters, with 1 in the middle} & \includegraphics[width=0.12\textwidth]{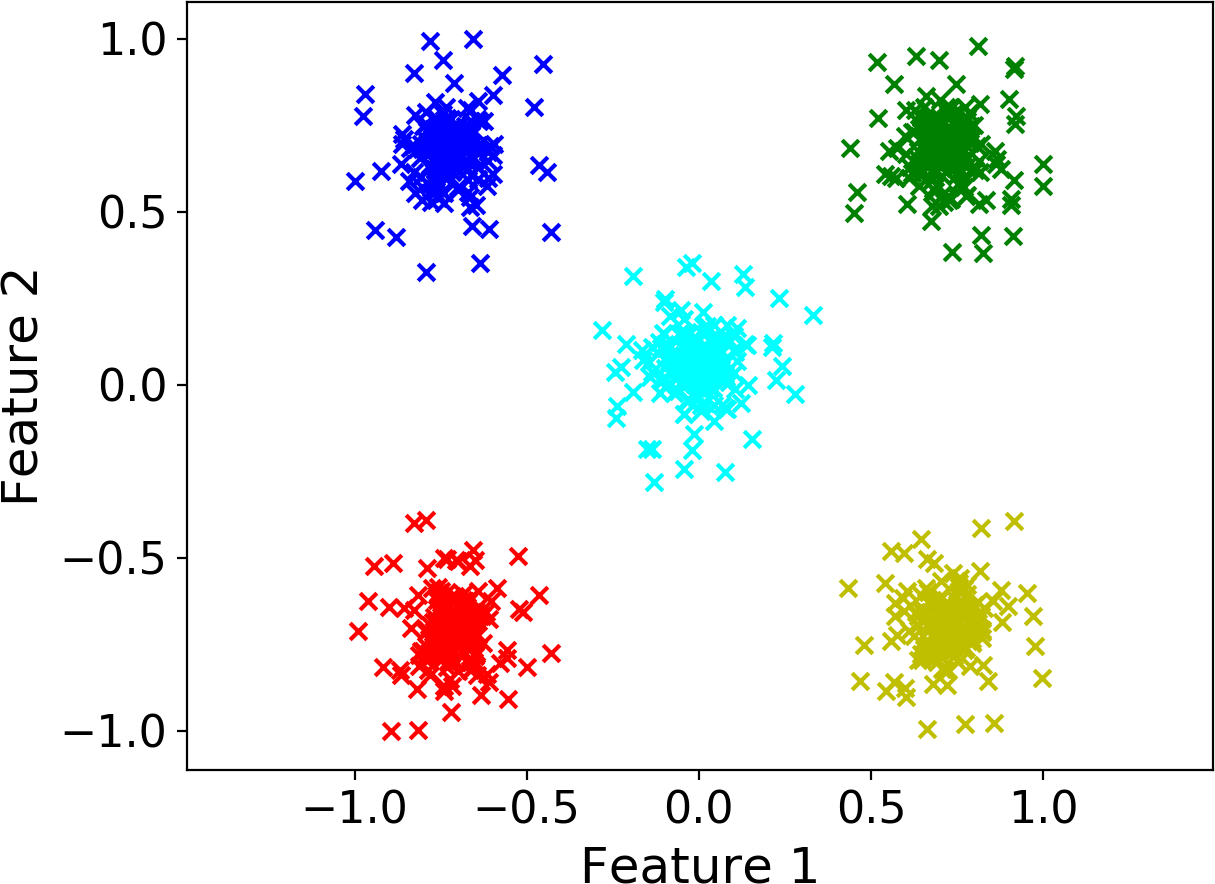} & \includegraphics[width=0.12\textwidth]{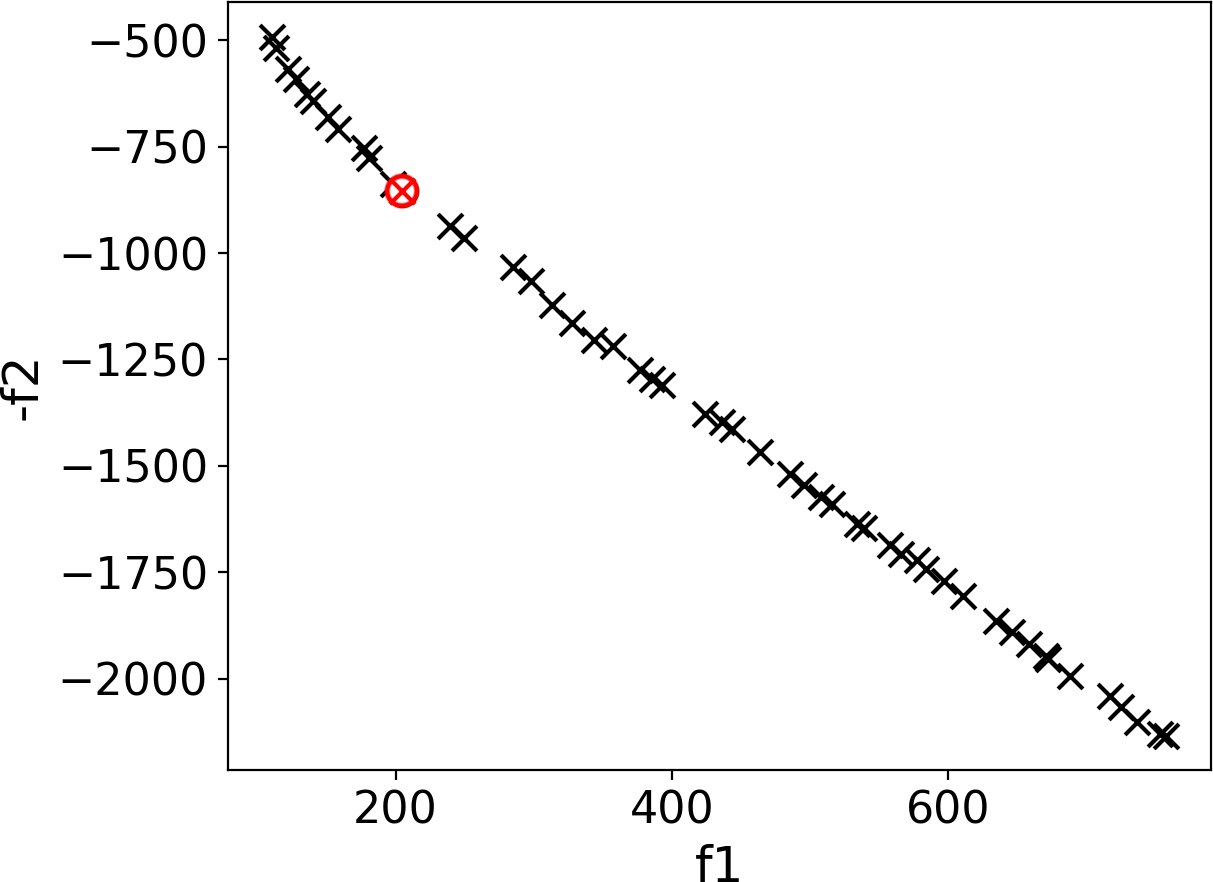} & \includegraphics[width=0.12\textwidth]{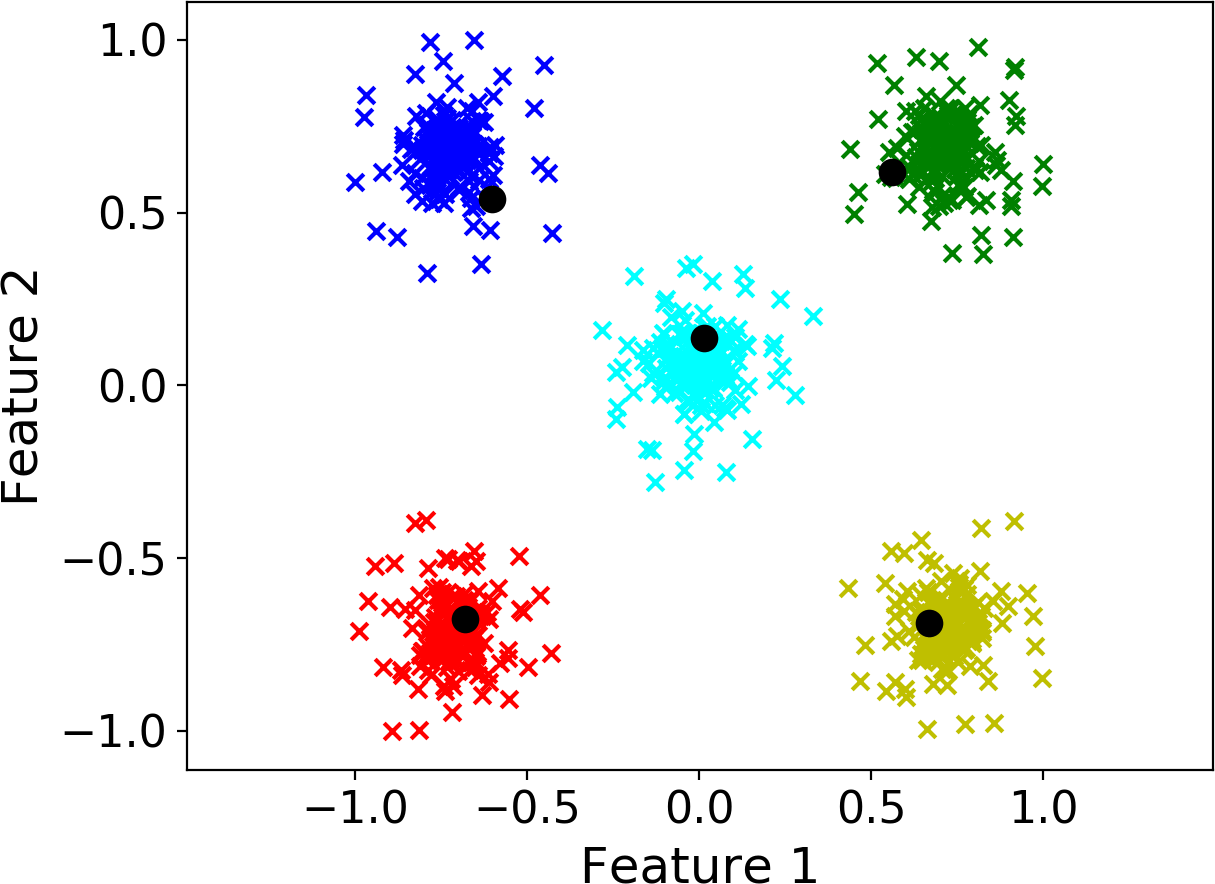} \\
        \hline
        \spheading{5 well-separated clusters, with 1 to the right} & \includegraphics[width=0.12\textwidth]{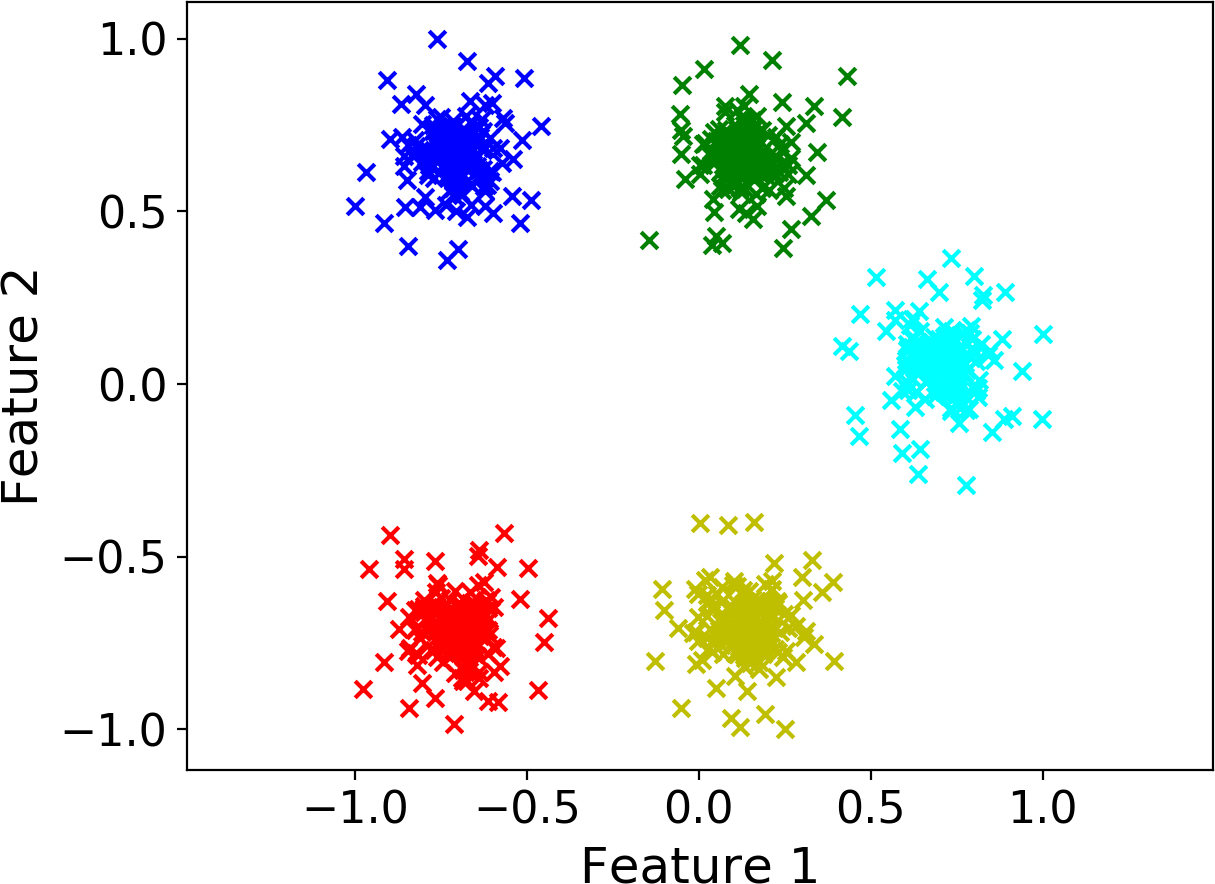} & \includegraphics[width=0.12\textwidth]{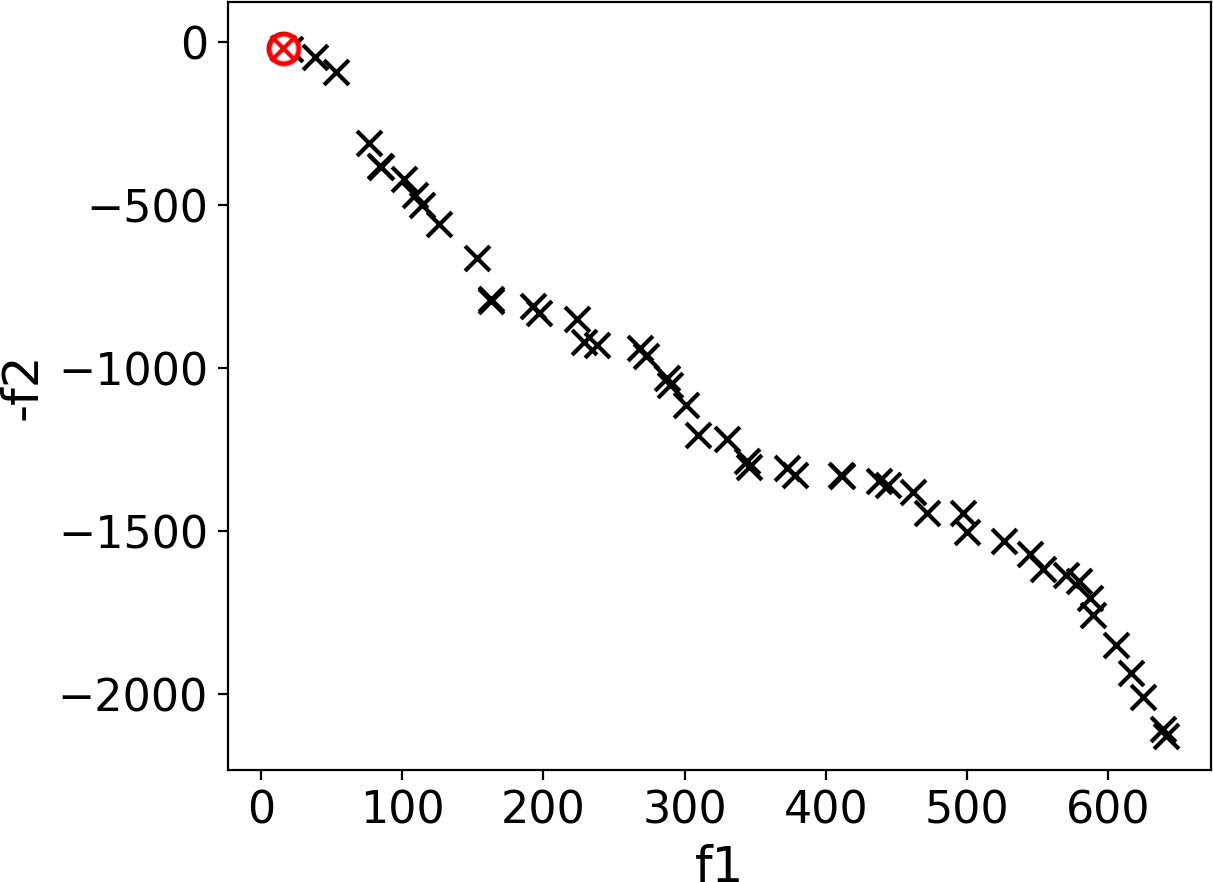} & \includegraphics[width=0.12\textwidth]{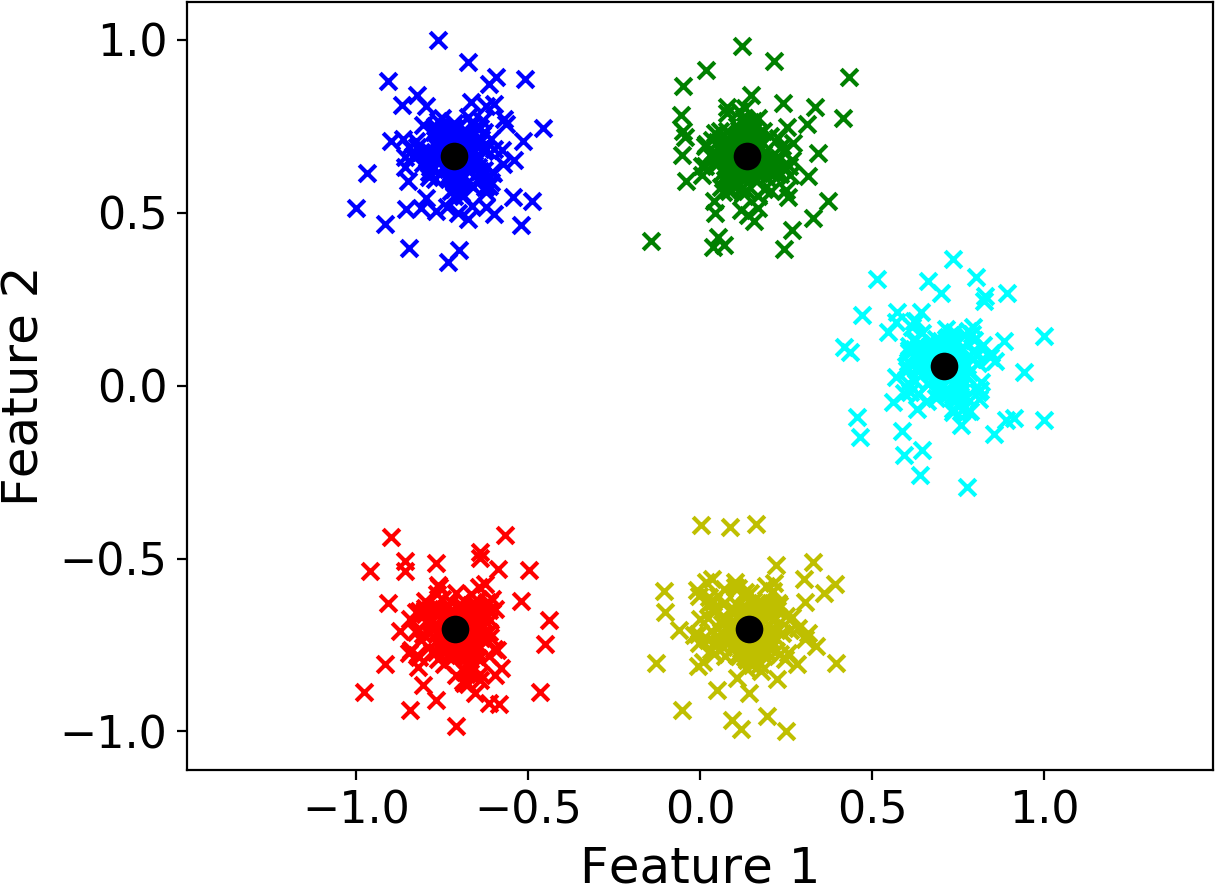} \\
        \hline
    \end{tabular}
\end{table}

% ~~~~~~~~~~~~~~~~~~~~~~~~~~~~~ %

\subsection{Comparison of Pareto Fronts}
\label{sec_pf}

\begin{figure}
    \vspace{-1mm}
    \centering
	\includegraphics[width=0.45\textwidth]{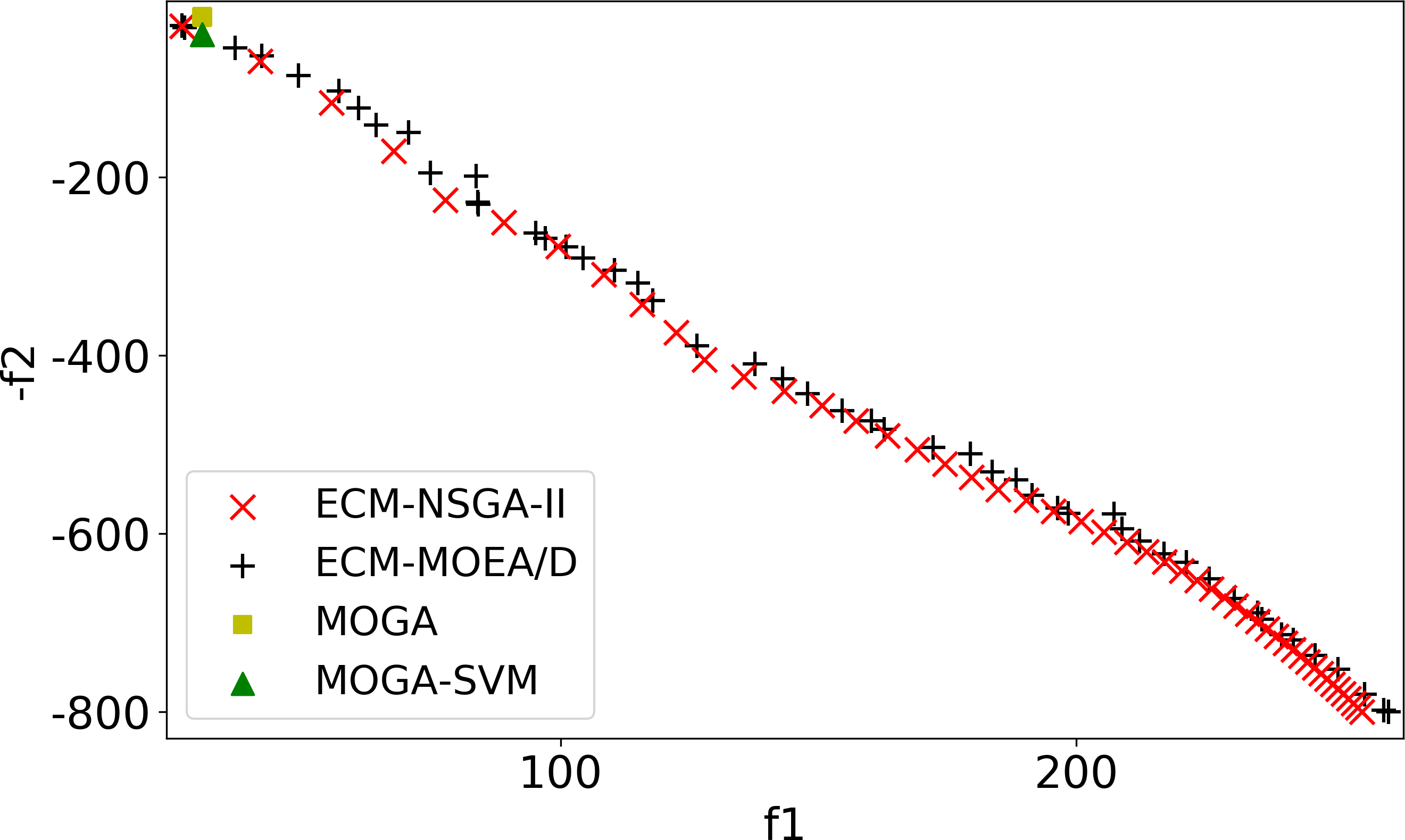}
	\caption{Comparison of Pareto fronts on the \emph{proximity1} synthetic dataset}
	\label{fig_pareto_comp}
	\vspace{-4mm}
\end{figure}

\begin{table}
\centering
{\scriptsize
\caption{\label{tab_ss_results}Comparison of Schott's Spacing Metric over synthetic and real datasets}
\begin{tabular}{l p{0.8cm} p{0.8cm} p{1cm} p{1cm}}
\hline
Dataset & MOGA & MOGA-SVM & ECM-NSGA-II & ECM-MOEA/D \\
\hline
proximity1 & 0.0032 & 0.0009 & 5.4672 & \textbf{12.1241} \\
proximity2 & 0.0067 & 0.0393 & 5.7940 & \textbf{14.4735} \\
proximity3 & 0.0005 & 0.0006 & 5.4845 & \textbf{14.5695} \\
proximity4 & 0.0130 & 0.0089 & 4.0915 & \textbf{13.7317} \\
proximity5 & 0.0169 & 0.0124 & 4.2396 & \textbf{17.9207} \\
spread1 & 0.0073 & 0.0035 & 7.7586 & \textbf{10.1440} \\
spread2 & 0.0016 & 0.0004 & 8.7522 & \textbf{10.7135} \\
spread3 & 0.0003 & 0.0034 & 5.9851 & \textbf{11.5221} \\
spread4 & 0.0079 & 0.0046 & 6.7693 & \textbf{11.9174} \\
spread5 & 0.0026 & 0.0051 & 5.6010 & \textbf{14.4791} \\
2d-4c-no0 & 0.0531 & 0.0270 & 23.5203 & \textbf{68.2197} \\
2d-4c-no1 & 0.0959 & 0.0674 & 16.9215 & \textbf{54.8581}\\
2d-4c-no2 & 0.3107 & 0.0682 & 11.5239 & \textbf{69.1033} \\
2d-4c-no3 & 0.0083 & 0.0110 & 10.0184 & \textbf{27.4230} \\
2d-4c-no4 & 0.4271 & 0.1615 & 7.3958 & \textbf{34.9729} \\
2d-4c-no5 & 0.0427 & 0.0587 & 21.1299 & \textbf{74.8314} \\
2d-4c-no6 & 0.0530 & 0.0876 & 26.7573 & \textbf{88.9569} \\
2d-4c-no7 & 0.4471 & 0.0978 & 10.4615 & \textbf{29.5684} \\
2d-4c-no8 & 0.0023 & 0.0073 & 17.5021 & \textbf{31.4343} \\
2d-4c-no9 & 0.0365 & 0.1292 & 13.5963 & \textbf{42.1341} \\
10d-4c-no0 & 0 & 0 & 20.0628 & \textbf{23.3810} \\
10d-4c-no1 & 0.4321 & 0.1629 & 11.6972 & \textbf{26.8866} \\
10d-4c-no2 & 0.4550 & 0.7392 & 10.9053 & \textbf{26.5723} \\
10d-4c-no3 & 0 & 0 & 19.5474 & \textbf{29.7252} \\
10d-4c-no4 & 0 & 0 & \textbf{15.0934} & 14.2382 \\
10d-4c-no5 & 0 & 0 & 10.2654 & \textbf{22.6472} \\
10d-4c-no6 & 0.7411 & 0.5154 & 12.1797 & \textbf{20.0835} \\
10d-4c-no7 & 0.1167 & 0.0759 & 23.0278 & \textbf{29.3453} \\
10d-4c-no8 & 0.3362 & 0.3096 & 13.4835 & \textbf{26.6004} \\
10d-4c-no9 & 0.5165 & 0.2850 & 17.1202 & \textbf{26.9990} \\
\hline
B. scale & 0.0032 & 0.0009 & 5.4672 & \textbf{12.1241} \\
B. Tissue & 0.0555 & 0.0420 & 0.9547 & \textbf{14.7230} \\
wdbc & 2.0183 & 0.5215 & 12.7512 & \textbf{22.2190} \\
banknote & 5.3040 & 6.5190 & 13.0640 & \textbf{16.4670} \\
echo & 0.1930 & 0.4210 & 0.7680 & \textbf{1.1300} \\
Ecoli & 0.0230 & 0.01920 & 2.5770 & \textbf{7.9210} \\
Iris & 0.0159 & 0.0439 & 2.0320 & \textbf{2.6140} \\
magic & 0.0234 & 0.0195 & \textbf{2.5231} & 1.3950 \\
seeds & 0.1110 & 0.1196 & \textbf{1.5400} & 1.4900 \\
sonar & 6.4006 & 7.7954 & \textbf{189.9391} & 17.0041 \\
ukm & 0.7044 & 0.2100 & 2.6060 & \textbf{5.8870} \\
wine & 0.0515 & 0.0103 & 1.1233 & \textbf{1.1758} \\
colon cancer & 4.6602 & 4.1465 & \textbf{91.5271} & 37.8690 \\
lung cancer & 29.2246 & 23.3281 & \textbf{2614.5028} & 747.9953 \\
prostate cancer & 0.2376 & 0.3565 & \textbf{1133.2959} & 775.876322 \\
\hline
\end{tabular}
}
\end{table}

\begin{table}
\centering
{\scriptsize
\caption{\label{tab_ie_results}Comparison of Pareto fronts on real and synthetic datasets using Epsilon Indicator}
\begin{tabular}{p{1.5cm} p{0.75cm} p{0.8cm} p{0.9cm} p{0.75cm} p{0.8cm} p{0.92cm}}
\hline
 & \multicolumn{3}{c}{ECM-NSGA-II} & \multicolumn{3}{c}{ECM-MOEA/D} \\
\cmidrule{2-7}
\multirow{2}{*}{Dataset} & MOGA & MOGA-SVM & ECM-MOEA/D & MOGA & MOGA-SVM & ECM-NSGA-II \\
\hline
proximity1 & 26.6168 & 26.6180 & 1.0202 & 26.6130 & 26.6142 & \textbf{0.9999} \\
proximity2 & 11.2888 & 11.2888 & 1.0019 & 11.2883 & 11.2883 & \textbf{0.9999} \\
proximity3 & 5.4729 & 5.4729 & 1.0288 & 5.4706 & 5.4706 & \textbf{0.9996} \\
proximity4 & 3.2577 & 3.2577 & 1.0286 & 3.2575 & 3.2575 & \textbf{0.9999} \\
proximity5 & 2.5073 & 2.5045 & 1.0308 & 2.5072 & 2.5044 & \textbf{0.9999} \\
spread1 & 423.3354 & 439.0631 & 1.0396 & 423.0901 & 438.8086 & \textbf{0.9994} \\
spread2 & 107.0529 & 106.6815 & 1.0304 & 107.0333 & 106.6619 & \textbf{0.9998} \\
spread3 & 37.4213 & 37.6199 & 1.0445 & 37.4083 & 37.6069 & \textbf{0.9997} \\
spread4 & 24.1407 & 24.1058 & 1.0072 & 24.1347 & 24.0998 & \textbf{0.9998} \\
spread5 & 8.7974 & 8.8067 & 1.0049 & 8.7966 & 8.8058 & \textbf{0.9999} \\
2d-4c-no0 & 6.5977 & 6.5937 & 1.0464 & 6.3051 & 6.3012 & \textbf{0.9999} \\
2d-4c-no1 & 5.0339 & 5.0705 & 1.0154 & 4.9577 & 4.9938 & \textbf{0.9999} \\
2d-4c-no2 & 3.8840 & 3.4271 & 1.0212 & 3.8837 & 3.3560 & \textbf{0.9999} \\
2d-4c-no3 & 3.5216 & 3.5216 & 1.0020 & 3.5146 & 3.5146 & \textbf{0.9999} \\
2d-4c-no4 & 1.8235 & 1.8302 & 1.0348 & 1.8234 & 1.8303 & \textbf{0.9999} \\
2d-4c-no5 & 4.0648 & 4.0603 & 1.0001 & 4.0646 & 4.0601 & \textbf{0.9999} \\
2d-4c-no6 & 5.0136 & 5.0136 & 1.0026 & 5.0007 & 5.0007 & \textbf{0.9999} \\
2d-4c-no7 & 4.8158 & 5.3224 & 1.0027 & 4.8028 & 5.3080 & \textbf{0.9999} \\
2d-4c-no8 & 10.9662 & 10.8218 & 1.0070 & 10.8903 & 10.8212 & \textbf{0.9999} \\
2d-4c-no9 & 7.2298 & 7.1701 & 1.0149 & 7.1231 & 7.0643 & \textbf{0.9999} \\
10d-4c-no0 & 1.9592 & 1.9571 & 1.1234 & 1.9575 & 1.9554 & \textbf{0.9991} \\
10d-4c-no1 & 1.52825 & 1.5285 & 1.0540 & 1.5280 & 1.5280 & \textbf{0.9997} \\
10d-4c-no2 & 1.6823 & 1.6899 & 1.0068 & 1.6809 & 1.6886 & \textbf{0.9992} \\
10d-4c-no3 & 1.7986 & 1.7985 & 1.1495 & 1.7334 & 1.7332 & \textbf{0.9993} \\
10d-4c-no4 & 1.8202 & 1.8202 & 1.0941 & 1.8190 & 1.8190 & \textbf{0.9993} \\
10d-4c-no5 & 1.6006 & 1.6005 & 1.0466 & 1.5999 & 1.5999 & \textbf{0.9996} \\
10d-4c-no6 & 2.2364 & 2.2364 & 1.0561 & 2.2349 & 2.2349 & \textbf{0.9993} \\
10d-4c-no7 & 4.4895 & 4.4913 & 1.0346 & 4.4846 & 4.4864 & \textbf{0.9989} \\
10d-4c-no8 & 2.8754 & 2.8754 & 1.0443 & 2.8732 & 2.8732 & \textbf{0.9992} \\
10d-4c-no9 & 2.8778 & 2.8778 & 1.0614 & 2.8746 & 2.8746 & \textbf{0.9989} \\
\hline
B. scale & 1.0130 & 1.0129 & 1.0090 & 1.0080 & 1.0070 & \textbf{0.9973} \\
B. Tissue & 9.9587 & 9.1640 & 6.6150 & 12.0560 & 11.1250 & \textbf{3.5615} \\
wdbc & 144.6750 & 144.6750 & 1.0753 & 134.5460 & 134.5460 & \textbf{0.9810} \\
banknote & 8.7460 & 8.6840 & 1.0190 & 8.7450 & 8.6820 & \textbf{0.9990} \\
echo & 22.6990 & 22.7130 & \textbf{1.0530} & 105.7230 & 105.7820 & 4.6570 \\
Ecoli & 2.6260 & 3.6970 & 1.8710 & 2.5980 & 3.4520 & \textbf{1.0350} \\
Iris & 8.4800 & 8.2780 & \textbf{1.0080} & 10.5600 & 10.3000 & 1.2440 \\
magic & 1.1052 & 1.1052 & \textbf{1.0000} & 1.1052 & 1.1052 & 1.0047 \\
seeds & 25.2700 & 25.5100 & 2.3730 & 17.4700 & 17.6360 & \textbf{0.6910} \\
sonar & 4.1955 & 4.4084 & \textbf{1.0000} & 1.0508 & 1.0079 & 4.4088 \\
ukm & 2.7490 & 2.7490 & \textbf{0.9596} & 2.8640 & 2.8640 & 1.0590 \\
wine & 1.3861 & 1.3862 & 1.0145 & 1.3856 & 1.3857 & \textbf{0.9997} \\
colon cancer & 3.0893 & 3.0893 & \textbf{0.9999} & 1.0051 & 1.0051 & 3.0737 \\
lung cancer & 7.6850 & 7.6850 & \textbf{1.0000} & 7.7053 & 7.7053 & 1.0026 \\
prostate cancer & 2.9077 & 2.9081 & \textbf{0.9999} & 2.8927 & 2.8934 & 1.0051 \\
\hline
\end{tabular}
}
\end{table}

Schott's Spacing Metric (SSM) \cite{Liz08} is a measure of the diversity of the generated Pareto-optimal solutions. We use SSM to compare between the Pareto fronts of ECM-NSGA-II and ECM-MOEA/D and the set of solutions of MOGA and MOGA-SVM mapped to the same objective space. ECM-NSGA-II and ECM-MOEA/D obtain much higher values of SSM when compared with MOGA and MOGA-SVM, as detailed in Table \ref{tab_ss_results}. This attests to the fact that the objectives of MOGA and MOGA-SVM with a weak Pareto relation do not lend much diversity in terms of fuzziness. The entire set of solutions of MOGA and MOGA-SVM maps around the same point for the \emph{proximity1} dataset, as seen in Fig. \ref{fig_pareto_comp}. The higher values for ECM-MOEA/D compared to ECM-NSGA-II indicate that the former obtains a greater diversity of solutions. This can also be observed from Fig. \ref{fig_pareto_comp}.

We undertake further comparisons of the Pareto fronts using the Epsilon Indicator (EI) \cite{Liz08} which is the minimum factor by which all elements of the control Pareto front must be multiplied to have all its solutions dominated by a candidate Pareto front. Thus, values of EI lower than unity indicate dominance over the control front (for a minimization problem). Table \ref{tab_ie_results} lists the EI values with the Pareto fronts of ECM-NSGA-II and ECM-MOEA/D as control. The high values for MOGA and MOGA-SVM show that the Pareto fronts obtained by ECM-NSGA-II as well as ECM-MOEA/D dominate the solutions of MOGA and MOGA-SVM. The low values for ECM-NSGA-II w.r.t ECM-MOEA/D suggests that the Pareto front of the former generally dominates that of the latter, as can be seen from Fig. \ref{fig_pareto_comp} for the \emph{proximity1} dataset.

% ~~~~~~~~~~~~~~~~~~~~~~~~~~~~~ %

\section{Conclusions \& Future Work}

We propose a fuzzy CBC method by ECM using the MOO methods NSGA-II and MOEA/D, to produce fuzzy clusterings at different levels of fuzziness. The proposed methods are able to identify clusters with different levels of overlap, with ECM-NSGA-II producing slightly better results in terms of ARI. The MOEA/D variant, on the other hand, is generally observed to produce more uniformly spaced clusterings along Pareto fronts. Our experiments also suggest that ECM-MOEA/D is more resilient to closeness between clusters while ECM-NSGA-II is more resilient to clusters having disparate spreads. Hence, we recommend the use of ECM-NSGA-II when a single fuzzy clustering with high cluster compactness at an appropriate level of fuzziness is desired. ECM-MOEA/D may be used when a uniformly-spaced range of fuzzy clusterings at different levels of fuzziness is desired or when the clusters are known to be highly overlapping. Additionally, we present a method to select a suitable trade-off clustering from the Pareto front. Future investigations can be towards identifying an appropriate number of clusters, incorporating different distance metrics such as in multiple kernel clustering \cite{chen:2011,liu:2017}, or towards different methods such as fuzzy possibilistic clustering \cite{tsai:2012,yang:2006,saha:2018}.

\bibliographystyle{IEEEtran}
\bibliography{draft10}

%\end{document}

%%%%%%%%%% Merge with supplemental materials %%%%%%%%%%
%\widetext
\clearpage
\onecolumn
\begin{center}
\vspace{+10mm}
{\Huge Fuzzy Clustering to Identify Clusters at Different Levels of Fuzziness: An Evolutionary Multi-Objective Optimization Approach: Supplementary Material}

\vspace{+5mm}
{Avisek Gupta, Shounak Datta, and Swagatam Das, \emph{Senior Member, IEEE}}

\vspace{+10mm}
\end{center}
%%%%%%%%%% Merge with supplemental materials %%%%%%%%%%
%%%%%%%%%% Prefix a "S" to all equations, figures, tables and reset the counter %%%%%%%%%%
\setcounter{equation}{0}
\setcounter{figure}{0}
\setcounter{table}{0}
\setcounter{page}{1}
\makeatletter
\renewcommand{\theequation}{S\arabic{equation}}
\renewcommand{\thefigure}{S\arabic{figure}}
%\renewcommand{\bibnumfmt}[1]{[S#1]}
%\renewcommand{\citenumfont}[1]{S#1}
%%%%%%%%%% Prefix a "S" to all equations, figures, tables and reset the counter %%%%%%%%%%

\section{Tuning of parameters for ECM}

In this section we study the changes in the maximum Adjusted Rand Index (ARI) for different values of the parameters of the multi-objective optimization methods used. NSGA-II has the following paramters:

\begin{enumerate}
    \item \emph{pop}: Size of the population.
    \item \emph{FE}: Number of fitness evaluations.
    \item \emph{pool}: Fraction of the population undergoing genetic operations.
    \item \emph{tour}: During tournament selection, the number of solutions from which one is selected.
    \item \emph{mu}: Distribution index for crossover.
    \item \emph{mum}: Distribution index for mutation.
\end{enumerate}

MOEA/D has the following parameters:

\begin{enumerate}
    \item \emph{pop}: Size of the population.
    \item \emph{T}: Size of the neighbourhood.
    \item \emph{FE}: Number of fitness evaluations.
    \item \emph{F}: Parameter for mutation in Differential Evolution.
    \item \emph{Cr}: Parameter for crossover in Differential Evolution.
\end{enumerate}

\begin{figure}[b]
	\vspace{-3mm}
	\centering
	\subfloat[3 well-separated clusters]{\includegraphics[width=0.32\textwidth]{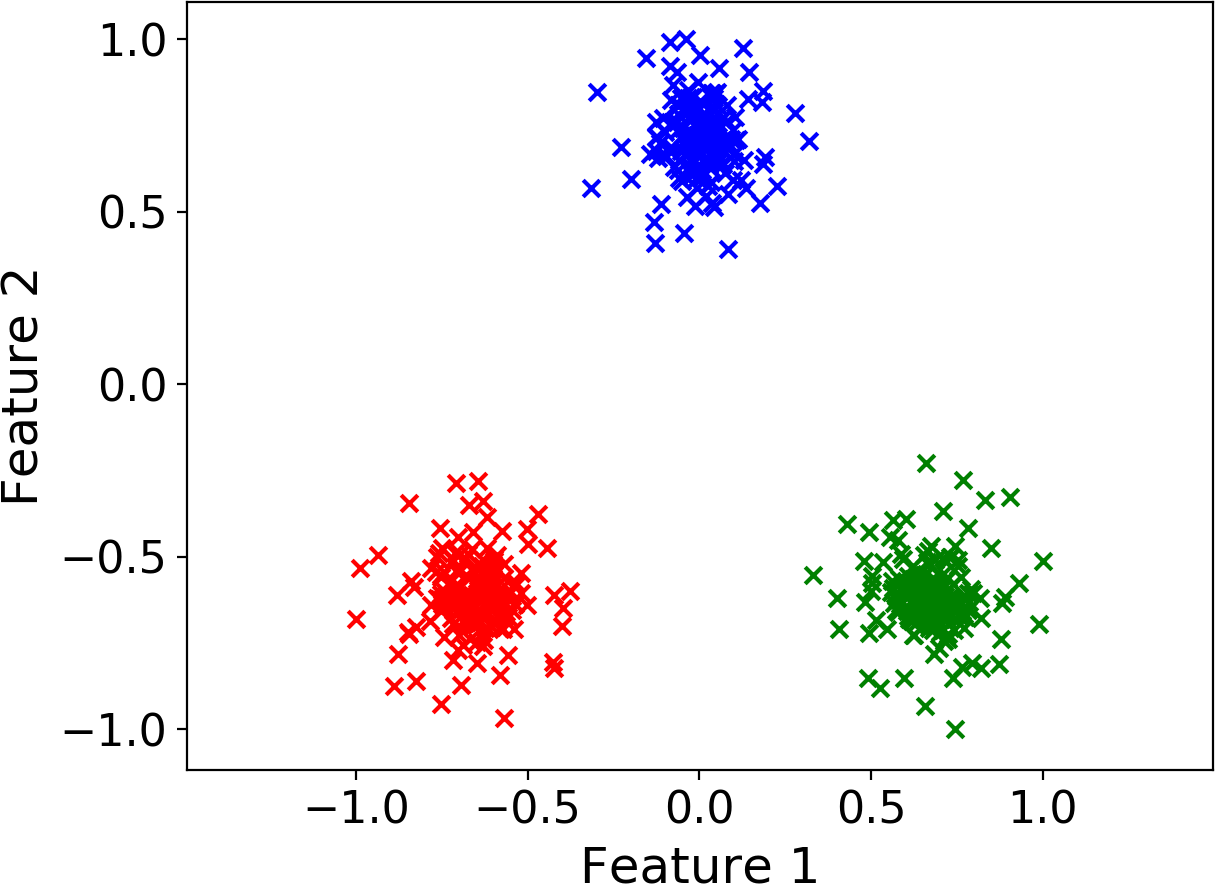}}
	\hfil
	\subfloat[3 clusters with slight overlap]{\includegraphics[width=0.32\textwidth]{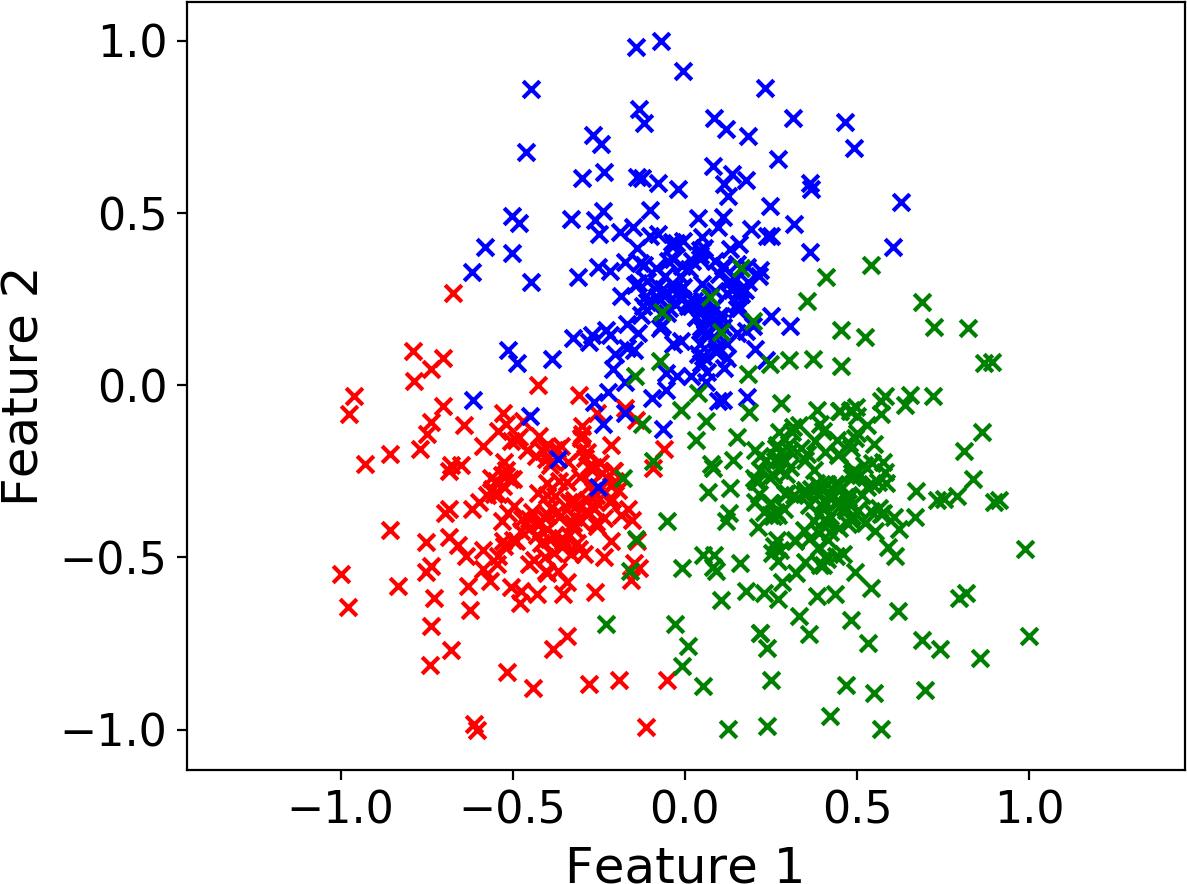}}
    \hfil
	\subfloat[3 clusters with high overlap]{\includegraphics[width=0.32\textwidth]{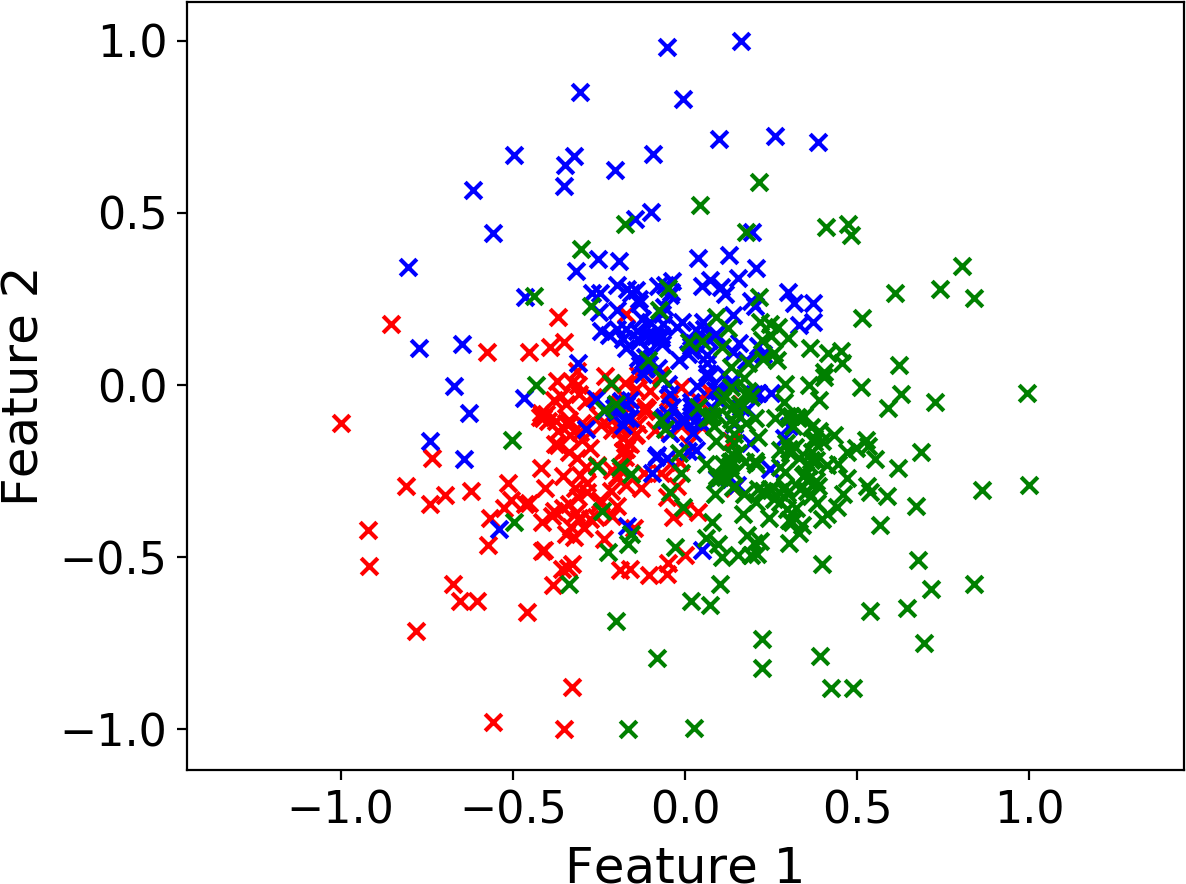}}
    \caption{Data set containing 3 clusters with different levels of overlap}
    \label{fig_3clusts}
\end{figure}

The effects of tuning the parameter values are observed over three datasets containing three clusters that are well-separated, slightly overlapped, or highly overlapped. The datasets are shown in Fig. \ref{fig_3clusts}.

\begin{figure}[h]
	\centering
	\includegraphics[width=0.8\textwidth]{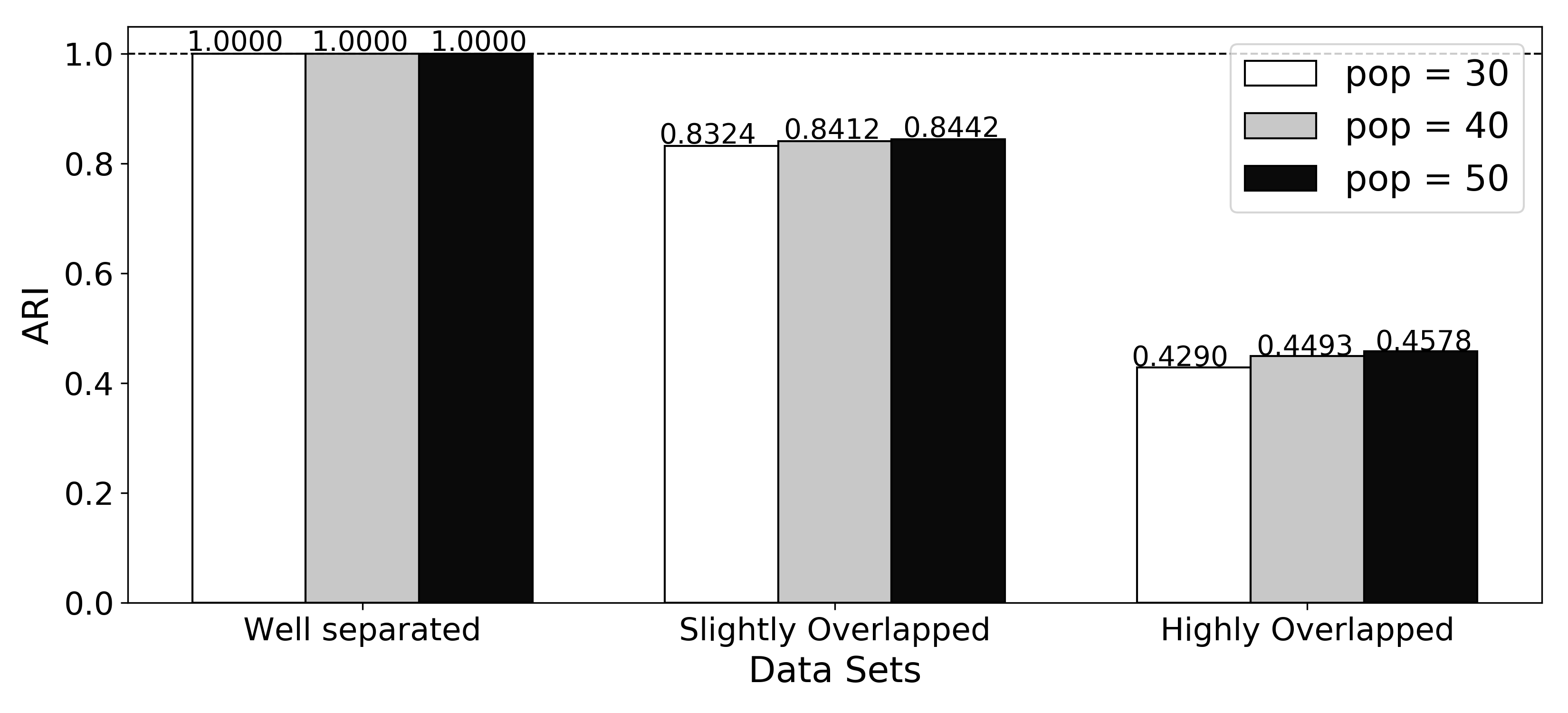}
    \caption{The variation in maximum ARI with variation in the population size ($pop$) for ECM-NSGA-II}
    \label{fig_nsga2_pop}
\end{figure}

\begin{figure}[h]
	\centering
	\includegraphics[width=0.8\textwidth]{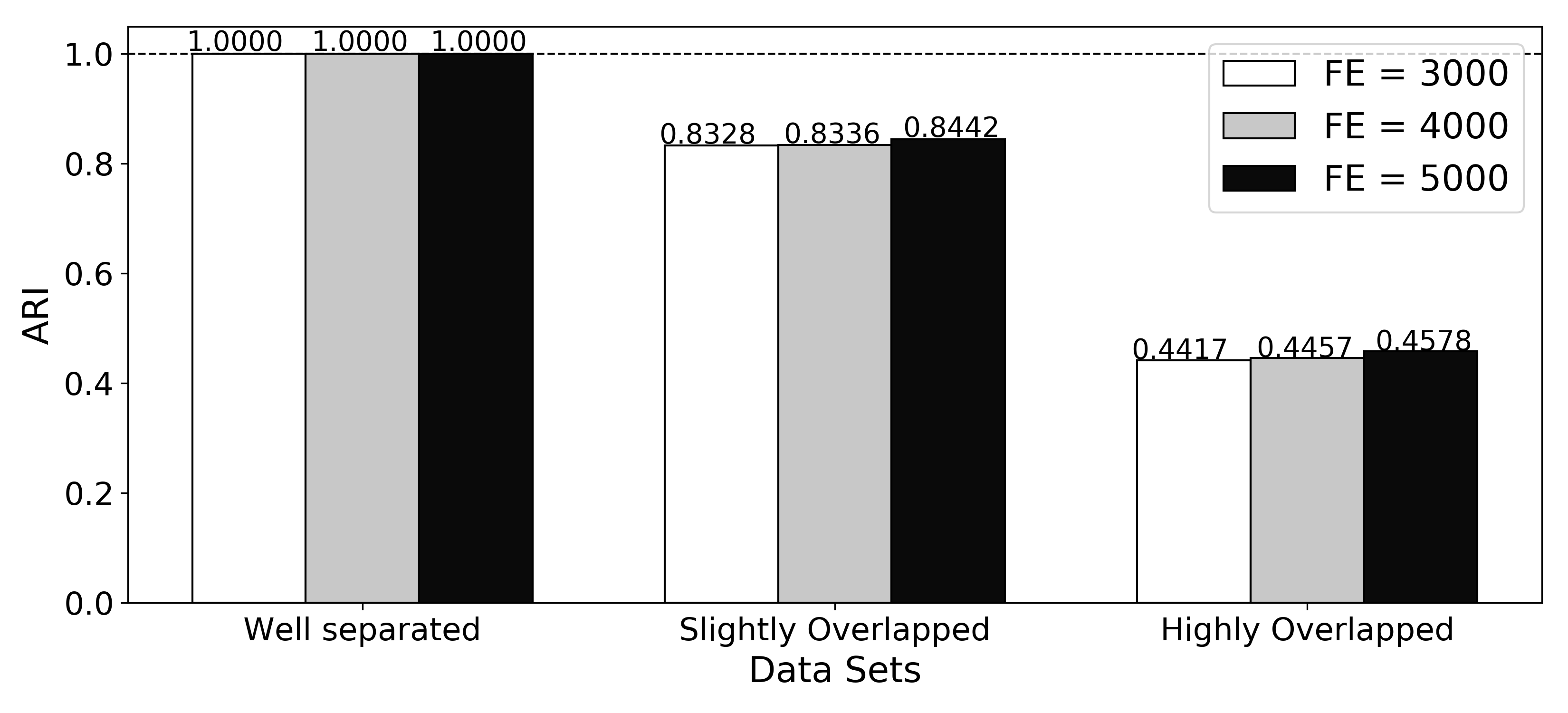}
    \caption{The variation in maximum ARI with variation in the number of fitness evaluations ($FE$) for ECM-NSGA-II}
    \label{fig_nsga2_FEmax}
\end{figure}

\begin{figure}[h]
	\centering
	\includegraphics[width=0.8\textwidth]{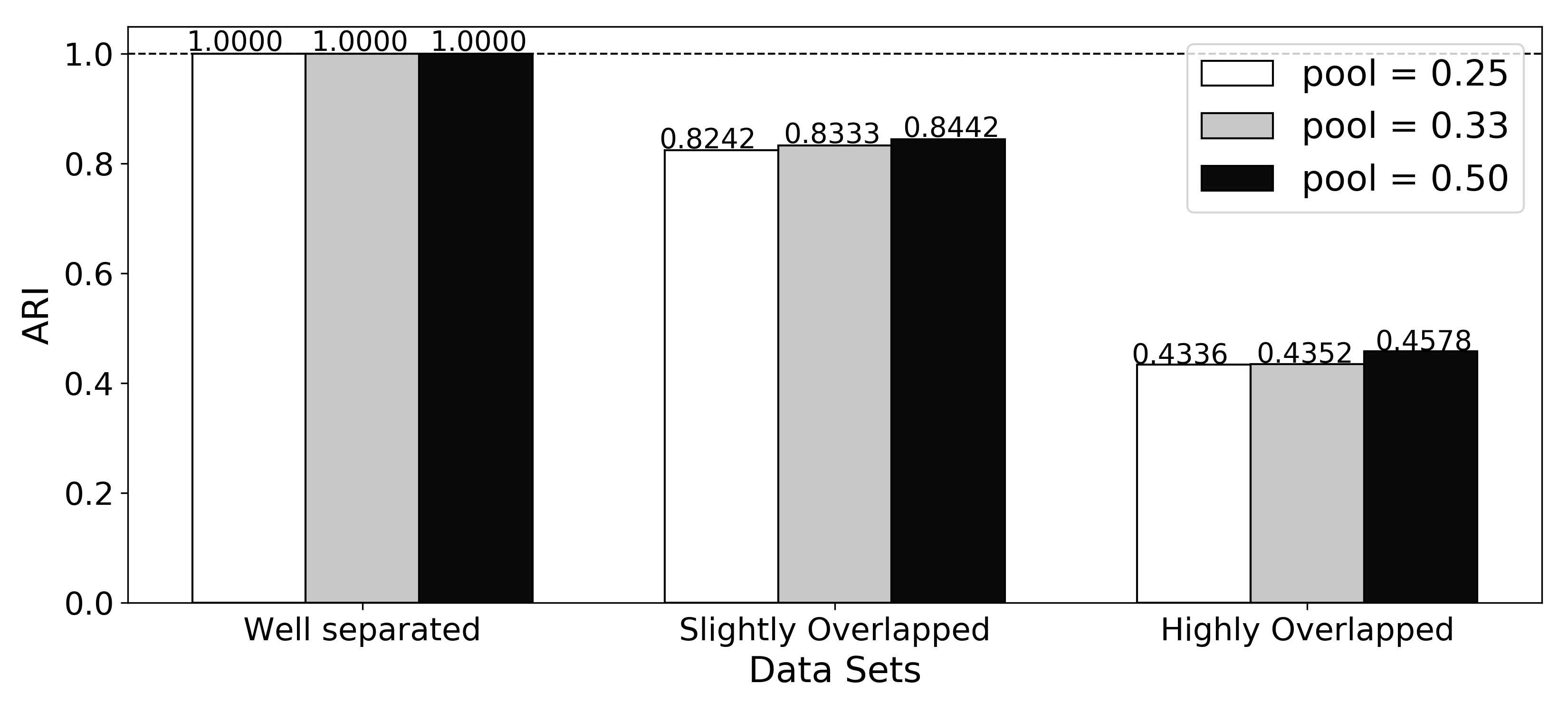}
    \caption{The variation in maximum ARI with variation in the fraction of population undergoing genetic operation ($pool$) for ECM-NSGA-II}
    \label{fig_nsga2_pool}
\end{figure}

\begin{figure}[h]
	\centering
	\includegraphics[width=0.8\textwidth]{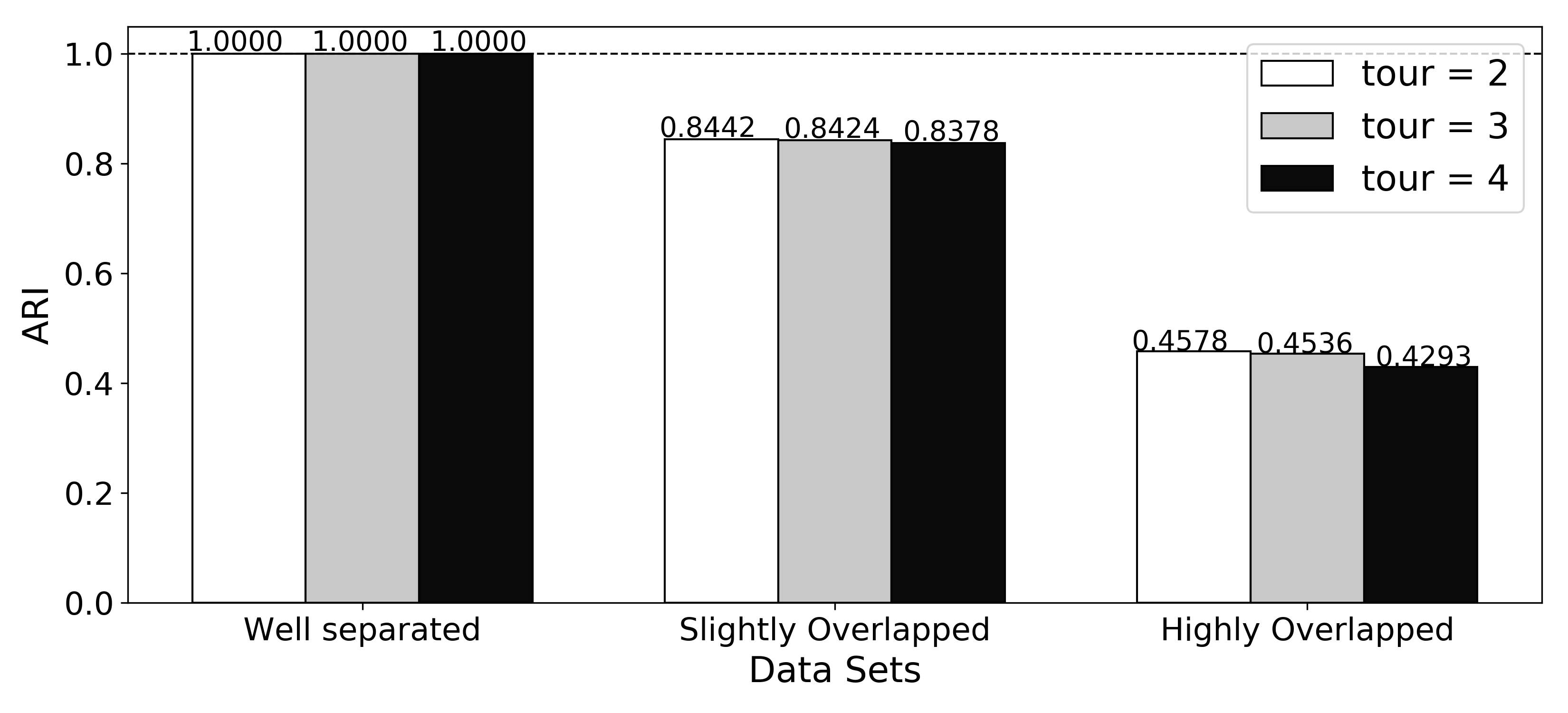}
    \caption{The variation in maximum ARI with variation in the number of solutions ($tour$) from which one solution is selected during tournament selection for ECM-NSGA-II}
    \label{fig_nsga2_tour}
\end{figure}

\begin{figure}[h]
	\centering
	\includegraphics[width=0.8\textwidth]{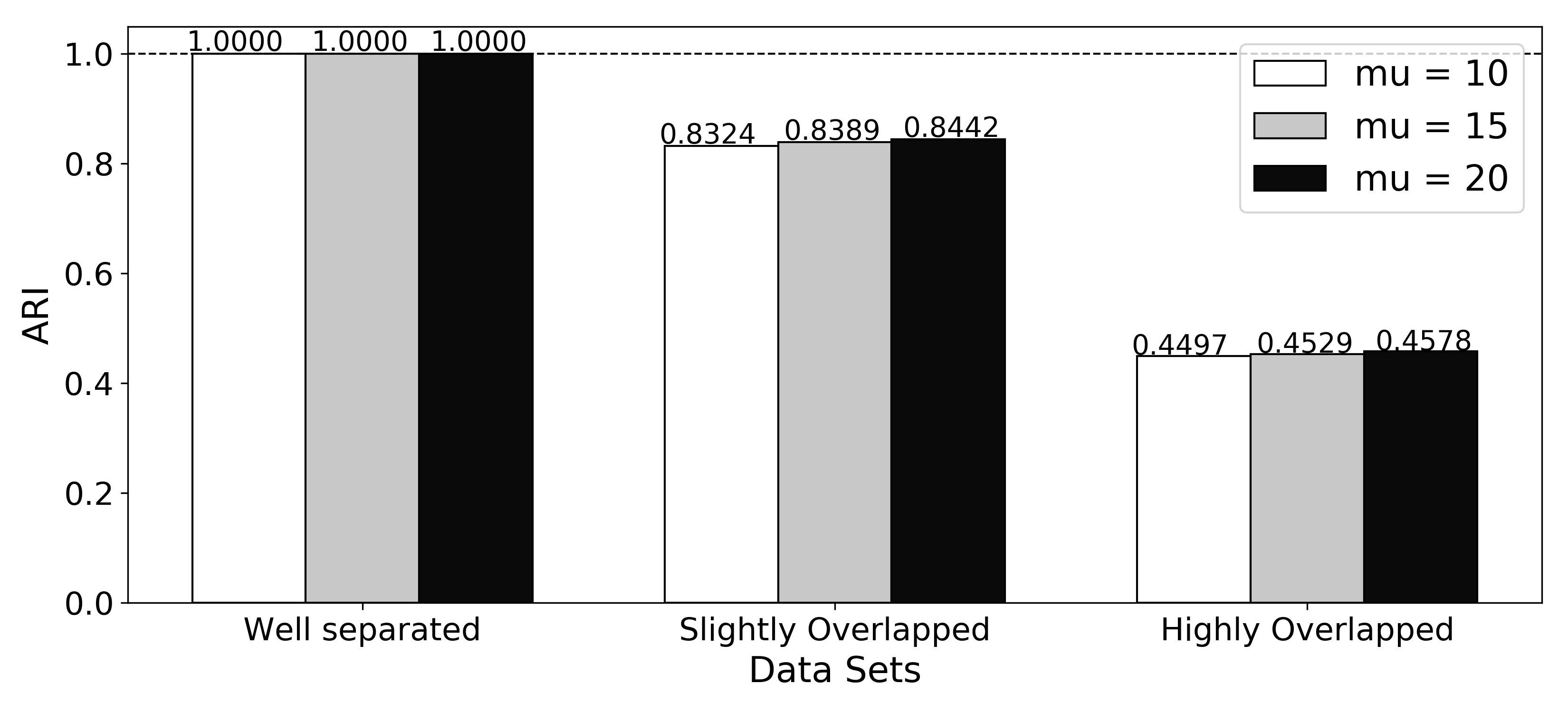}
    \caption{The variation in maximum ARI with variation in the distribution index for crossover ($mu$) for ECM-NSGA-II}
    \label{fig_nsga2_mu}
\end{figure}

\begin{figure}[h]
	\centering
	\includegraphics[width=0.8\textwidth]{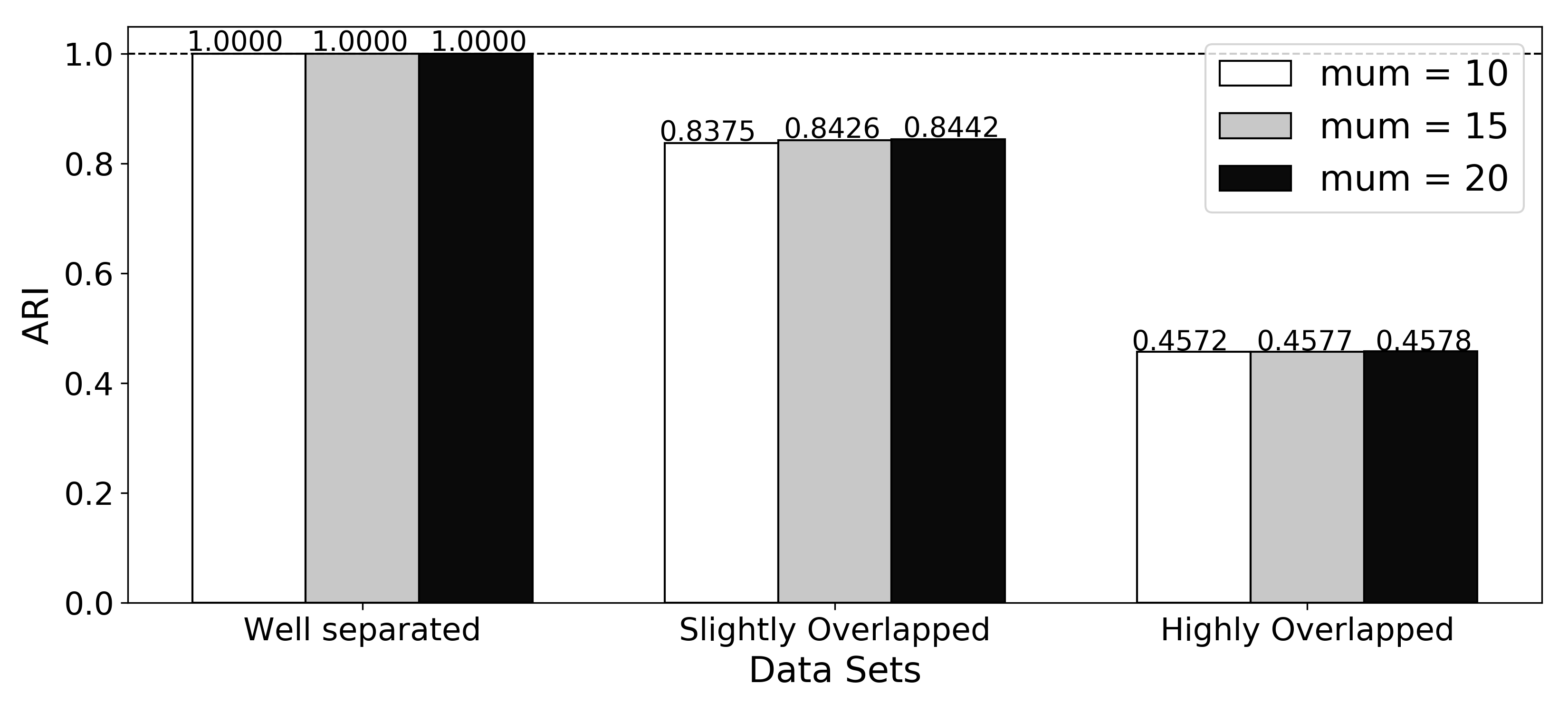}
    \caption{The variation in maximum ARI with variation in the distribution index for mutation ($mum$) for ECM-NSGA-II}
    \label{fig_nsga2_mum}
\end{figure}

In our experiments, we ran ECM-NSGA-II with the following parameter values: $pop=50$, $FE=5000$, $pool=0.5$, $tour=2$, $mu=20$, and $mum=20$. In Figs. \ref{fig_nsga2_pop} to \ref{fig_nsga2_mum}, we observe the variations in the best ARI achieved for different values of each parameter. When varying a single parameter, the other parameters are set to the values we used in our experiments. In each figure, one can observe that the variation in best ARI is very small, suggesting that the methods are quite resilient to the choice of parameters. For the parameters $pop$, $FE$, $pool$, $mu$ and $mum$, the best ARI achieved increases with increase in the parameter values considered in the experiments. For $tour$, the maximum ARI decreases with increase in parameter values.

\begin{figure}[h]
	\centering
	\includegraphics[width=0.8\textwidth]{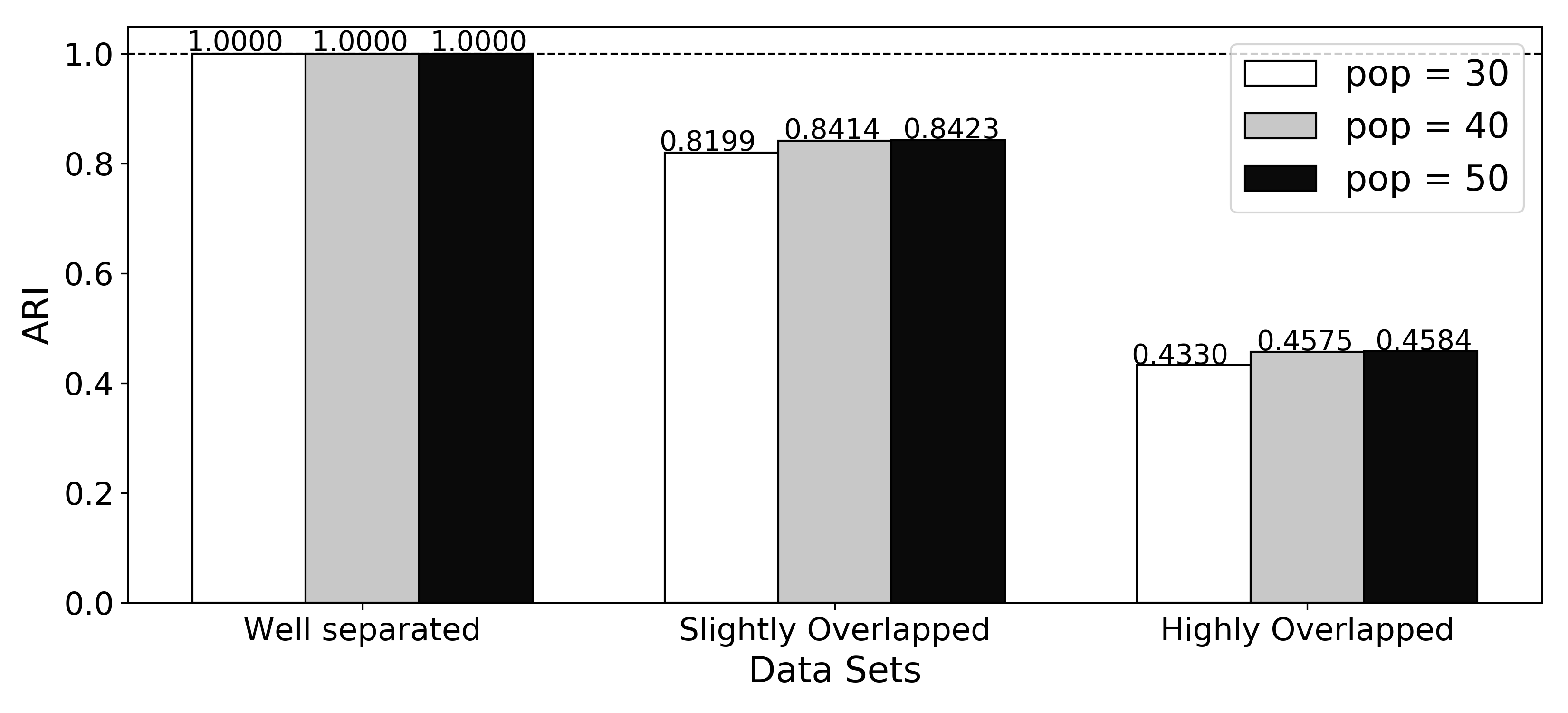}
    \caption{The variation in maximum ARI with variation in the population size ($pop$) for ECM-MOEA/D}
    \label{fig_moead_pop}
\end{figure}

\begin{figure}[h]
	\centering
	\includegraphics[width=0.8\textwidth]{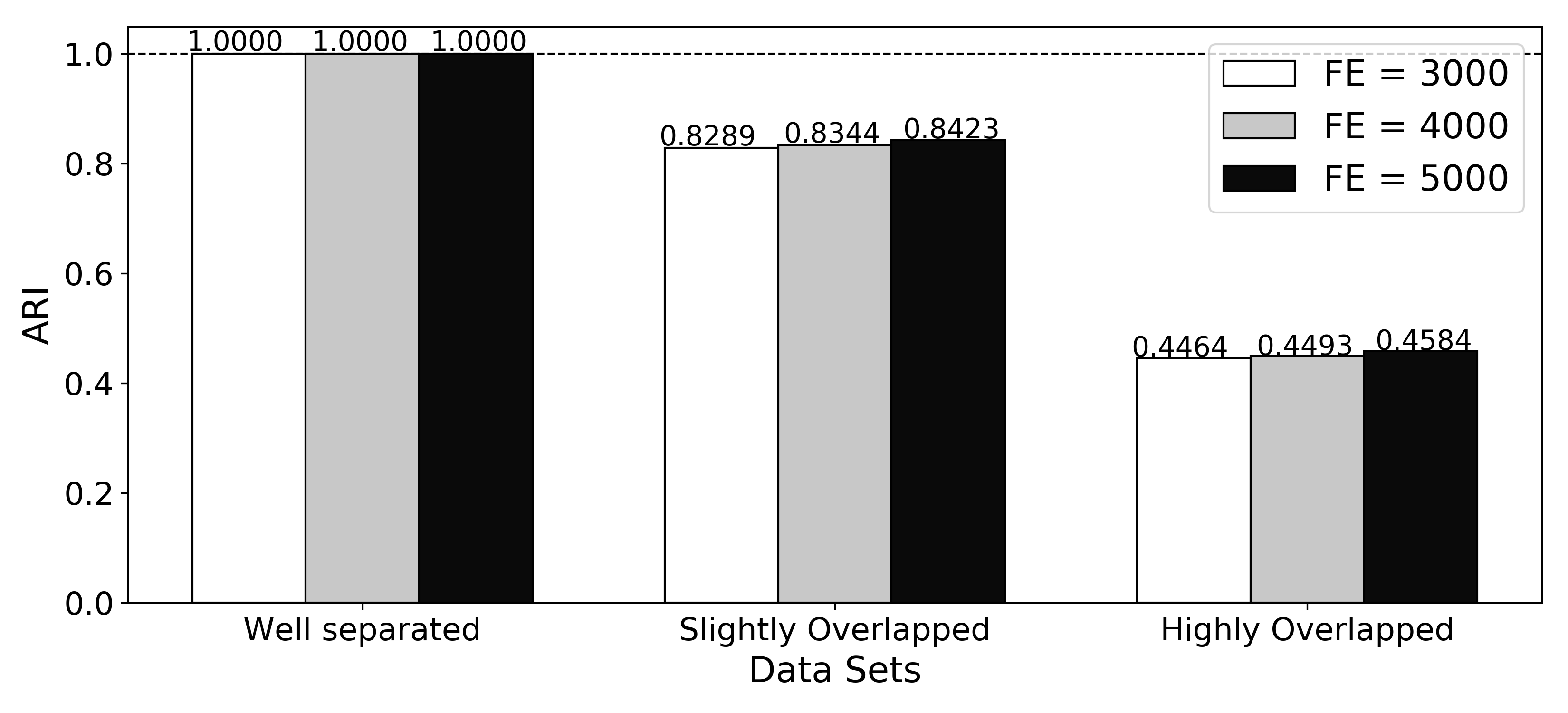}
    \caption{The variation in maximum ARI with variation in the number of fitness evaluations ($FE$) for ECM-MOEA/D}
    \label{fig_moead_FEmax}
\end{figure}

\begin{figure}[h]
	\centering
	\includegraphics[width=0.8\textwidth]{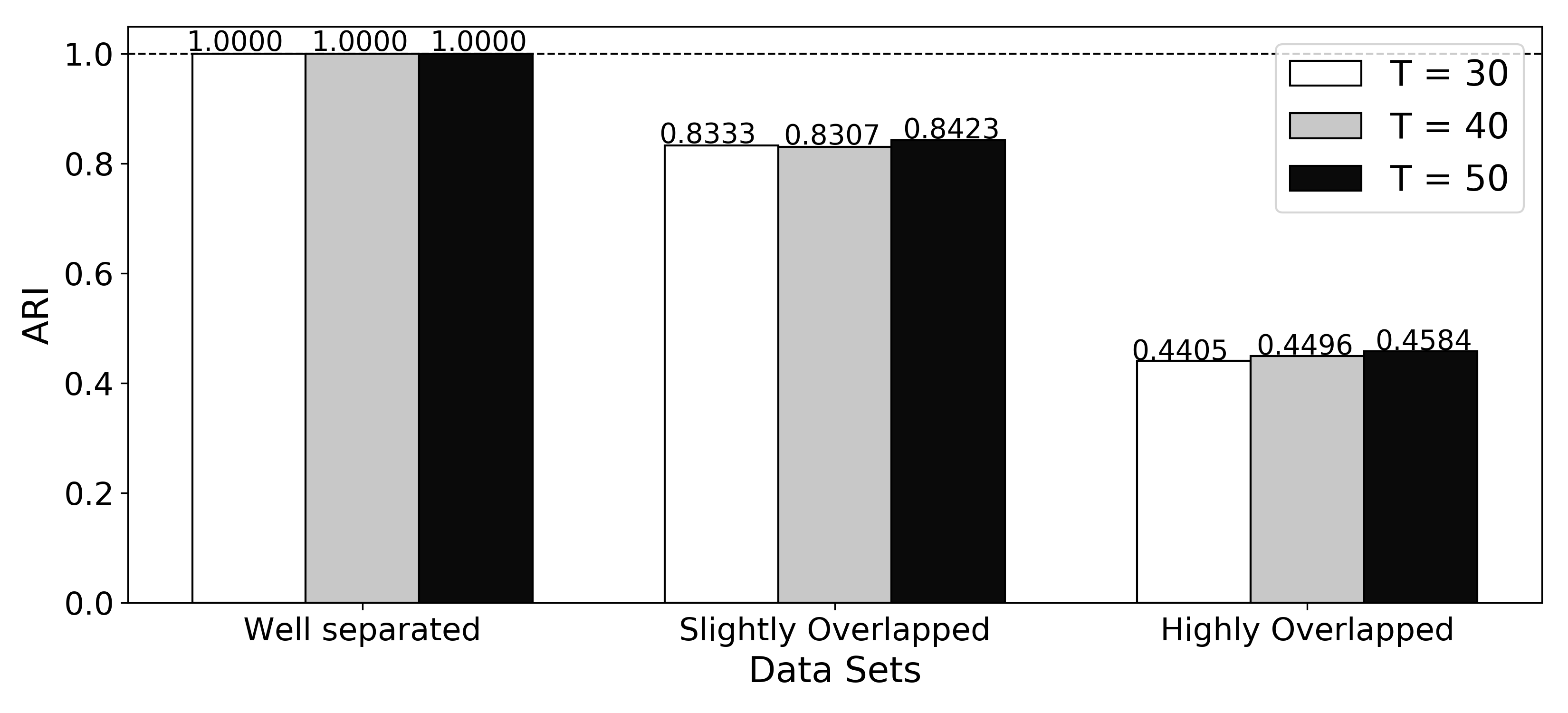}
    \caption{The variation in maximum ARI with variation in the neighbourhood size ($T$) for ECM-MOEA/D}
    \label{fig_moead_T}
\end{figure}

\begin{figure}[h]
	\centering
	\includegraphics[width=0.8\textwidth]{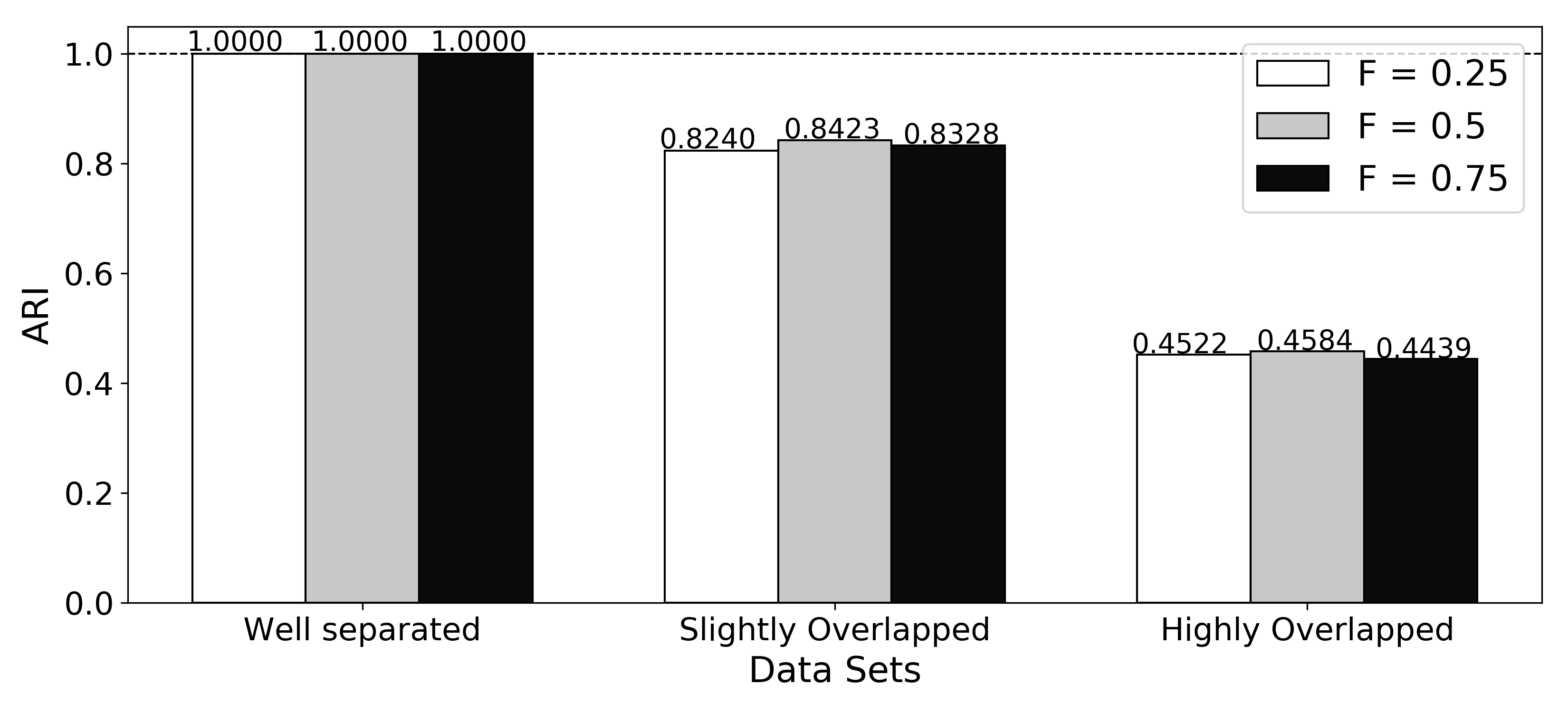}
    \caption{The variation in maximum ARI with variation in the mutation parameter $F$ in Differential Evolution for ECM-MOEA/D}
    \label{fig_moead_F}
\end{figure}

\begin{figure}[h]
	\centering
	\includegraphics[width=0.8\textwidth]{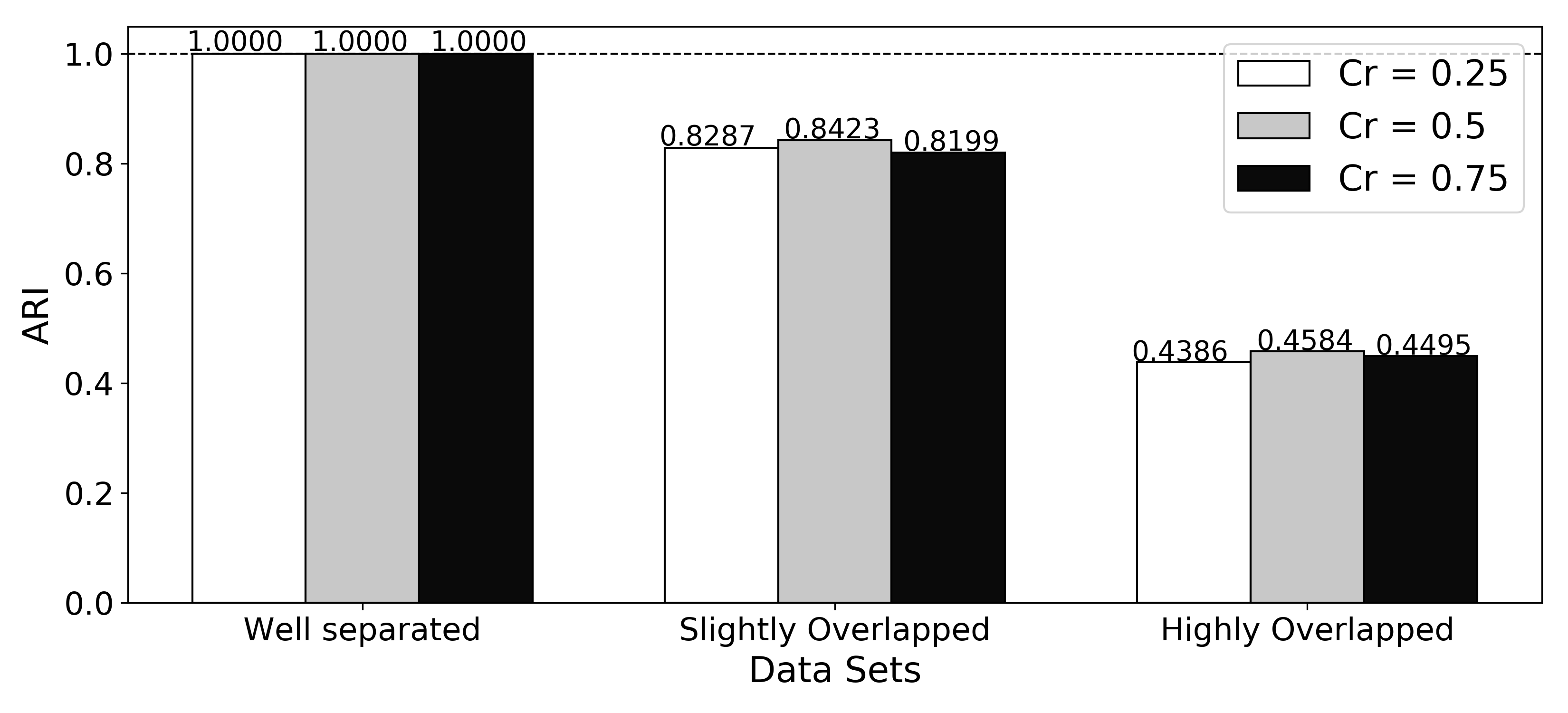}
    \caption{The variation in maximum ARI with variation in crossover parameter $Cr$ in Differential Evolution for ECM-MOEA/D}
    \label{fig_moead_Cr}
\end{figure}

Similarly, in our experiments, ECM-MOEA/D was executed with the following parameters values: $pop=50$, $FE=5000$, $T=50$, $F=0.5$, $Cr=0.5$. Figs. \ref{fig_moead_pop} to \ref{fig_moead_Cr} show small variations in the best ARI achieved with variations in each parameter, while keeping all other parameters fixed. This indicates that the methods are quite resilient to the choice of parameter values. For $pop$, $FE$ and $T$, the best ARI achieved increases with increase in the parameter values considered in the experiments. For $F$ and $Cr$, the best ARI is respectively achieved when $F=0.5$ and $Cr=0.5$.

\end{document}